\documentclass[10pt,journal, twocolumn]{IEEEtran}
\ifCLASSOPTIONcompsoc
\else
\fi

\ifCLASSINFOpdf
\else
\fi

\usepackage[table]{xcolor}
\usepackage{times}
\usepackage{epsfig}
\usepackage{graphicx}
\usepackage{amsmath}
\usepackage{amssymb}
\usepackage{epsfig}
\usepackage{graphicx}
\usepackage{amsmath}
\usepackage{amssymb}
\usepackage{bbm}
\usepackage{pifont}
\newcommand{\cmark}{\ding{51}}%
\newcommand{\xmark}{\ding{55}}%

\usepackage{array,setspace, amsfonts,url,bm, cellspace}
\usepackage{tikz}
\usepackage{epstopdf}
\usetikzlibrary{positioning}
\usetikzlibrary{arrows}
\usepackage{pifont}
\usepackage{multirow, hhline}
\usetikzlibrary{decorations.text}
\usetikzlibrary{shapes.geometric,calc}
\definecolor{mygray}{RGB}{208,208,208}
\definecolor{mymagenta}{RGB}{226,0,116}

\newcolumntype{M}[1]{>{\centering\arraybackslash}m{#1}}

\newcolumntype{C}[1]{>{\centering\let\newline\\\arraybackslash\hspace{0pt}}m{#1}}

\usepackage[pagebackref=true,breaklinks=true,letterpaper=true,colorlinks,bookmarks=false]{hyperref}

\begin{document}

\title{The P-DESTRE: A Fully Annotated Dataset for Pedestrian Detection, Tracking, Re-Identification and Search from Aerial Devices}

\author{\IEEEauthorblockN{S.V. Aruna Kumar, Ehsan Yaghoubi, Abhijit Das, B.S.~Harish and Hugo Proen\c{c}a,~\IEEEmembership{Senior Member,~IEEE}}
\IEEEcompsocitemizethanks{A. Kumar, E. Yaghoubi and H. Proen\c{c}a are with the IT: Instituto de Telecomunica\c{c}\~{o}es, Department of Computer Science, University of Beira Interior, Portugal, E-mail: arunkumarsv55@gmail.com, D2389@ubi.pt, hugomcp@di.ubi.pt}
\IEEEcompsocitemizethanks{A. Das is with the India Statistical Institute, Kolkata, India, E-mail: abhijitdas2048@gmail.com}
\IEEEcompsocitemizethanks{B. Harish is with the Department of Information Science and Engineering, JSS Science and Technology University, Mysuru, India, E-mail: bsharish@jssstuniv.in}
\thanks{Manuscript received ? ?, 2020; revised ? ?, ?.}}

\markboth{arxiv,~2020}%
{Shell \MakeLowercase{\textit{et al.}}: Bare Demo of IEEEtran.cls for Computer Society Journals}

\IEEEtitleabstractindextext{%
\begin{abstract}

Over the last decades, the world has been witnessing growing threats to the security in urban spaces, which has augmented the relevance given to visual surveillance solutions able to detect, track and identify persons of interest in crowds. In particular, unmanned aerial vehicles (UAVs) are a potential tool for this kind of analysis, as they provide a cheap way for data collection, cover large and difficult-to-reach areas, while reducing human staff demands. In this context, all the available datasets are exclusively suitable for the pedestrian \emph{re-identification} problem, in which the multi-camera views per ID are taken on a single day, and allows the use of clothing appearance features for identification purposes. Accordingly, the main contributions of this paper are two-fold: 1) we announce the UAV-based P-DESTRE dataset, which is the first of its kind to provide consistent ID annotations across multiple days, making it suitable for the extremely challenging problem of \emph{person search}, i.e., where no clothing information can be reliably used. Apart this feature, the P-DESTRE annotations enable the research on UAV-based pedestrian detection, tracking, re-identification and soft biometric solutions; and 2) we compare the results attained by state-of-the-art pedestrian detection, tracking, re-identification and search techniques in well-known surveillance datasets, to the effectiveness obtained by the same techniques in the P-DESTRE data. Such comparison enables to identify the most problematic data degradation factors of UAV-based data for each task, and can be used as baselines for subsequent advances in this kind of technology. The dataset and the full details of the empirical evaluation carried out are freely available at \url{http://p-destre.di.ubi.pt/}.

\end{abstract}

}

\maketitle

\IEEEdisplaynontitleabstractindextext

\IEEEpeerreviewmaketitle

\section{Introduction}

\IEEEPARstart{V}ideo-based surveillance regards \emph{the act of watching a person or a place, esp. a person believed to be involved with criminal activity or a place where criminals gather}\footnote{\url{https://dictionary.cambridge.org/dictionary/english/surveillance}}. Over the years, this kind of technologies has been used in far more applications than their roots in crimes detection, such as traffic control and management of physical infrastructures. The pioneer generation of video surveillance systems was based in closed-circuit television (CCTV) networks, being limited by the stationary nature of the cameras. More recently, unmanned aerial vehicles (UAVs) have been regarded as a solution to overcome such limitations: UAVs provide a fast and cheap way for data collection, and can easily assess confined spaces, producing minimal noise while reducing the staff demands and cost. 

\begin{figure}[ht!]
\begin{center}
\begin{tikzpicture}

\def\deltaX{0}

\draw (-0.85, 0) node[rectangle, rotate=90] {\scriptsize{\textbf{Re-identification}}};  

\draw (0+\deltaX,0) node(n1)  {\includegraphics[height=2.25 cm]{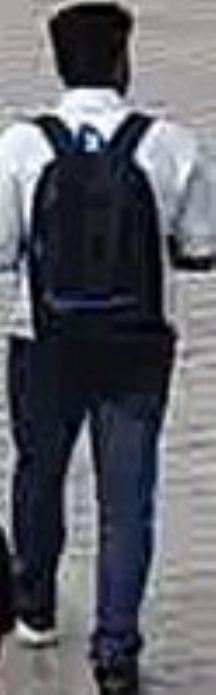}};       
\draw (1+\deltaX,0) node(n1)  {\includegraphics[height=2.25 cm]{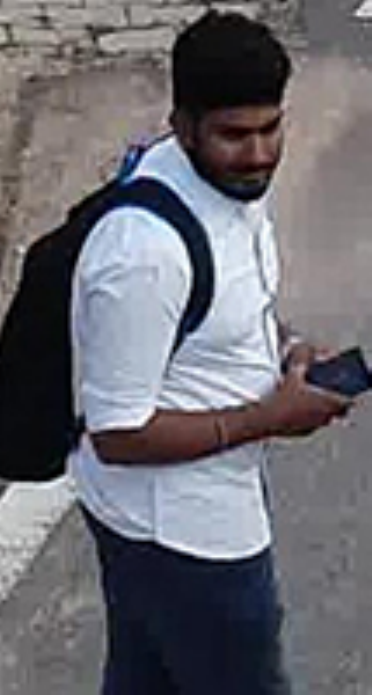}};       
\draw (2.05+\deltaX,0) node(n1)  {\includegraphics[height=2.25 cm]{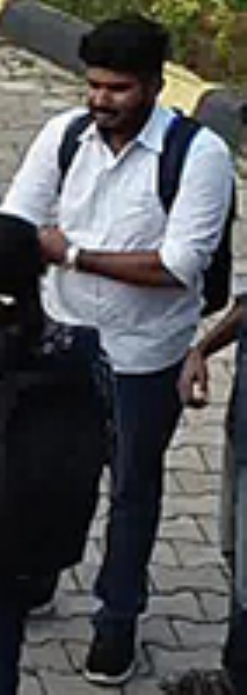}};       
\draw (2.95+\deltaX,0) node(n1)  {\includegraphics[height=2.25 cm]{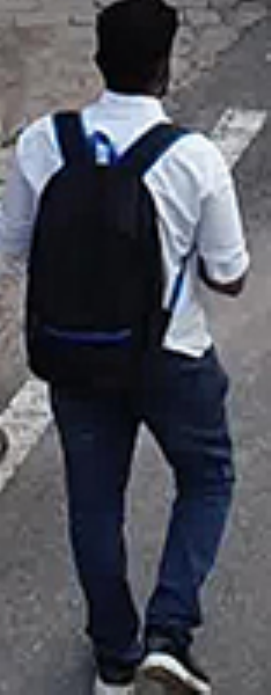}};    

\draw (4.15+\deltaX,0) node(n1)  {\includegraphics[height=2.25 cm]{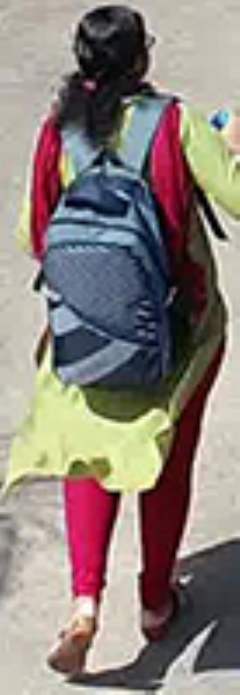}};    
\draw (5.00+\deltaX,0) node(n1)  {\includegraphics[height=2.25 cm]{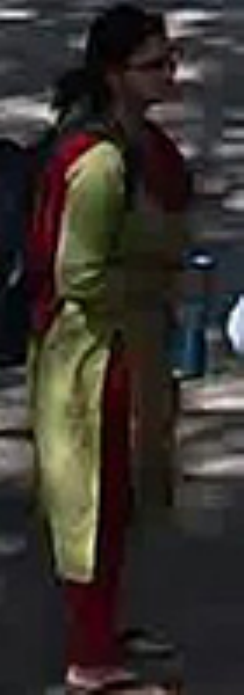}};    
\draw (5.850+\deltaX,0) node(n1)  {\includegraphics[height=2.25 cm]{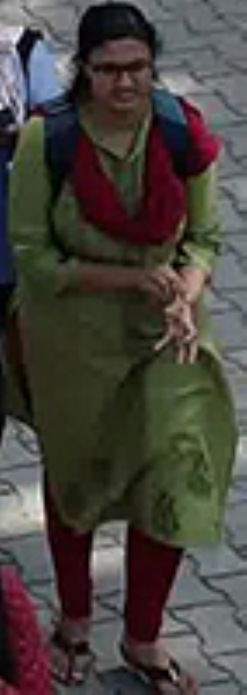}};    
\draw (6.65+\deltaX,0) node(n1)  {\includegraphics[height=2.25 cm]{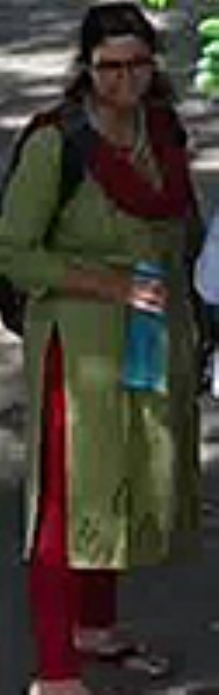}};    

\draw [very thick, rounded corners, red ] (-1.175+\deltaX, -1.275) rectangle (7.15+\deltaX,1.275);

\def\deltaY{-2.75}

\draw (-0.85, 0+\deltaY) node[rectangle, rotate=90] {\scriptsize{\textbf{Search}}};  

\draw (0,0+\deltaY) node(n1)  {\includegraphics[height=2.25 cm]{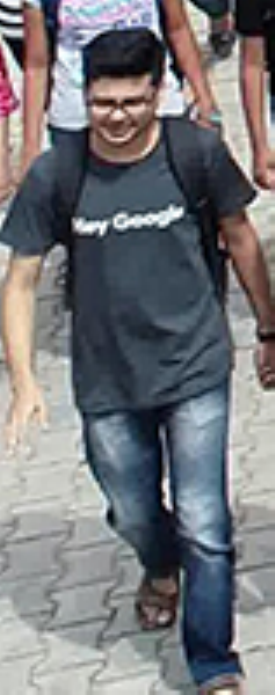}};       
\draw (1,0+\deltaY) node(n1)  {\includegraphics[height=2.25 cm]{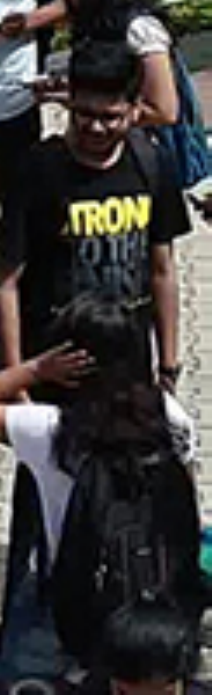}};       
\draw (2.05,0+\deltaY) node(n1)  {\includegraphics[height=2.25 cm]{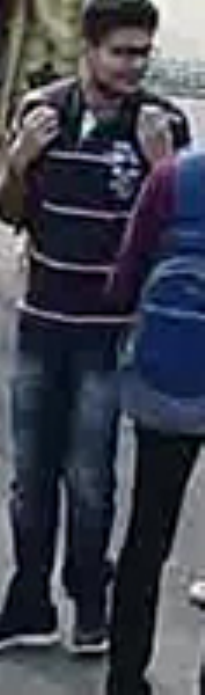}};       
\draw (2.95,0+\deltaY) node(n1)  {\includegraphics[height=2.25 cm]{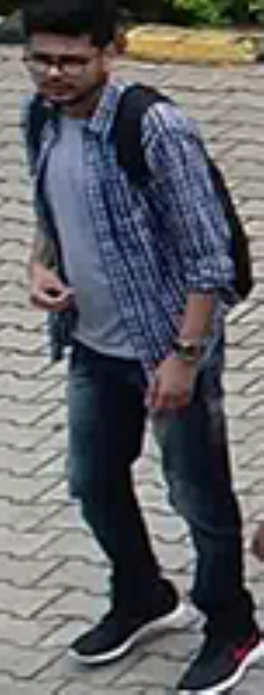}};    

\draw (4.15,0+\deltaY) node(n1)  {\includegraphics[height=2.25 cm]{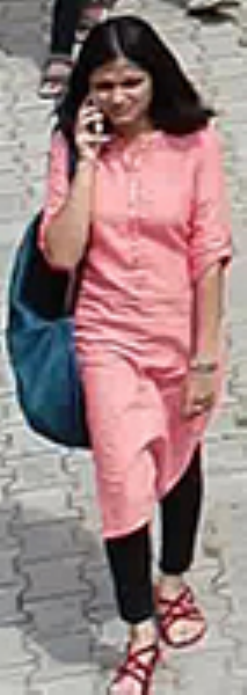}};    
\draw (5.00,0+\deltaY) node(n1)  {\includegraphics[height=2.25 cm]{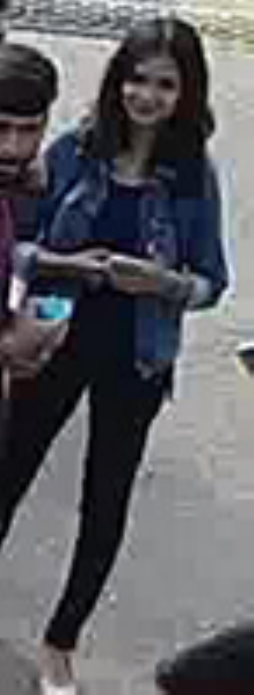}};    
\draw (5.850,0+\deltaY) node(n1)  {\includegraphics[height=2.25 cm]{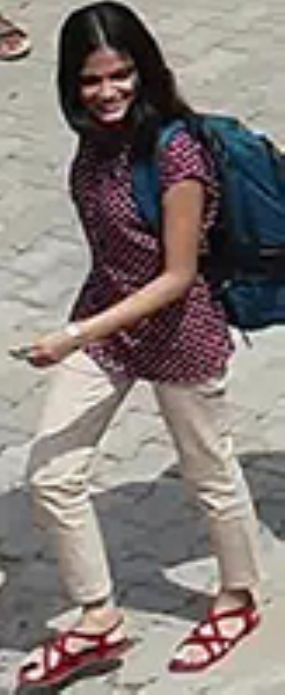}};    
\draw (6.65,0+\deltaY) node(n1)  {\includegraphics[height=2.25 cm]{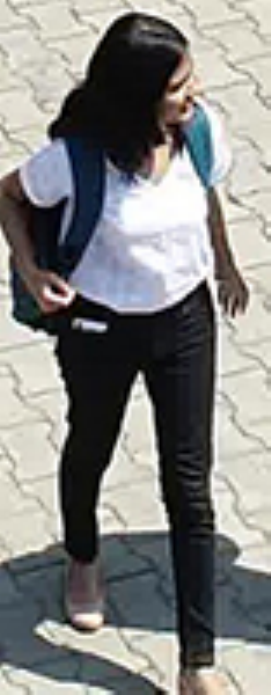}};    

\draw [very thick, rounded corners, green ] (-1.175, -1.275+\deltaY) rectangle (7.15,1.275+\deltaY);     

\draw [dashed, very thick] (3.6, 1.2) -- (3.6,-1.2);

\draw [dashed, very thick] (3.6, 1.2+\deltaY) -- (3.6,-1.2+\deltaY);

\end{tikzpicture}
    \caption{Key difference between the pedestrian \emph{re-identification} (upper row) and \emph{search} (bottom row) problems. In the former case, it is assumed that the subjects keep the same clothes between consecutive observations, which does not happen in the person search problem. This feature increases significantly the difficulties of correctly matching identities, as most of the state-of-the-art human re-identification techniques rely in clothing appearance-based features.}
        \label{fig:key}
    \end{center}
\end{figure}
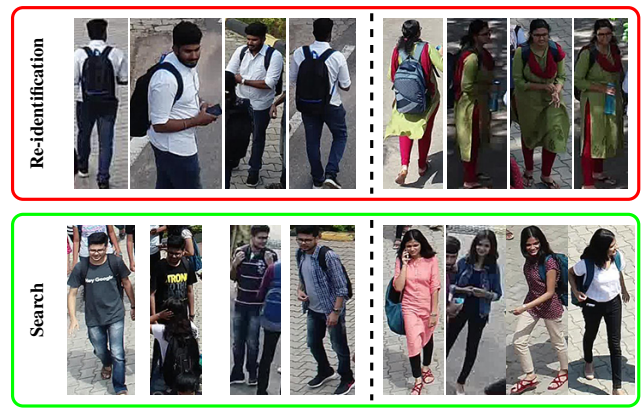

Being at the core of video surveillance, many efforts have been put in the development of pedestrian analysis methods that work in \emph{real-world} conditions, which is seen as a \emph{grand challenge}\footnote{\url{https://en.wikipedia.org/wiki/Grand_Challenges}}. In particular, the problem of identifying pedestrians in crowded scenes, based on low resolution data and partially occluded silhouettes, becomes especially difficult when the time elapsed between consecutive observations of a person denies the use of clothing-based features (bottom row of Fig.~\ref{fig:key}).

To date, the research on pedestrians analysis has been conducted on databases (e.g.,~\cite{Li2014},~\cite{Singh2018} and~\cite{Grigorev2019}) with two main weaknesses: 1) they contain data with short lapses of time between consecutive observations of each ID (typically within a single day), which allows to use clothing appearance features in identity matching (top row of Fig.~\ref{fig:key}); 2) they have a limited availability of soft biometric annotations, which denies the use of this kind of features to prune the space of  identities possible for a query. There are even datasets related to other problems (e.g., such as gait recognition~\cite{Zheng2011}), that have been used in surveillance experiments, but where the data acquisition conditions are highly different of the typically seen in real-world environments.

As a tool to support further advances in UAV-based pedestrian analysis, the P-DESTRE is the result of a joint effort from researchers in two universities of Portugal and India. It is a multi-session set of UAV-based videos, taken in outdoor crowded environments. ''\emph{DJI Phantom 4}''\footnote{\url{https://www.dji.com/pt/phantom-4}} drones controlled by human operators flew over various scenes of both universities \emph{campi}, with the data acquired to simulate the everyday conditions in urban environments. All the subjects offered explicitly as volunteers and they were asked to simply ignore the UAVs. Also, the P-DESTRE set is fully annotated at the frame level (by human experts), providing three families of meta-data:

1) \textbf{Bounding boxes}. The position of each pedestrian at every frame of each scene is provided as a bounding box, which enables to use the data for object detection, tracking and semantic segmentation purposes;

2) \textbf{Soft biometrics labels}. Each pedestrian is fully characterised by 16 labels: \{'gender', 'age', 'height', 'body volume', 'ethnicity', 'hair colour', 'hairstyle', 'beard', 'moustache', 'glasses', 'head accessories',  'body accessories', 'action' and 'clothing information' (x3)\}, which also allows to use the data for soft biometrics and action recognition problems;

3) \textbf{IDs}. Each pedestrian has a unique identifier consistent over all the data acquisition days/sessions, which is the  feature of the dataset that makes it suitable for various identification problems. The \emph{unknown} identities are also annotated, which enables to use them as distractors and augment the challenges in performing robust identification.
 
As a consequence of the above types of annotation, the key discriminating feature between the P-DESTRE and related datasets is the \emph{pedestrian search} problem, where the data is acquired over large lapses of time (e.g., various days/weeks), keeping consistent ID labels between observations. In this problem, the identification techniques cannot rely in clothing appearance-based features, which is the key property that distinguishes \emph{search} from the (less challenging) \emph{re-identification} problem (Fig.~\ref{fig:key}), where the consecutive observations of each ID are assumed to have been taken in short intervals of time and clothing appearance features can be reliably used.

In summary, we provide the following contributions:

\begin{itemize}
\item we announce the free availability of the P-DESTRE dataset fore research purposes, which is the first of its kind that is fully annotated at the frame level, and designed to support the research on UAV-based person search. In addition, the P-DESTRE set can be used in human detection, tracking, re-identification, and soft biometrics experiments. It is composed of over 14 million bounding boxes, extracted from video sequences containing 261 known identities; 
\item we provide a systematic review of the related work in the scope of the P-DESTRE dataset, and compare its main discriminating features with respect to the related sets;
\item We report the results that state-of-the-art methods in pedestrian detection, tracking, re-identification and search attain in UAV-based data. To serve as baselines, upon our empirical evaluation, we also provide the results attained by the same techniques in well-known visual surveillance data sets; 
\item We discuss the strengths and weaknesses of the existing solutions for each of the four problems considered, pointing for the further improvements that are required to enable the deployment of this kind of technologies focused. \\
\end{itemize}

The remainder of this paper is organized as follows: Section~\ref{sec:Related} summarizes the most relevant research in the scope of the  novel dataset. Section~\ref{sec:ProposedSet} provides a detailed description of the P-DESTRE data. Section~\ref{sec:Results} discusses the results observed in our empirical evaluation, and the conclusions are given in Section~\ref{sec:Conclusions}.

\section{Related Work}
\label{sec:Related}

This section summarizes the previous works in the scope of the P-DESTRE dataset. We start by describing the most relevant UAV-based datasets for general object detection and tracking purposes. Then, we pay special attention to datasets that focus particularly the problems of pedestrian detection, tracking, re-identification and search, comparing them from various perspectives.

\subsection{UAV-Based Datasets}

Various datasets of UAV-based data are available to the research community, with most of them serving for object detection and tracking purposes. The 'Object deTection in Aerial images'~\cite{Xia2018} set supports research on multi-class object detection, and has 2,806 images that contain over 188K instances of 15 categories. The 'Stanford drone dataset'~\cite{Robiquet2016} provides video data for object tracking, containing 60 videos from 8 scenes, annotated for 6 classes of objects. Similarly, the 'UAV123'~\cite{Mueller2016} set  provides 123 video sequences from aerial viewpoints, that contain more than 110K frames annotated with bounding boxes for object detection/tracking.
The 'VisDrone'~\cite{Zhu2018} consists of 288 videos/261,908 frames, with over 2.6M bounding boxes covering pedestrians, cars, bicycles, and tricycles. Finally, the largest freely available source is the 'Multidrone'~\cite{Mademlis2019} set, that provides data for multiple category object detection and tracking analysis. It contains videos of many different actions, under various weather conditions and in multiple places, yet not all the data are annotated.

\subsection{Pedestrian Analysis Datasets}

\begin{table*}[h]
    \centering
     \caption{Comparison between the P-DESTRE and the existing datasets that support the research in pedestrian detection, tracking and re-identification (appearing in chronological order).}
    \begin{tabular}{|p{1.75cm}|c|c|c|c|c|c|c|c|c|c|c|}\hline
         \multirow{2}{*}{ \textbf{\scriptsize{Dataset}}} &         \multirow{2}{*}{ \textbf{\scriptsize{Camera}}} &          \multirow{2}{*}{\textbf{\scriptsize{Format}}} & \multicolumn{5}{c|}{ \textbf{\scriptsize{Task}}}&   \multirow{2}{*}{\textbf{\scriptsize{Identities}}} &   \multirow{2}{*}{\textbf{\scriptsize{Bound. Box}}}  &          \multirow{2}{*}{\textbf{\scriptsize{Environment}}} & \multirow{2}{*}{\textbf{\scriptsize{Height (m)}}} \\ \cline{4-8}
         
         & & &  \textbf{\scriptsize{Detection}} & \textbf{\scriptsize{Tracking}} & \textbf{\scriptsize{ReID}} & \textbf{\scriptsize{Search}}& \textbf{\scriptsize{Action Rec.}} & & & & \\ \hline

        \scriptsize{PRID-2011~\cite{Hirzer2011}} & \scriptsize{UAV} & \scriptsize{Still} & \xmark & \xmark & \cmark &\xmark & \xmark & \scriptsize{1,581} & \scriptsize{40K}  & \scriptsize{Surveillance} & \scriptsize{$[20, 60]$ } \\ \hline

        \scriptsize{CUHK03~\cite{Li2014}} & \scriptsize{CCTV} & \scriptsize{Still} & \xmark & \xmark & \cmark &\xmark & \xmark & \scriptsize{1,467}& \scriptsize{13K}  & \scriptsize{Surveillance} & \scriptsize{-} \\ \hline

        \scriptsize{iLIDS-VID~\cite{Wang2014}} & \scriptsize{CCTV} & \scriptsize{Video} & \xmark & \xmark & \cmark &\xmark & \xmark & \scriptsize{300}& \scriptsize{42K}  & \scriptsize{Surveillance} & \scriptsize{-} \\ \hline

        \scriptsize{MRP~\cite{layne2014investigating}} & \scriptsize{UAV} & \scriptsize{Video} & \cmark & \cmark & \cmark &\xmark & \xmark & \scriptsize{28} & \scriptsize{4K} & \scriptsize{Surveillance} & \scriptsize{$<10$ } \\ \hline

        \scriptsize{PRAI-1581~\cite{Wang2014}} & \scriptsize{UAV} & \scriptsize{Still} & \xmark & \xmark & \cmark &\xmark & \xmark & \scriptsize{1,581}& \scriptsize{39K}  & \scriptsize{Surveillance} & \scriptsize{[20, 60]} \\ \hline

        \scriptsize{CSM~\cite{Ahmed2015}} & \scriptsize{(Various)} & \scriptsize{Video} & \xmark & \xmark & \xmark &\cmark & \xmark & \scriptsize{1,218} & \scriptsize{11M} & \scriptsize{TV} & \scriptsize{-} \\ \hline
        
        \scriptsize{Market1501~\cite{zheng2015scalable}} & \scriptsize{CCTV} & \scriptsize{Still} & \cmark & \cmark & \cmark &\xmark & \xmark & \scriptsize{1,501} & \scriptsize{32,668} & \scriptsize{Surveillance} & \scriptsize{$<10$ } \\ \hline

        \scriptsize{Mini-drone~\cite{Bonetto2015}} & \scriptsize{UAV} & \scriptsize{Videos} & \cmark & \cmark & \xmark &\xmark & \cmark & \scriptsize{-} & \scriptsize{$>$ 27K} & \scriptsize{Surveillance} & \scriptsize{$<10$ } \\ \hline

        \scriptsize{Mars~\cite{Zheng2016}} & \scriptsize{CCTV} & \scriptsize{Video} & \xmark & \xmark & \cmark &\xmark & \xmark & \scriptsize{1,261}& \scriptsize{20K}  & \scriptsize{Surveillance} & \scriptsize{-} \\ \hline

        \scriptsize{AVI~\cite{Singh2018}} & \scriptsize{UAV} & \scriptsize{Still} & \xmark & \xmark & \xmark &\xmark & \cmark & \scriptsize{5,124} & \scriptsize{10K}  & \scriptsize{Surveillance} & \scriptsize{$[2, 8]$ } \\ \hline

        \scriptsize{DukeMTMC-VideoReID~\cite{Wu2018}} & \scriptsize{CCTV} & \scriptsize{Video} & \xmark & \xmark & \cmark &\xmark & \xmark & \scriptsize{1,812}& \scriptsize{815K}  & \scriptsize{Surveillance} & \scriptsize{-} \\ \hline

        \scriptsize{iQIYI-VID~\cite{Liu2019}} & \scriptsize{(Various)} & \scriptsize{Video} & \xmark & \xmark & \xmark &\cmark & \xmark & \scriptsize{5,000} & \scriptsize{600K} & \scriptsize{TV} & \scriptsize{-} \\ \hline
        
        \scriptsize{DRone HIT~\cite{Grigorev2019}} & \scriptsize{UAV} & \scriptsize{Still} & \xmark & \xmark & \cmark &\xmark & \xmark & \scriptsize{101}& \scriptsize{40K}  & \scriptsize{Surveillance} & \scriptsize{25} \\ \hline

    \cellcolor{gray!50}\scriptsize{P-DESTRE} &     \cellcolor{gray!50}\scriptsize{UAV} &    \cellcolor{gray!50}\scriptsize{Video} &     \cellcolor{gray!50}\cmark &     \cellcolor{gray!50}\cmark &    \cellcolor{gray!50}\cmark &    \cellcolor{gray!50}\cmark &     \cellcolor{gray!50}\cmark &     \cellcolor{gray!50}\scriptsize{253} &     \cellcolor{gray!50}\scriptsize{$>$ 14.8M} &     \cellcolor{gray!50}\scriptsize{Surveillance} &     \cellcolor{gray!50}\scriptsize{$[5.5, 6.7]$} \\ \hline

    \end{tabular}
    \label{tab:related}
\end{table*}

As summarized in Table~\ref{tab:related}, various datasets for supporting pedestrian analysis research have been released in the past. The pioneer initiative was the 'PRID-2011'~\cite{Hirzer2011}, containing 400 image sequences of 200 pedestrians. 'CUHK03'~\cite{Li2014} aimed at providing enough data for deep learning-based solutions, and contains images collected from 5 cameras, comprising 1,467 identities and 13,164 bounding boxes. The 'iLIDS-VID'~\cite{Wang2014} set was the first to release video data, comprising 600 sequences of 300 individuals,  with sequences length ranging from 23 to 192 frames. The 'MRP'~\cite{layne2014investigating} was the first attempt to actually provide an UAV-based dataset specifically for the re-identification problem, containing a relatively short number of identities (28) and 4,000 bounding boxes.  Released at roughly the same time, the 'PRAI-1581'~\cite{Wang2014} reproduces undoubtedly real surveillance conditions, but UAVs flew at too high altitude to enable re-identification experiments (up to 60 meters). This set has 39,461 images of 1,581 identities, and is mainly used for detection and tracking purposes. The 'Market-1501'~\cite{zheng2015scalable} was collected using 6 cameras in front of a supermarket, and contains 32,668 bounding boxes of 1,501 identities. Its extension ('MARS'~\cite{Zheng2016}) was the first video-based set specifically devoted to pedestrian re-identification. Singularly, the 'Mini-drone'~\cite{Bonetto2015} set was created mostly to support abnormal event detection analysis, and can also be used for pedestrian detection, tracking and re-identification purposes (but not search). 

Subsequently, the 'DukeMTMC-VideoReID'~\cite{Wu2018} has exclusively pedestrian re-identification purposes and - as a pioneer feature - it also defines a performance evaluation protocol, enumerating the 702 identities used for training, the 702 identities for testing, and the 408 identities that act as distractors. In total, this set comprises 369,656 frames of 2,196 sequences for training and 445,764 frames of 2,636 sequences for testing. The discriminating feature of the 'AVI'~\cite{Singh2018} set, is the support of pose estimation/abnormal event detection experiments, with humans in each frame annotated with 14 body keypoints.  Even more recently, the 'DRoneHIT'~\cite{Grigorev2019} set also supports image-based pedestrian re-identification experiments from aerial data, containing 101 identities, each one with about 459 images. 

Finally, the 'CSM'~\cite{Ahmed2015} and 'iQIQI-VID'~\cite{Liu2019}  sets were included in this summary because they were the unique cases that previously released data regarding the person search problem in particular. However, their video sequences have notoriously different features from surveillance environments and predominantly regard TV shows and movies, where the identities are famous celebrities.  

Among the analyzed datasets, note that the Market1501, MARS, CUHK03, iLIDS-VID and DukeMTMC-VideoReID were collected using stationary cameras, and their data have notoriously different features of the resulting from UAV-based acquisition. Also, even though the PRAI-1581 and DRone HIT sets were collected using UAVs, they do not provide consistent identity information between acquisition sessions, and cannot be used in person search problem.

 \section{The P-DESTRE Dataset}
 \label{sec:ProposedSet}

 \subsection{Data Acquisition Devices and Protocols}

The P-DESTRE dataset is the result of a joint effort from researchers in two universities: the University of Beira Interior\footnote{\url{http://www.ubi.pt}}  (Portugal) and the JSS Science and Technology University\footnote{\url{https://jssstuniv.in}} (India).  In order to enable the research on pedestrian identification from UAV-based data, a set of DJI$^{\scriptsize{\textregistered}}$ Phantom 4\footnote{\url{https://www.dji.com/pt/phantom-4-pro-v2}} drones controlled by human operators flew over various scenes of both university campi, acquiring data that simulate the everyday conditions in outdoor urban environments.   

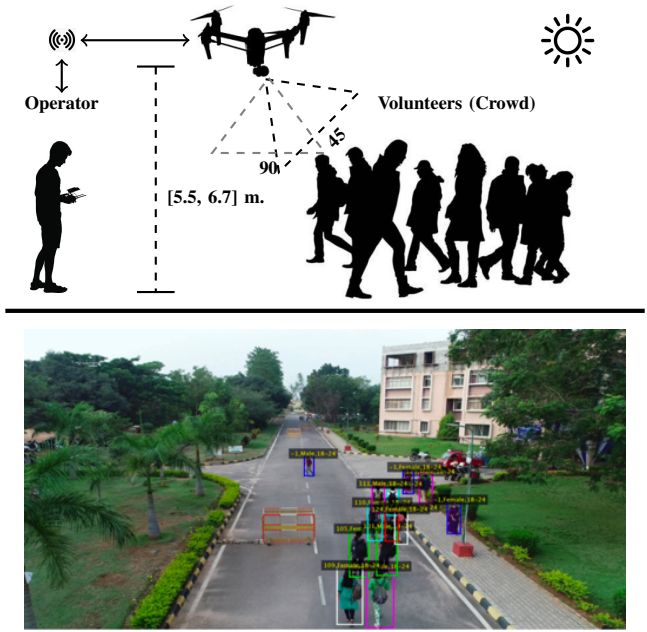
\begin{figure}[ht!]
\begin{center}
\begin{tikzpicture}

\draw (-3.1,-0.4) node(n1)  {\includegraphics[height=1.0 cm]{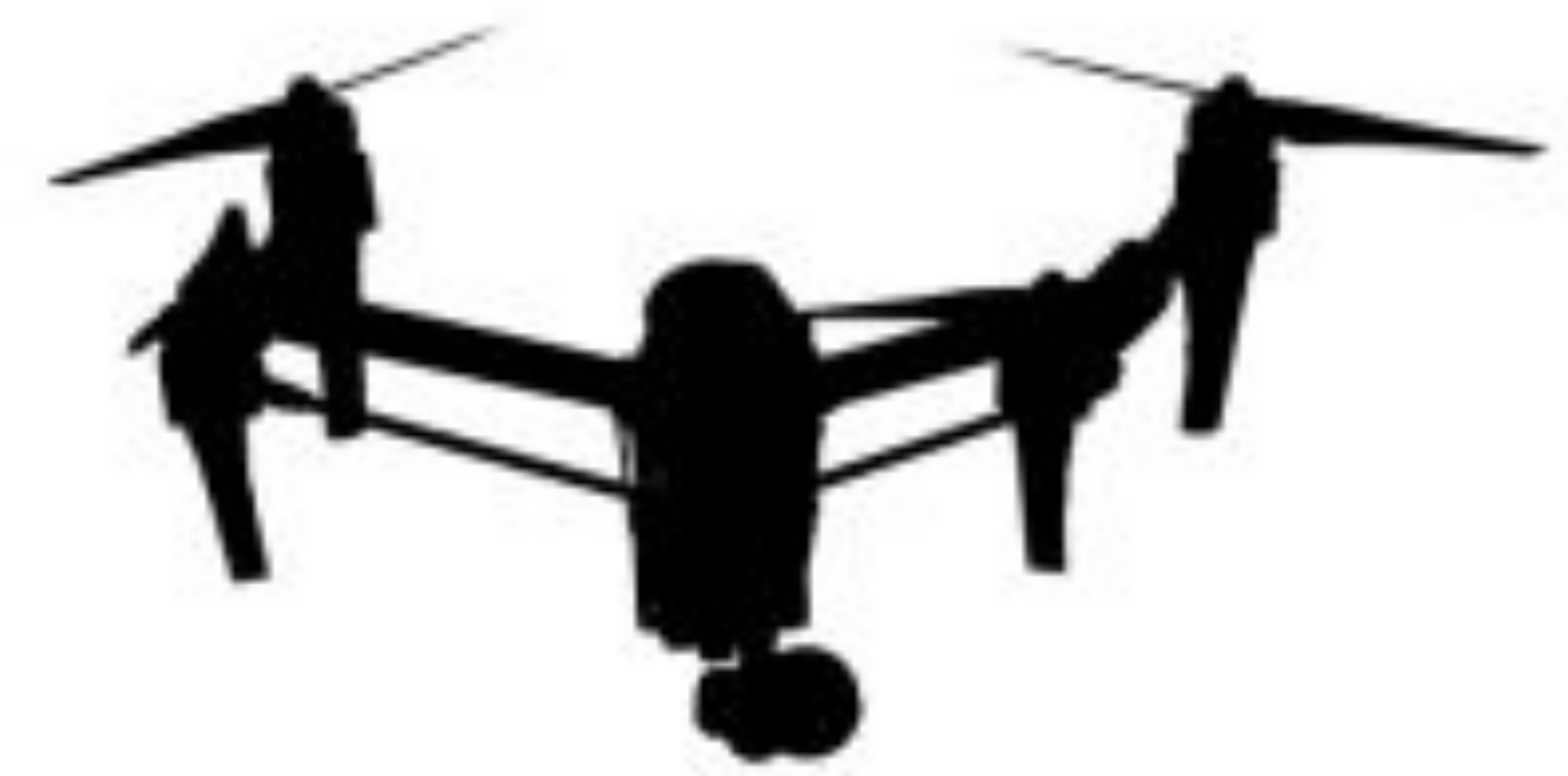}};       
\draw (-0.75,-2.75) node(n1)  {\includegraphics[height=2.25 cm]{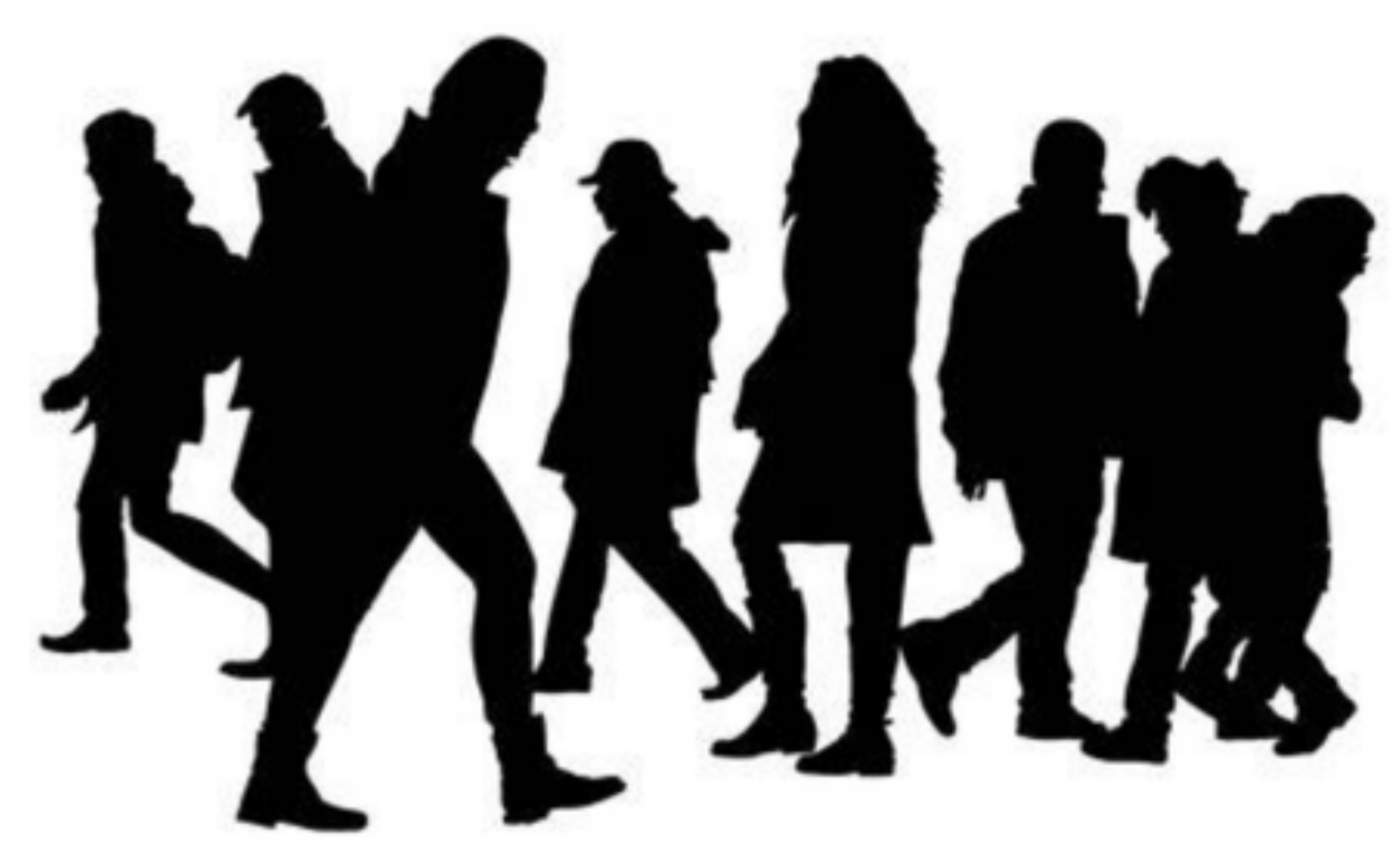}};       

\draw (-5.75,-2.75) node(n1)  {\includegraphics[height=2.0 cm]{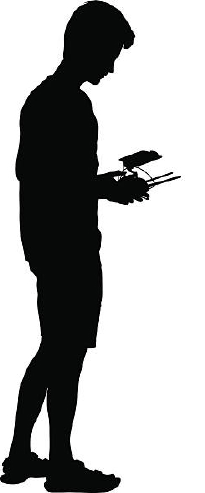}};       

\draw (1.0,-0.4) node(n1)  {\includegraphics[height=0.75 cm]{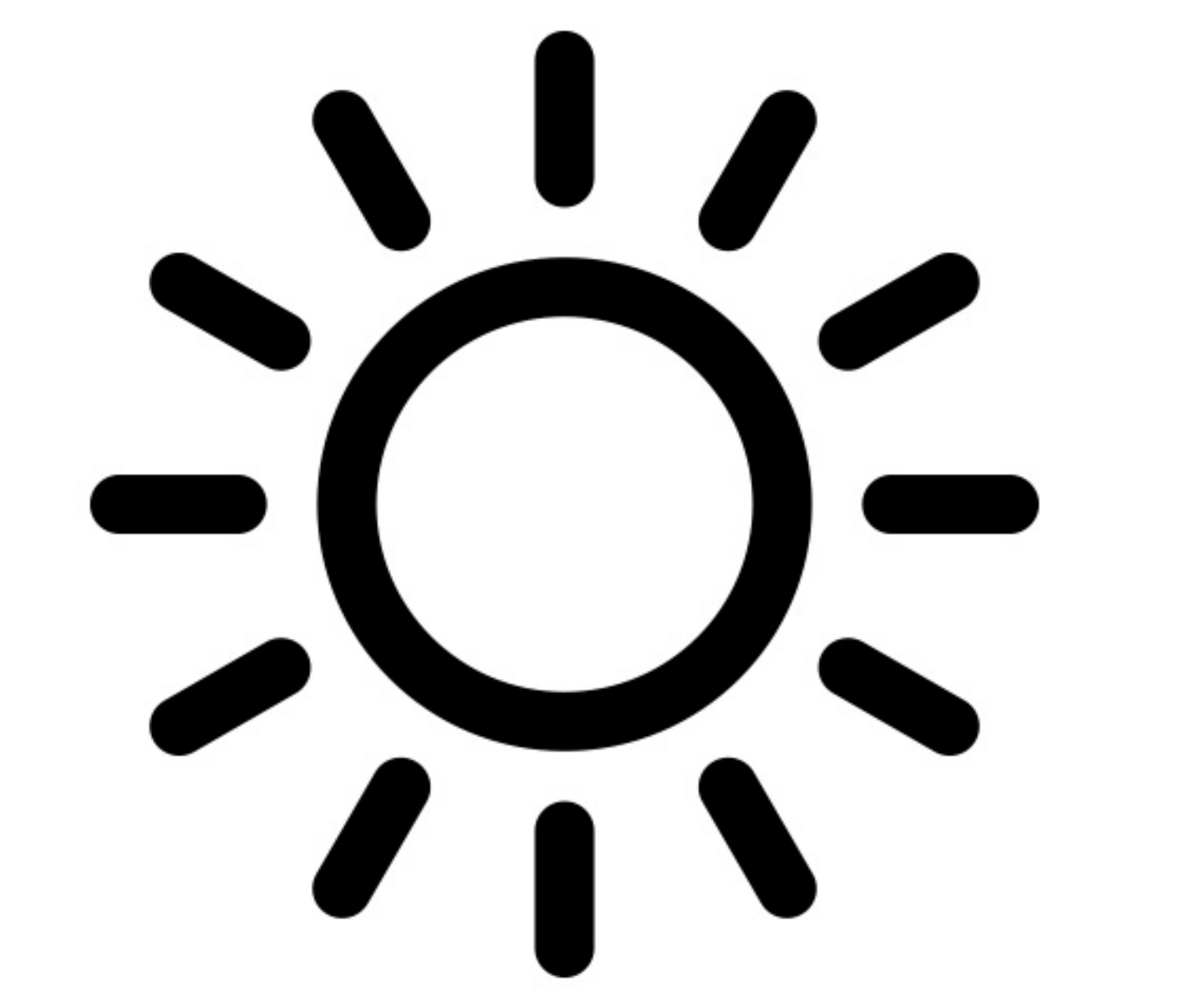}};       

\draw (-5.75,-0.4) node(n1)  {\includegraphics[height=0.3 cm]{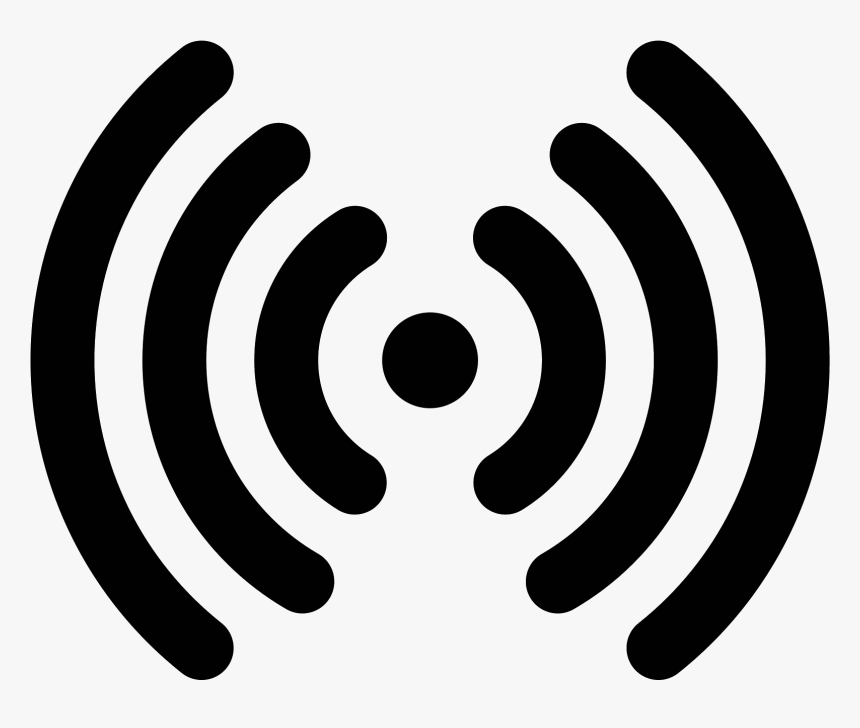}};       

\def\deltaX{-0.6-2.2}
\def\deltaY{0.3+2.2}
\draw [dashed, thick, rotate=45] (0+\deltaX, -1+\deltaY) -- (0.75+\deltaX, -2.0+\deltaY) -- (-0.75+\deltaX, -2.0+\deltaY) -- (0+\deltaX,-1+\deltaY);

\def\deltaX{0.0-3}
\def\deltaY{0.1}
\draw [dashed, thick, gray, rotate=0] (0+\deltaX, -1+\deltaY) -- (0.75+\deltaX, -2.0+\deltaY) -- (-0.75+\deltaX, -2.0+\deltaY) -- (0+\deltaX,-1+\deltaY);

\draw (0.9-3.0, -1.7-0) node[rectangle, rotate=45] {\scriptsize{\textbf{45º}}};  

\draw (0.0-3, -2.1) node[rectangle] {\scriptsize{\textbf{90º}}};  

\draw [ultra thick] (-6.5,-4) -- (2, -4);  

\def\deltaX{1.25}
\draw [thick, dashed] (-5.75+\deltaX, -0.75) -- (-5.75+\deltaX, -3.75);
\draw [thick] (-6.0+\deltaX, -0.75) -- (-5.5+\deltaX, -0.75);
\draw [thick] (-6.0+\deltaX, -3.75) -- (-5.5+\deltaX, -3.75);
\draw (-4.95+\deltaX, -2.5) node[rectangle] {\scriptsize{\textbf{[5.5, 6.7] m.}}};  

\draw (-5.75, -1.25) node[rectangle] {\scriptsize{\textbf{Operator}}};  

\draw (-0.5, -1.25) node[rectangle] {\scriptsize{\textbf{Volunteers (Crowd)}}};  

\draw [thick, <->] (-5.75, -1.1) -- (-5.75, -0.65);

\draw [thick, <->] (-5.5, -0.4) -- (-4.05, -0.4);

\draw (-2.25,-6.25) node(n1)  {\includegraphics[width=8.0 cm]{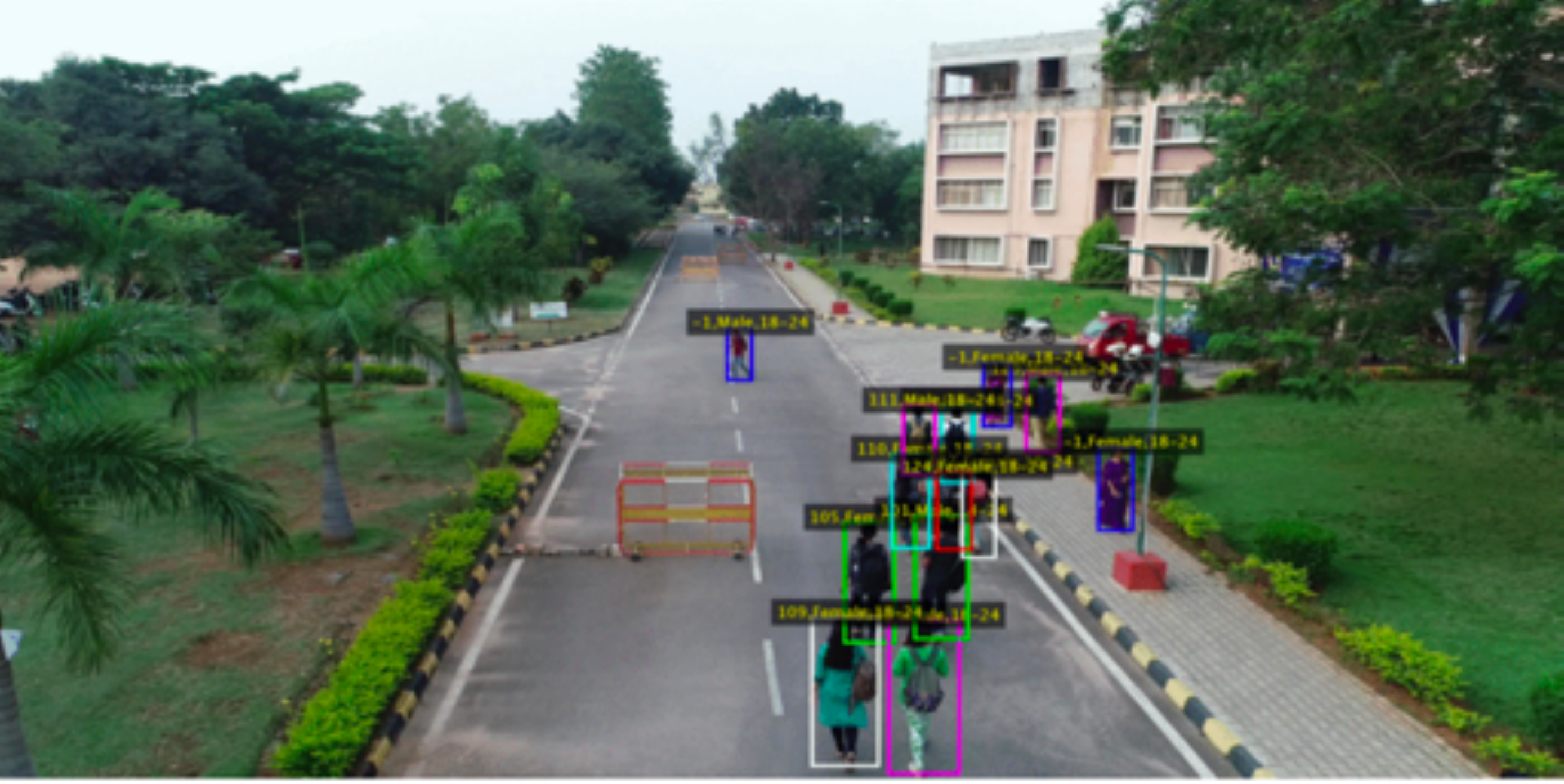}};    

\end{tikzpicture}
    \caption{At top: schema of the data acquisition protocol used in the P-DESTRE dataset. Human operators controlled DJI Phantom 4 aircrafts in various scenes of two university \emph{campi}, flying at altitudes between 5.5- and 6.7-meters, with gimbal pitch angles between 45$^{\circ}$ to 90$^{\circ}$. The image at the bottom provides one example of a full scene of the P-DESTRE set.}
        \label{fig:data_aquisition}
    \end{center}
\end{figure}

All subjects in the dataset offered explicitly as volunteers and they were asked to completely ignore the UAVs (Fig.~\ref{fig:data_aquisition}), that were flying at altitudes between 5.5 and 6.7 meters, with the camera pitch angles varying between 45$^{\circ}$ to 90$^{\circ}$. Volunteers were students of both universities (in the 18-24 age interval, $>$ 90\%), $\approx$ 65/35\% males/females, and of predominantly two ethnicities ('white' and 'indian'). About 28\% of the volunteers were using glasses, 10\% of them were using sunglasses. Data were recorded at 30fps, with 4K spatial resolution ($3,840 \times 2,160$), and stored in "mp4" format, with H.264 compression. The key features of the data acquisition settings are summarized in Table~\ref{tab:summary}, and additional details can be found at the corresponding webpage\footnote{\url{http://p-destre.di.ubi.pt/download.html}}.

\begin{table}[h!]
\centering
     \caption{The P-DESTRE data acquisition main features.}
     \label{tab:summary}
\begin{tabular}{|p{4cm}|p{4cm}|}
\hline
\multicolumn{2}{|l|}{\textbf{\scriptsize{Image Acquisition Settings}}} \\ \hline
\scriptsize{Camera: 1/2.3Ó CMOS, Effective pixels:12.4 M}  & \scriptsize{Frame Size: 3,840 $\times$ 2,160}  \\ \hline
\scriptsize{Lens 0 FOV 94¡ 20 mm (35 mm format equivalent) f/2.8 focus at $\infty$}  & \scriptsize{ISO Range: 100-3200}  \\ \hline
\scriptsize{Camera Pitch Angle: [45$^{\circ}$, 90$^{\circ}$]}  & \scriptsize{Drone Altitude: [5.5, 6.7] meters}  \\ \hline
\scriptsize{Format: MP4, 30 fps}  & \scriptsize{Bit Depth: 24 bit}  \\ \hline
\multicolumn{2}{|l|}{\textbf{\scriptsize{Volunteers}}} \\ \hline
\scriptsize{Total IDs: 269}  & \scriptsize{Gender: Male: 175 (65\%); Female: 94 (35\%)}  \\ \hline
\end{tabular}
\end{table}

\subsection{Annotation Data}

The P-DESTRE dataset is fully annotated at the frame level, by human experts. We provide one text file for each video, using the same file naming protocol (plus the ".txt" extension). The annotation process was divided into three phases: 1) human detection; 2) tracking; and 3) identification and soft biometrics characterisation. 

At first, the well-known Mask R-CNN~\cite{He2017} method was used to provide an initial estimate of the position of every pedestrian in the scene, with the resulting data subjected to human verification and correction. Next, the deep sort method~\cite{Wojke2017} provided the preliminary tracking information, which again was corrected manually. As result of these two initial steps, we obtained the rectangular bounding boxes providing the regions-of-interest (ROI) of every pedestrian in each frame/video. The final phase of the annotation process was carried out manually, with human annotators that  knew personally the volunteers of each university setting the ID information and characterising the samples according to the soft labels.  

Table~\ref{tab:annotation} provides the details of the labels annotated for every instance (pedestrian/frame) in the dataset, along with the ID information, the bounding box that defines the ROI and the frame information. For every label, we also provide a list of its possible values. 

\begin{table}[h!]
\centering
     \caption{The P-DESTRE dataset annotation protocol. For each video, a text file provides the annotation at frame level, with the ROI of each pedestrian in the scene, together with the ID information and 16 other soft biometric labels}
     \label{tab:annotation}
\begin{tabular}{|c|p{5.75cm}|}
\hline
\textbf{\scriptsize{Attributes}}  & \textbf{\scriptsize{Values}}  \\ \hline
\scriptsize{Frame  }              & \scriptsize{\textbf{1}, \textbf{2}, \ldots}                                                                                                                                                                                  \\ \hline
\scriptsize{ID          }         & \scriptsize{\textbf{-1}: 'Unknown', \textbf{1}, \textbf{2}, \ldots }   \\ \hline
\scriptsize{Bounding Box}         & \scriptsize{$[\bm{x}, \bm{y}, \bm{h}, \bm{w}]$ (Top left column, top left row, height, width)  }  \\ \hline
\scriptsize{Age              }    & \scriptsize{\textbf{0}: 0-11, \textbf{1}: 12-17, \textbf{2}: 18-24, \textbf{3}: 25-34, \textbf{4}: 35-44, \textbf{5}: 45-54, \textbf{6}: 55-64, \textbf{7}: $>$ 65, \textbf{8}: 'Unknown' }\\ \hline
\scriptsize{Height           }    & \scriptsize{\textbf{0}: 'Child', \textbf{1}: 'Short', \textbf{2}: 'Medium', \textbf{3}: 'Tall', \textbf{4}: 'Unknown'    }   \\ \hline
\scriptsize{Body Volume} & \scriptsize{\textbf{0}: 'Thin', \textbf{1}: 'Medium', \textbf{2}: 'Fat', \textbf{3}: 'Unknown'}  \\ \hline
\scriptsize{Ethnicity        }    & \scriptsize{\textbf{0}: 'White', \textbf{1}: 'Black', \textbf{2}: 'Asian', \textbf{3}: 'Indian', \textbf{4}: 'Unknown'   }     \\ \hline
\scriptsize{Hair Color       }    & \scriptsize{\textbf{0}: 'Black', \textbf{1}: 'Brown', \textbf{2}: 'White', \textbf{3}: 'Red', \textbf{4}: Gray', \textbf{5}: 'Occluded', \textbf{6}: 'Unknown'  }                                                                                                                                 \\ \hline
\scriptsize{Hairstyle          }  & \scriptsize{\textbf{0}: 'Bald', \textbf{1}: 'Short', \textbf{2}: 'Medium', \textbf{3}: 'Long', \textbf{4}: Horse Tail,' \textbf{5}: 'Unknown'     }                                                                                                                                        \\ \hline
\scriptsize{Beard   }& \scriptsize{\textbf{0}: 'Yes', \textbf{1}: 'No', \textbf{2}: 'Unknown'    }                                                                                                                                                                       \\ \hline
\scriptsize{Moustache  }   &\scriptsize{\textbf{0}: 'Yes', \textbf{1}: 'No', \textbf{2}: 'Unknown'    }                                                                                                                                                                       \\ \hline
\scriptsize{Glasses  }& \scriptsize{\textbf{0}: 'Yes', \textbf{1}: 'Sunglass', \textbf{2}: 'No', \textbf{3}: 'Unknown' }   \\ \hline
\scriptsize{Head Accessories}     & \scriptsize{\textbf{0}: 'Hat', \textbf{1}: 'Scarf', \textbf{2}: 'Neckless', \textbf{3}: 'Occluded', \textbf{4}: 'Unknown'  }   \\ \hline
\scriptsize{Upper Body Clothing } & \scriptsize{\textbf{0}: 'T-shirt', \textbf{1}: 'Blouse', \textbf{2}: 'Sweater', \textbf{3}: 'Coat', \textbf{4}: 'Bikini', \textbf{5}: 'Naked', \textbf{6}: 'Dress', \textbf{7}: 'Uniform', \textbf{8}: 'Shirt', \textbf{9}: 'Suit', \textbf{10}: 'Hoodie', \textbf{11}: 'Cardigan'}  \\ \hline
\scriptsize{Lower Body Clothing } & \scriptsize{\textbf{0}: 'Jeans', \textbf{1}: 'Leggins', \textbf{2}: 'Pants', \textbf{3}: 'Shorts', \textbf{4}: 'Skirt', \textbf{5}: 'Bikini', \textbf{6}: 'Dress', \textbf{7}: 'Uniform', \textbf{8}: 'Suit', \textbf{9}: 'Unknown ' }  \\ \hline
\scriptsize{Feet             }    & \scriptsize{\textbf{0}: 'Sport', \textbf{1}: 'Classic', \textbf{2}: 'High Heels', \textbf{3}: 'Boots', \textbf{4}: 'Sandals, \textbf{5}: 'Nothing', \textbf{6}: Unknown' }    \\ \hline
\scriptsize{Accessories  }        & \scriptsize{\textbf{0}: 'Bag', \textbf{1}: 'Backpack', \textbf{2}: 'Rolling', \textbf{3}: 'Umbrella', \textbf{4}: 'Sportif', \textbf{5}: 'Market', \textbf{6}: 'Nothing', \textbf{7}: 'Unknown' }                                                                                                  \\ \hline
\scriptsize{Action  }   & \scriptsize{\textbf{0}: 'Walk', \textbf{1}: 'Run', \textbf{2}: 'Stand', \textbf{3}: 'Sit', \textbf{4}: 'Cycle', \textbf{5}: 'Exercise', \textbf{6}: 'Pet', \textbf{7}: 'Phone', '\textbf{8}: 'Leave Bag', \textbf{9}: 'Fall', \textbf{10}: 'Fight', \textbf{11}: 'Date', \textbf{12}: 'Offend', \textbf{13}: 'Trade'}\\ \hline
\end{tabular}
\end{table}

\subsection{Typical Data Degradation Factors}

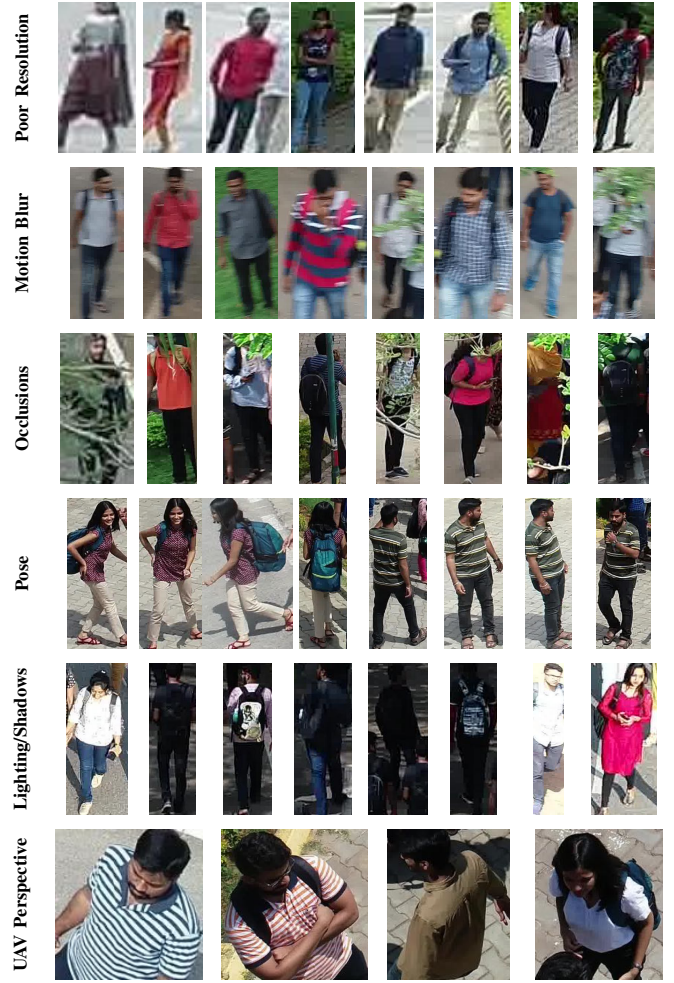
\begin{figure}[ht!]
\begin{center}
\begin{tikzpicture}

\draw (0,0) node(n1)  {\includegraphics[height=2.0 cm]{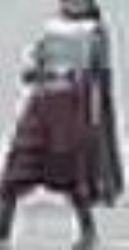}};       
\draw (1.0,0) node(n1)  {\includegraphics[height=2.0 cm]{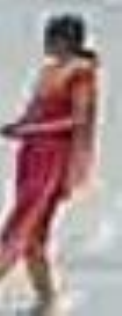}};     
\draw (2.0,0) node(n1)  {\includegraphics[height=2.0 cm]{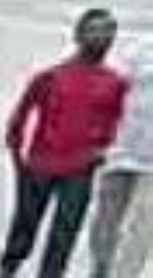}};     
\draw (3.0,0) node(n1)  {\includegraphics[height=2.0 cm]{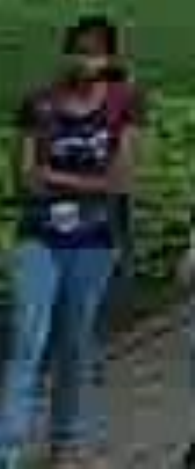}};     
\draw (4.0,0) node(n1)  {\includegraphics[height=2.0 cm]{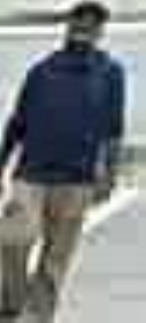}};     
\draw (5.0,0) node(n1)  {\includegraphics[height=2.0 cm]{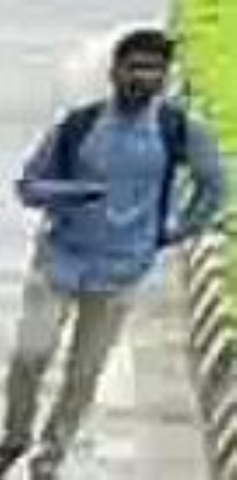}};     
\draw (6.0,0) node(n1)  {\includegraphics[height=2.0 cm]{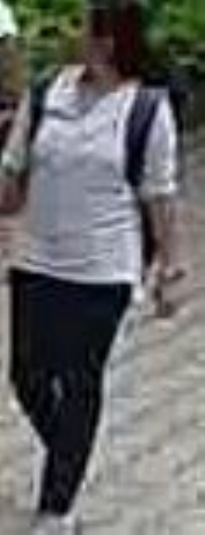}};     
\draw (7.0,0) node(n1)  {\includegraphics[height=2.0 cm]{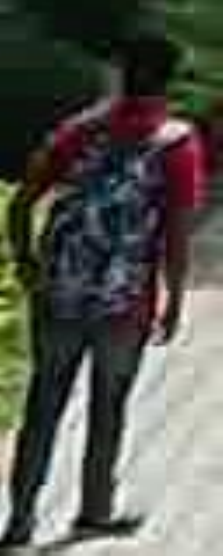}};     
\draw (-1, 0) node[rectangle, rotate=90] {\scriptsize{\textbf{Poor Resolution}}};

\def\deltaY{-2.2}
\draw (0,\deltaY) node(n1)  {\includegraphics[height=2.0 cm]{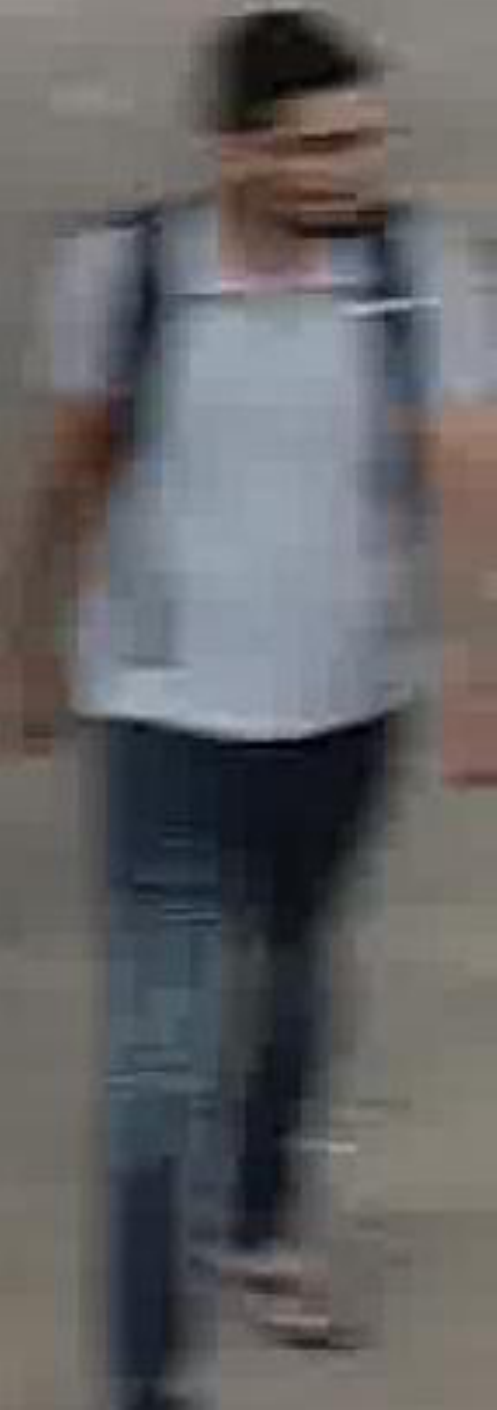}};       
\draw (1.0,\deltaY) node(n1)  {\includegraphics[height=2.0 cm]{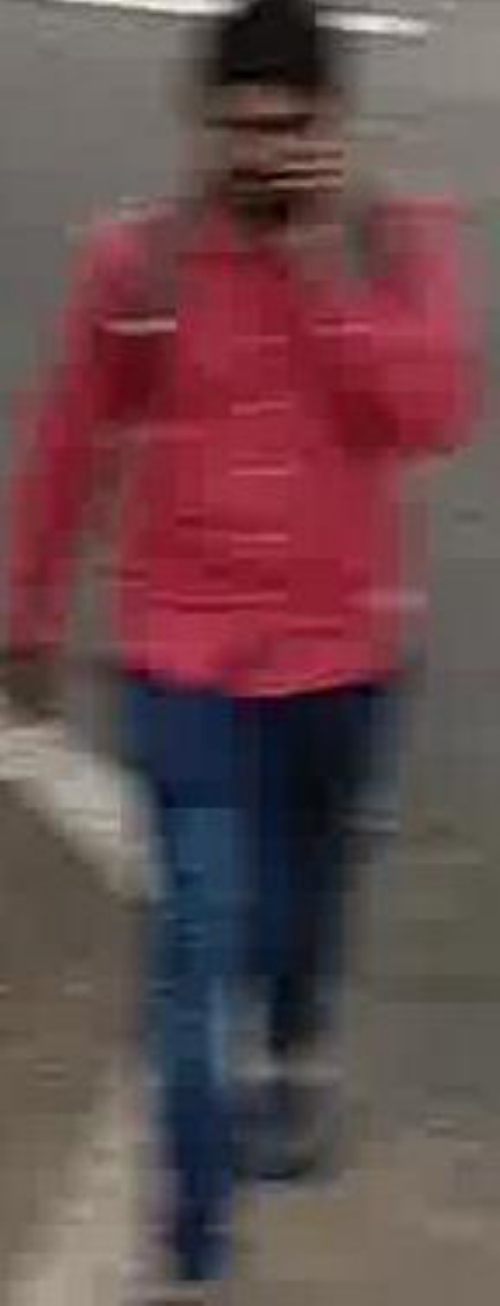}};     
\draw (2.0,\deltaY) node(n1)  {\includegraphics[height=2.0 cm]{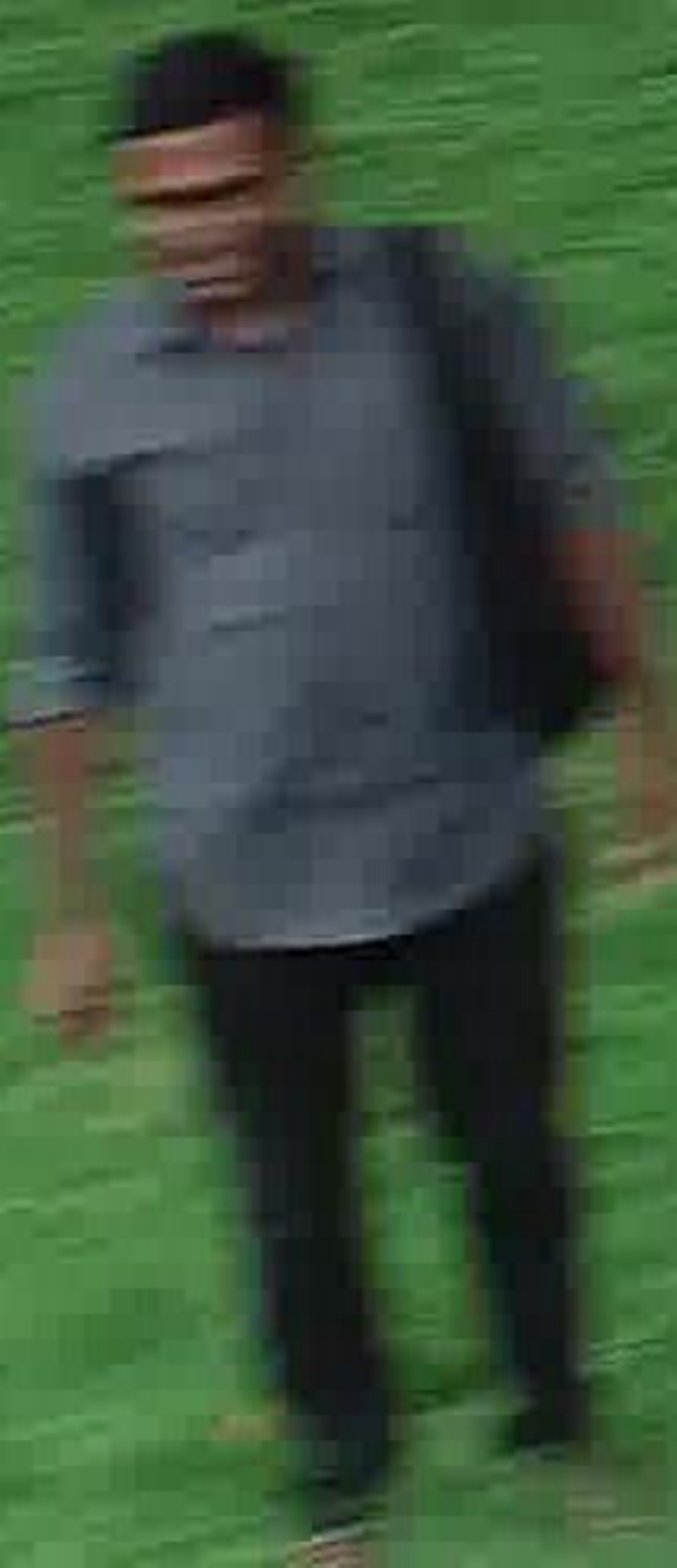}};     
\draw (3.0,\deltaY) node(n1)  {\includegraphics[height=2.0 cm]{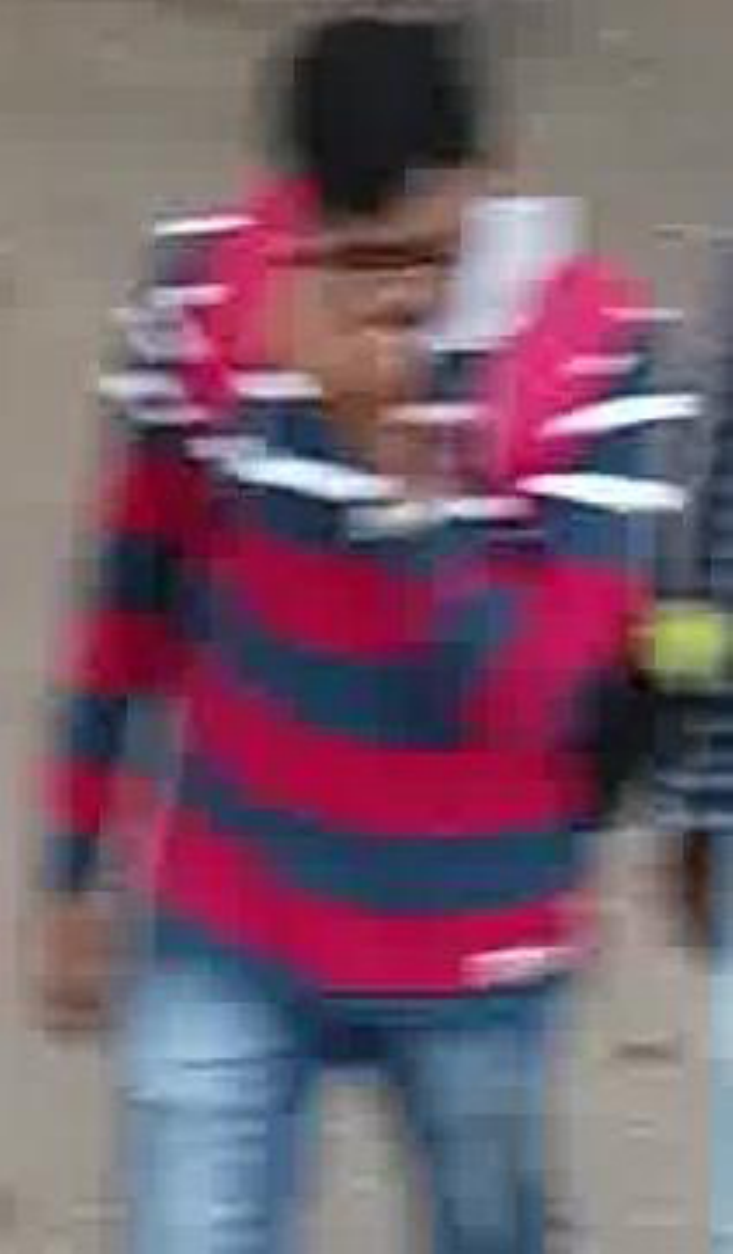}};     
\draw (4.0,\deltaY) node(n1)  {\includegraphics[height=2.0 cm]{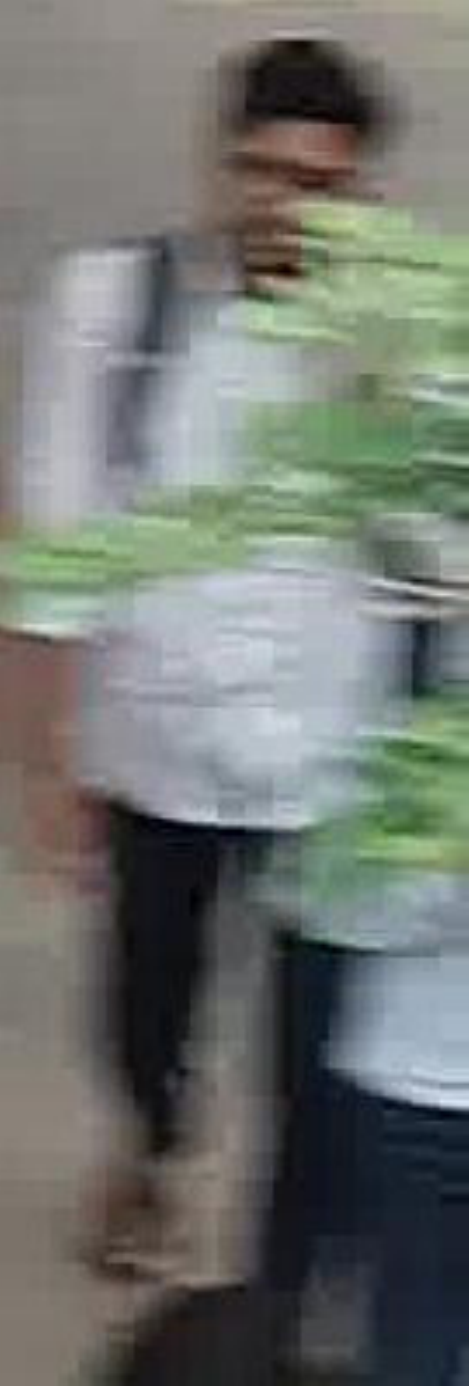}};     
\draw (5.0,\deltaY) node(n1)  {\includegraphics[height=2.0 cm]{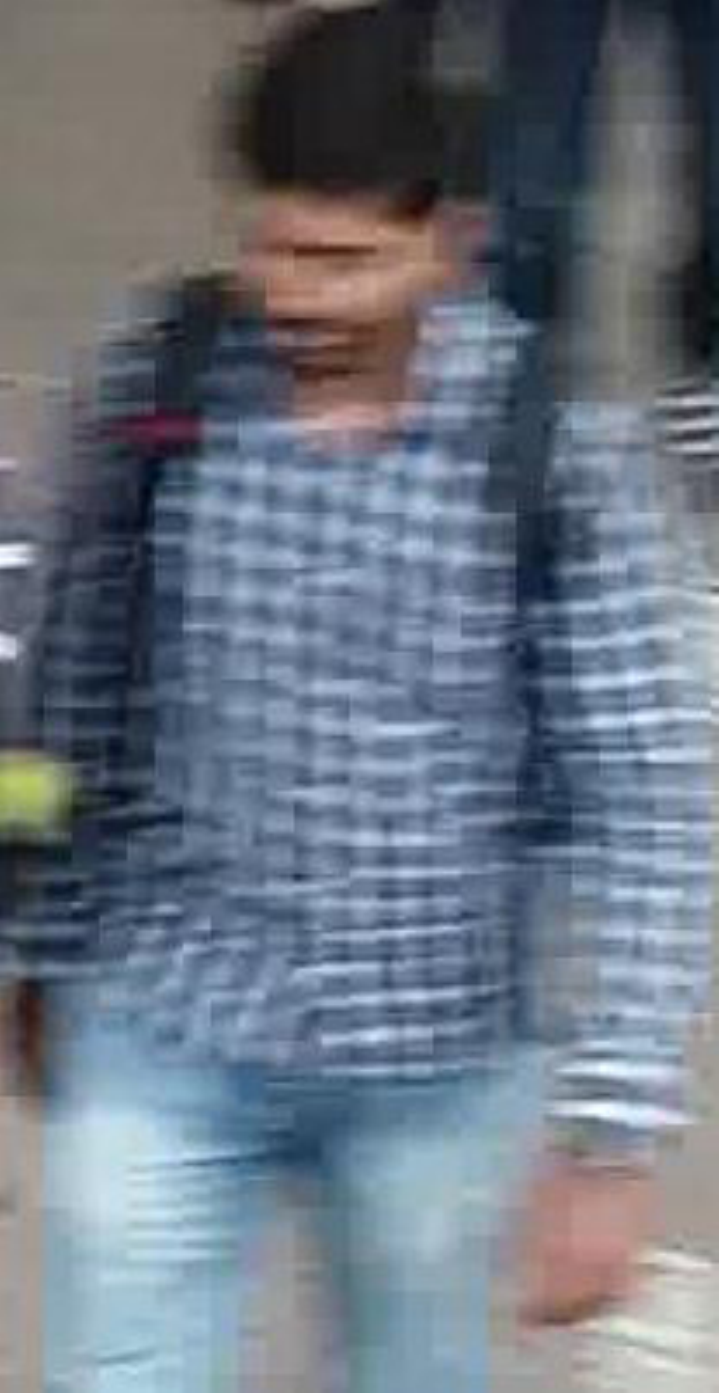}};     
\draw (6.0,\deltaY) node(n1)  {\includegraphics[height=2.0 cm]{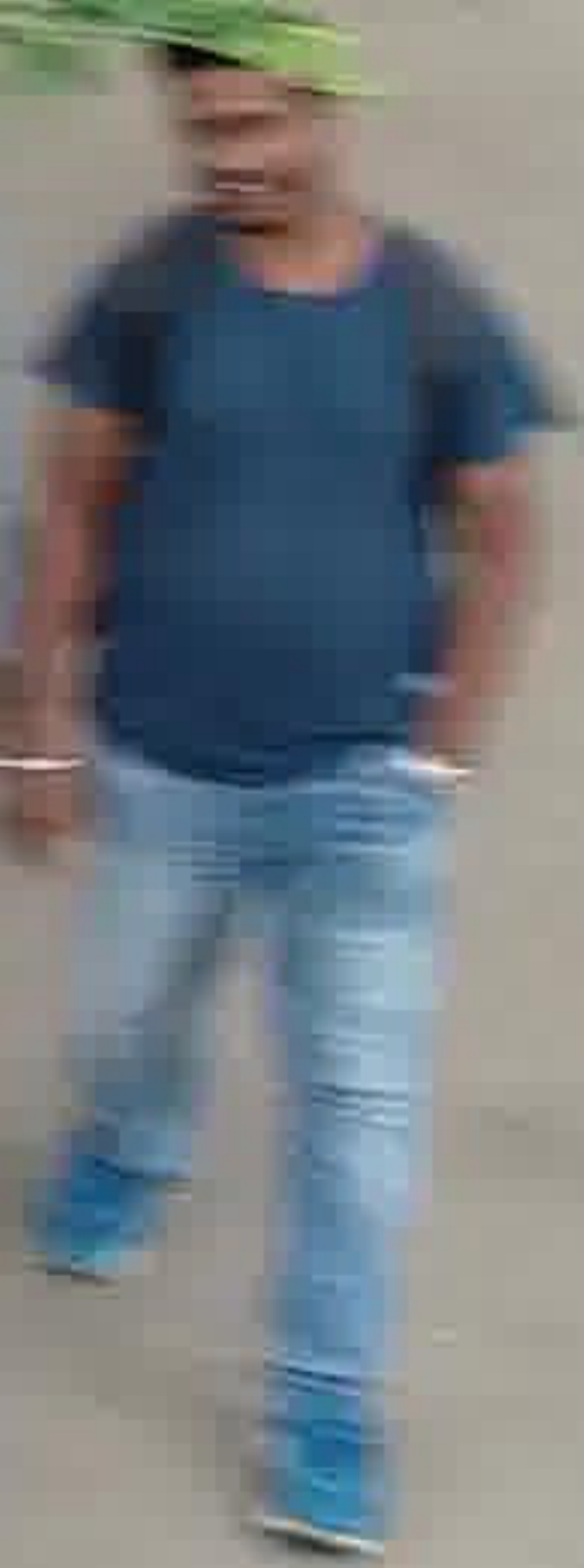}};     
\draw (7.0,\deltaY) node(n1)  {\includegraphics[height=2.0 cm]{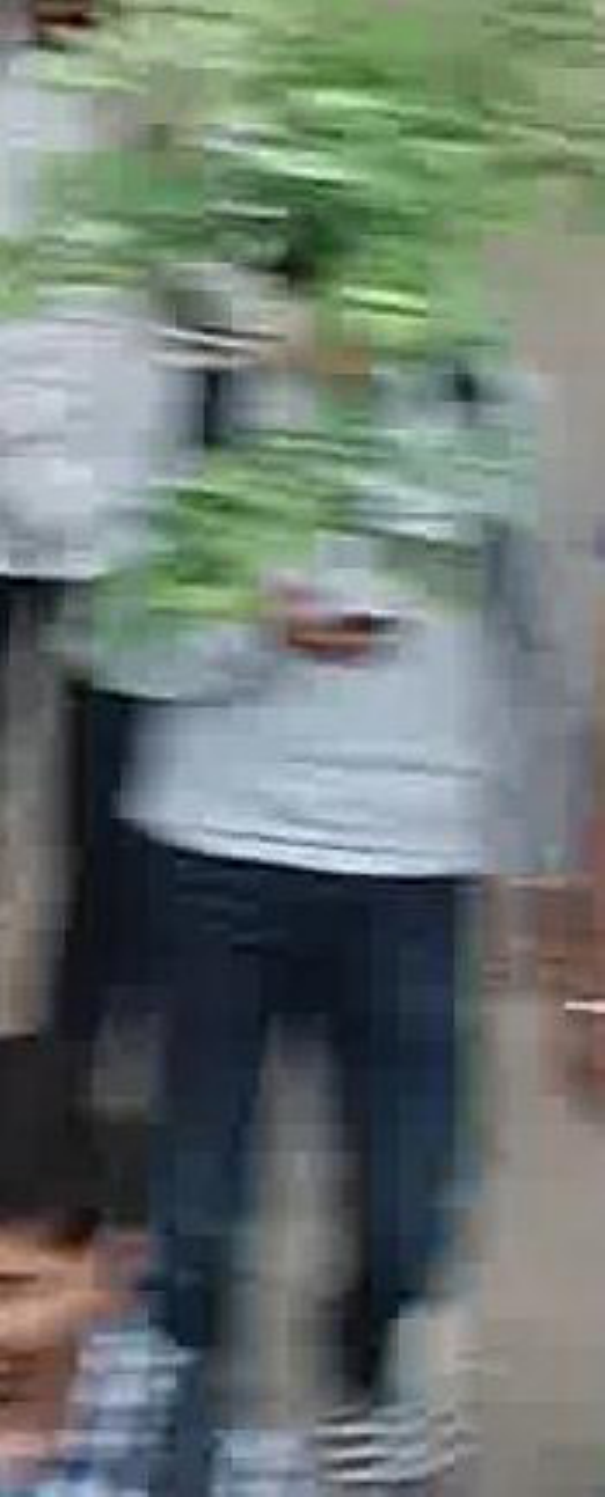}};     
\draw (-1, \deltaY) node[rectangle, rotate=90] {\scriptsize{\textbf{Motion Blur}}};  

\def\deltaY{-4.4}
\draw (0,\deltaY) node(n1)  {\includegraphics[height=2.0 cm]{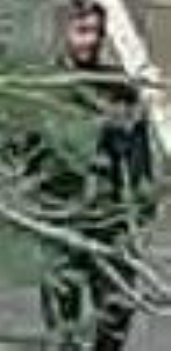}};       
\draw (1.0,\deltaY) node(n1)  {\includegraphics[height=2.0 cm]{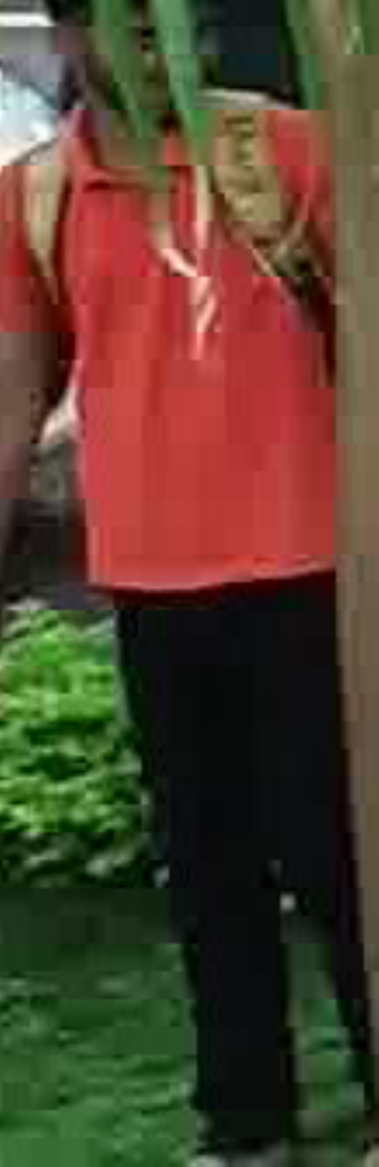}};     
\draw (2.0,\deltaY) node(n1)  {\includegraphics[height=2.0 cm]{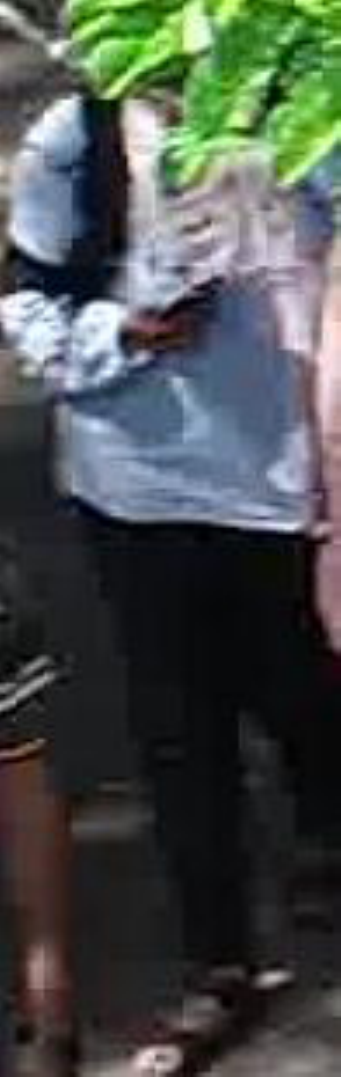}};     
\draw (3.0,\deltaY) node(n1)  {\includegraphics[height=2.0 cm]{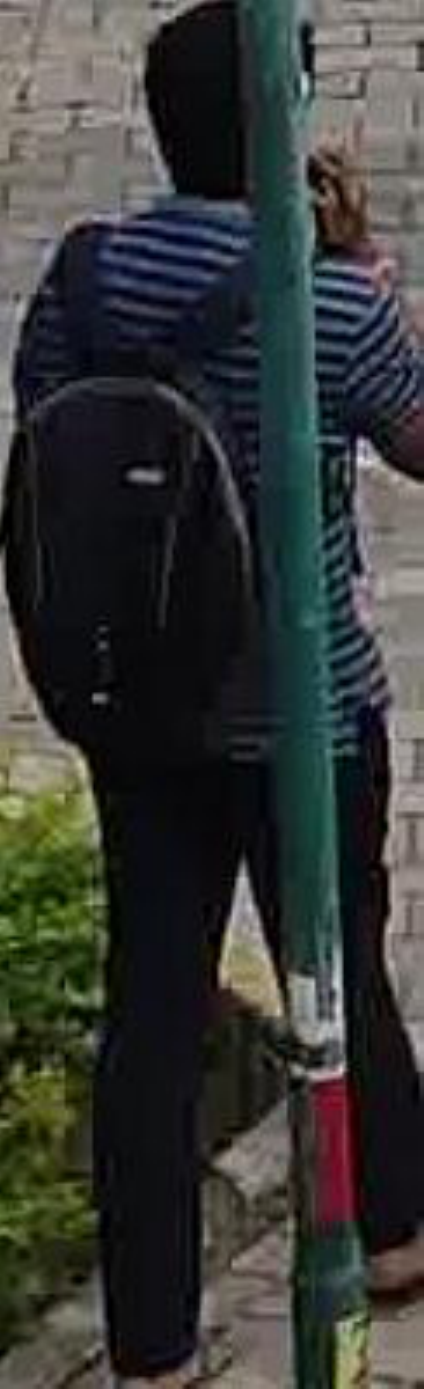}};     
\draw (4.0,\deltaY) node(n1)  {\includegraphics[height=2.0 cm]{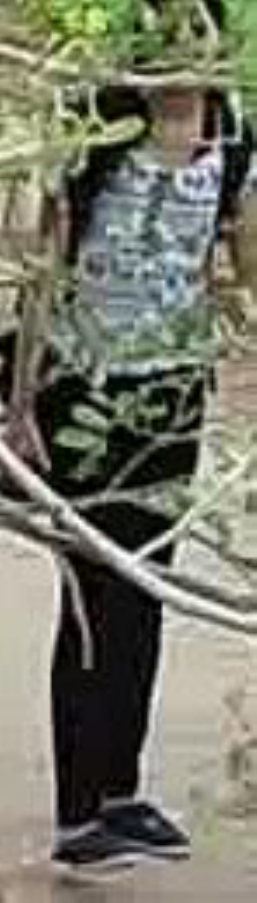}};     
\draw (5.0,\deltaY) node(n1)  {\includegraphics[height=2.0 cm]{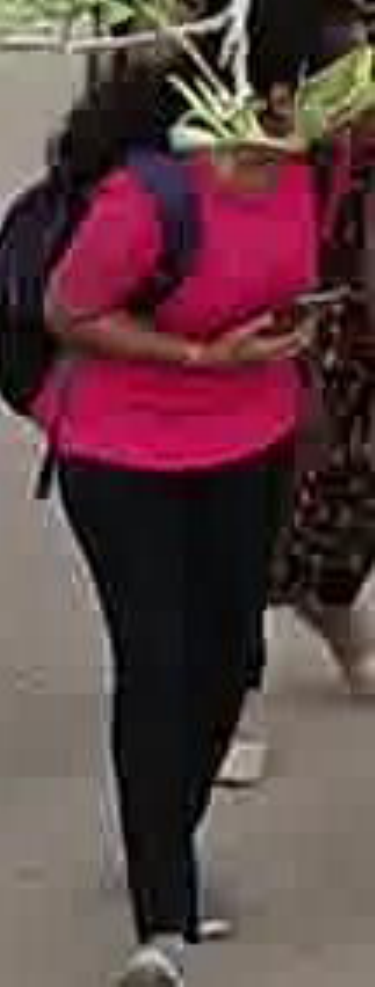}};     
\draw (6.0,\deltaY) node(n1)  {\includegraphics[height=2.0 cm]{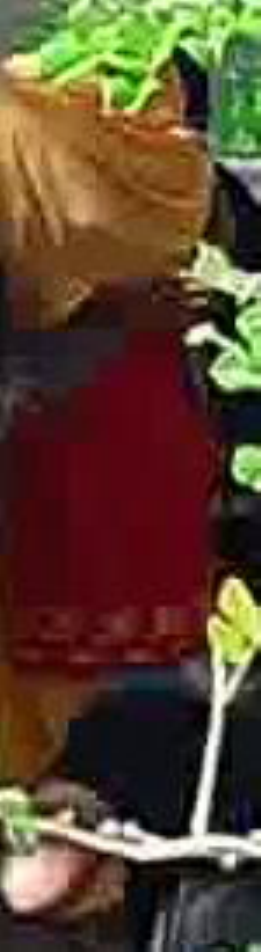}};     
\draw (7.0,\deltaY) node(n1)  {\includegraphics[height=2.0 cm]{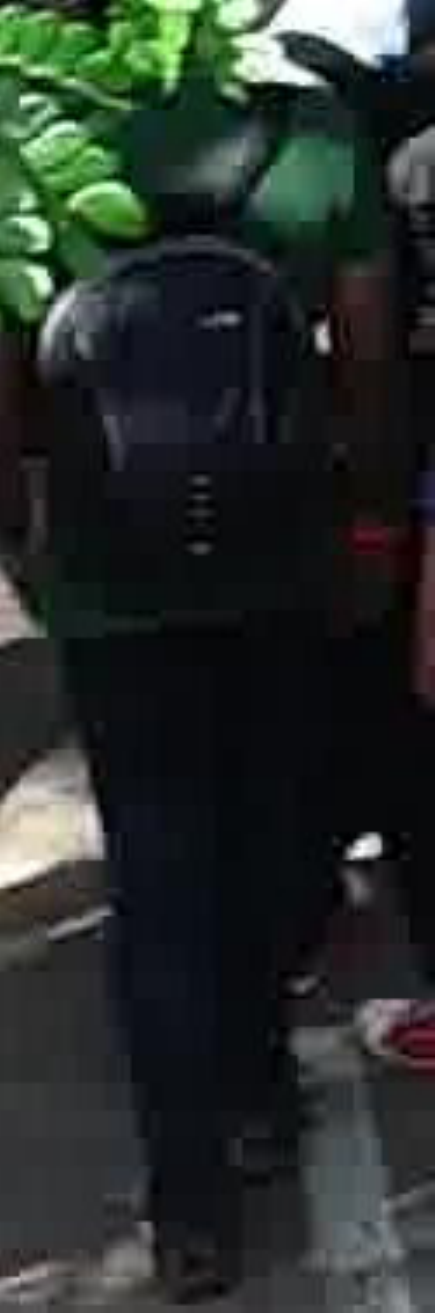}};     
\draw (-1, \deltaY) node[rectangle, rotate=90] {\scriptsize{\textbf{Occlusions}}};  

\def\deltaY{-6.6}
\draw (0,\deltaY) node(n1)  {\includegraphics[height=2.0 cm]{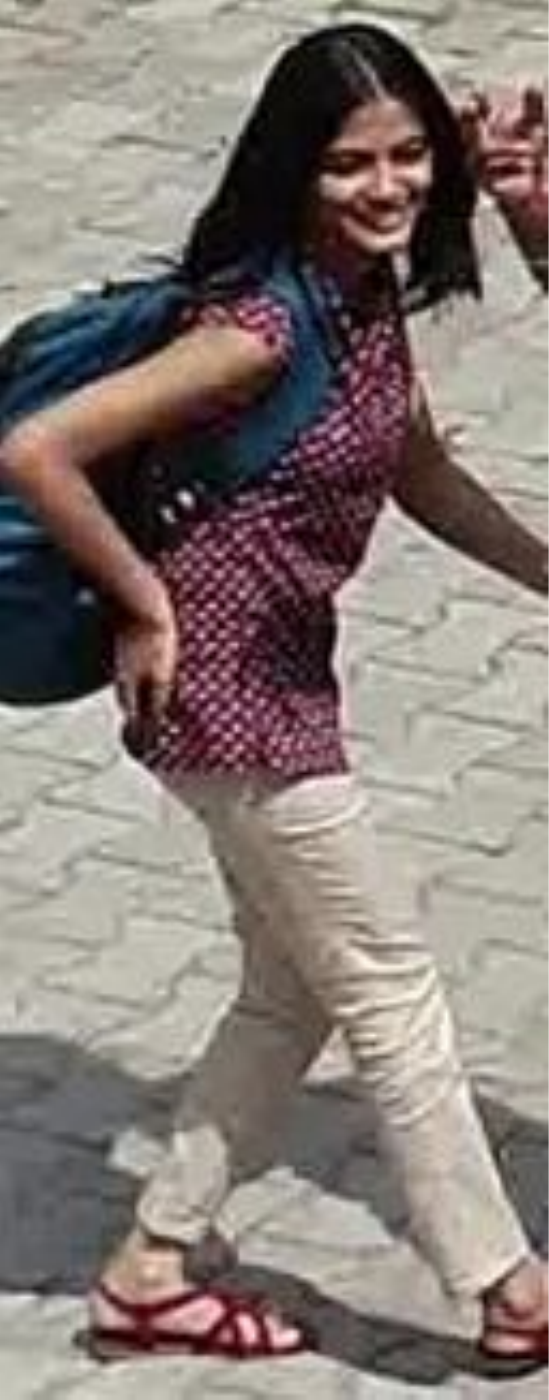}};       
\draw (1.0,\deltaY) node(n1)  {\includegraphics[height=2.0 cm]{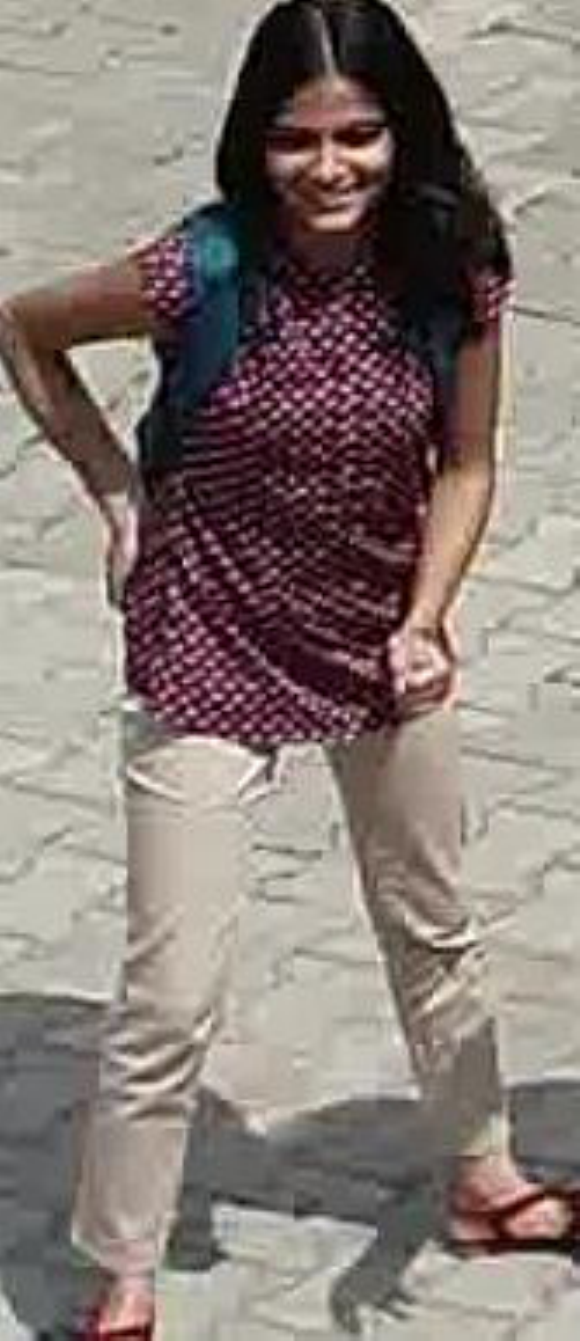}};     
\draw (2.0,\deltaY) node(n1)  {\includegraphics[height=2.0 cm]{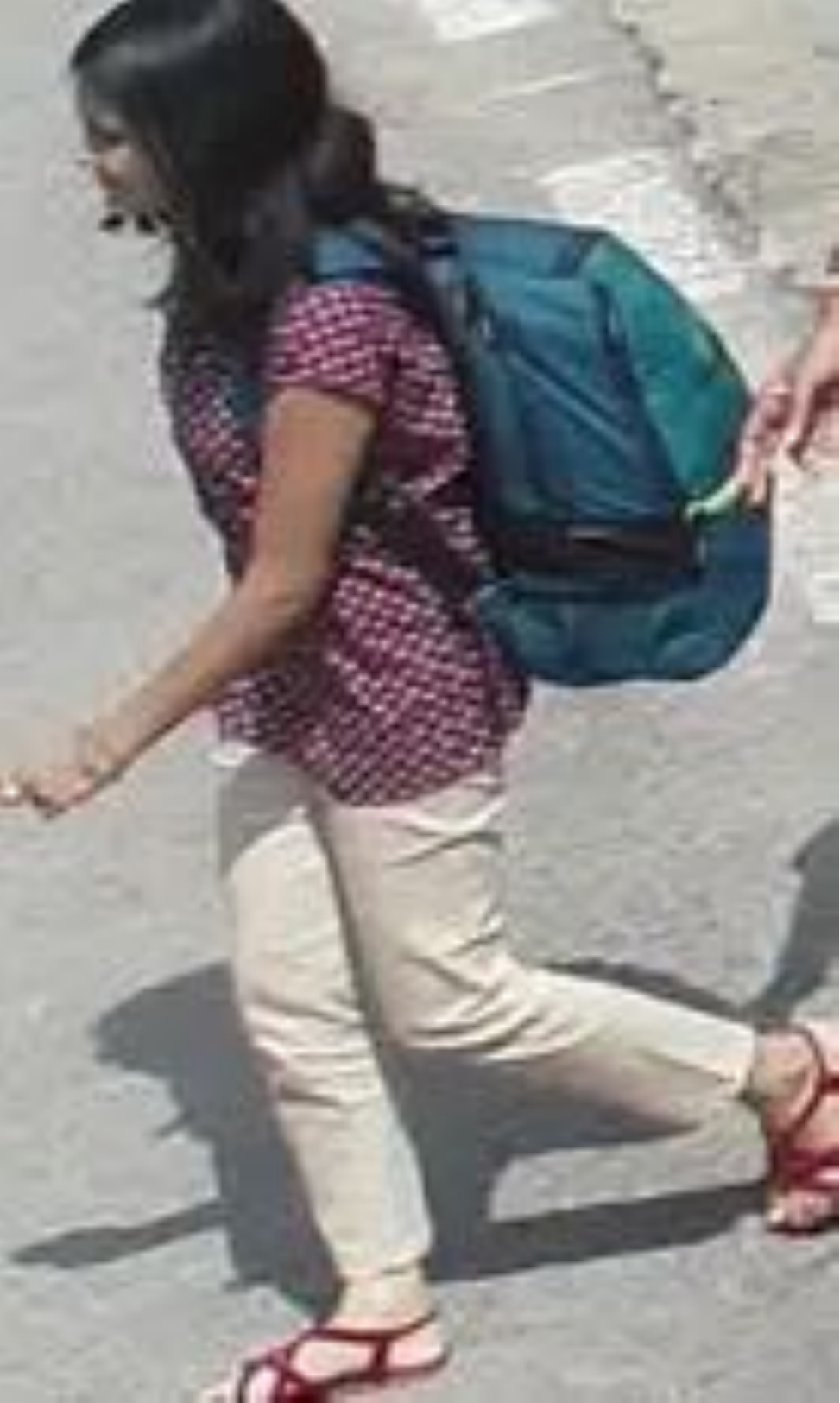}};     
\draw (3.0,\deltaY) node(n1)  {\includegraphics[height=2.0 cm]{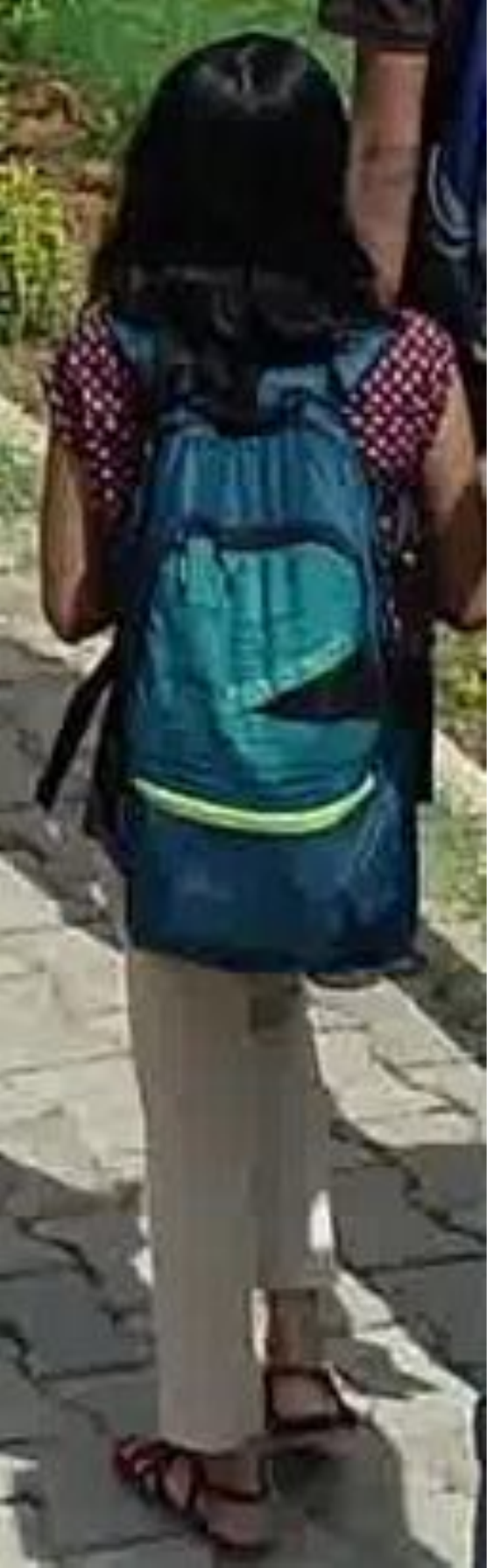}};     
\draw (4.0,\deltaY) node(n1)  {\includegraphics[height=2.0 cm]{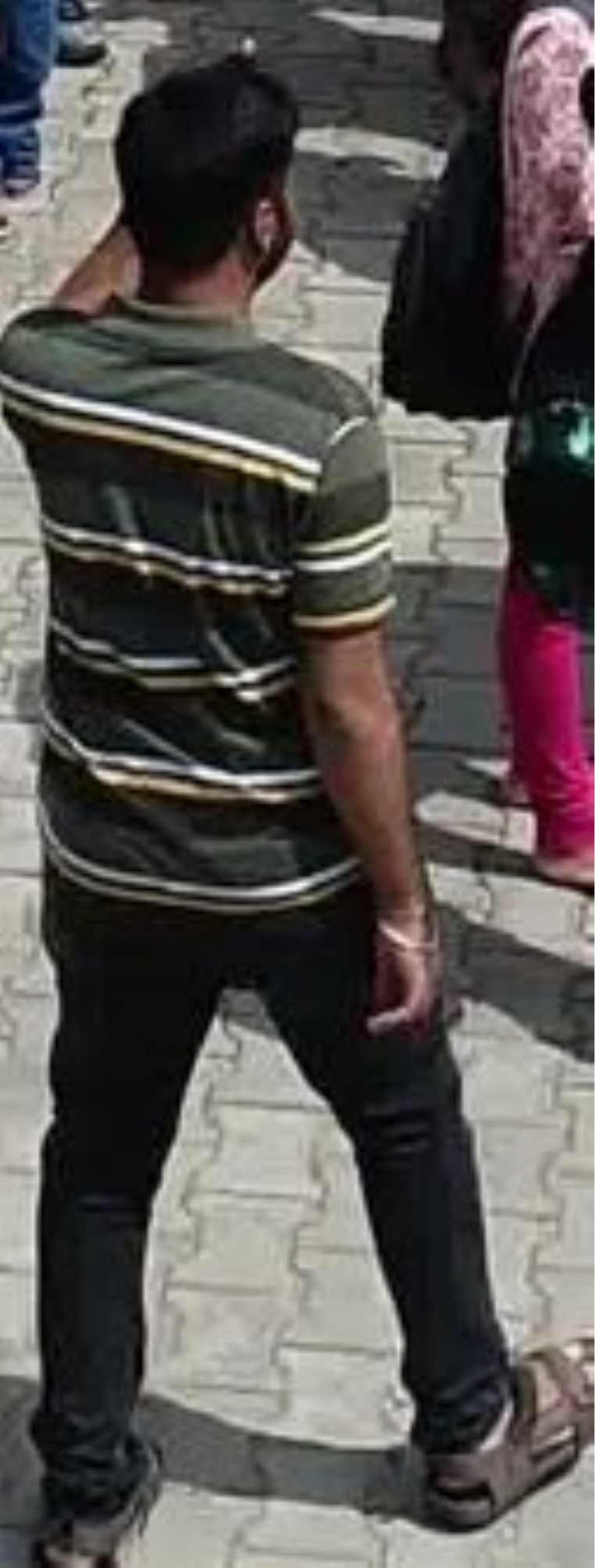}};     
\draw (5.0,\deltaY) node(n1)  {\includegraphics[height=2.0 cm]{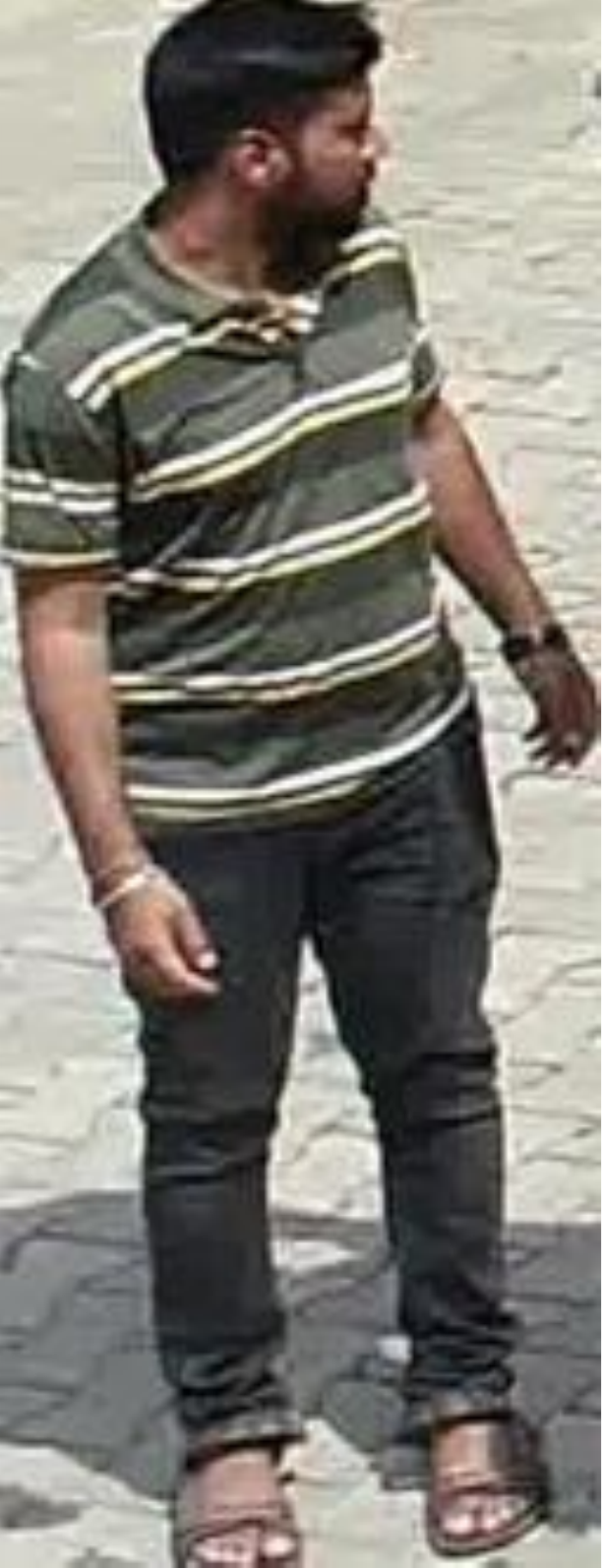}};     
\draw (6.0,\deltaY) node(n1)  {\includegraphics[height=2.0 cm]{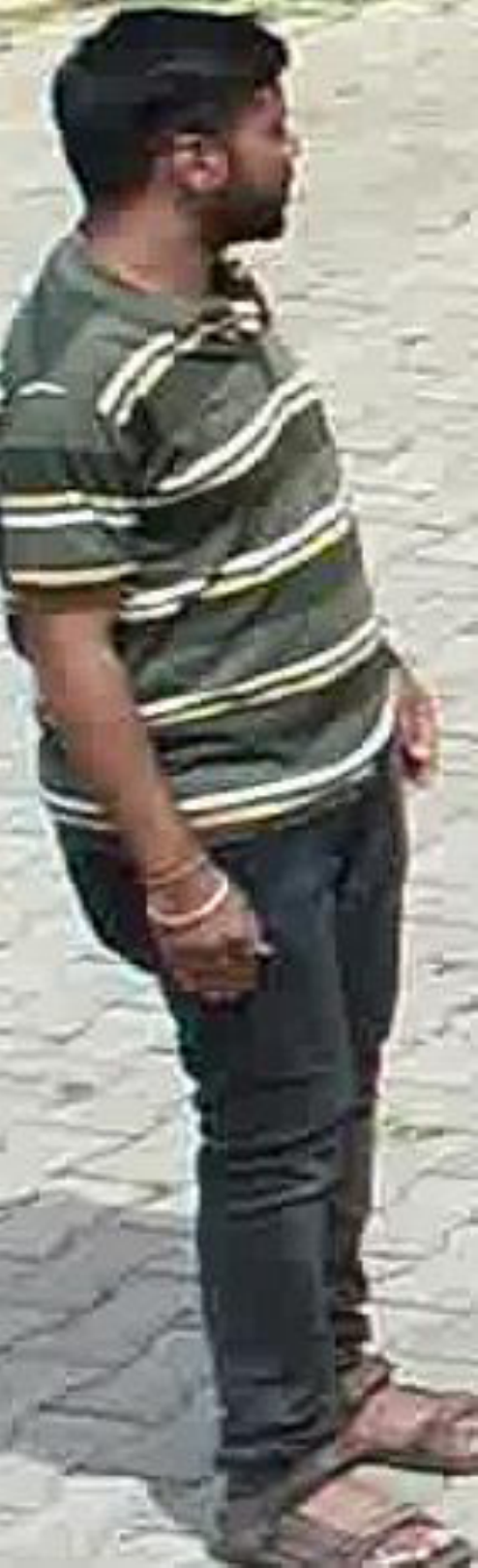}};     
\draw (7.0,\deltaY) node(n1)  {\includegraphics[height=2.0 cm]{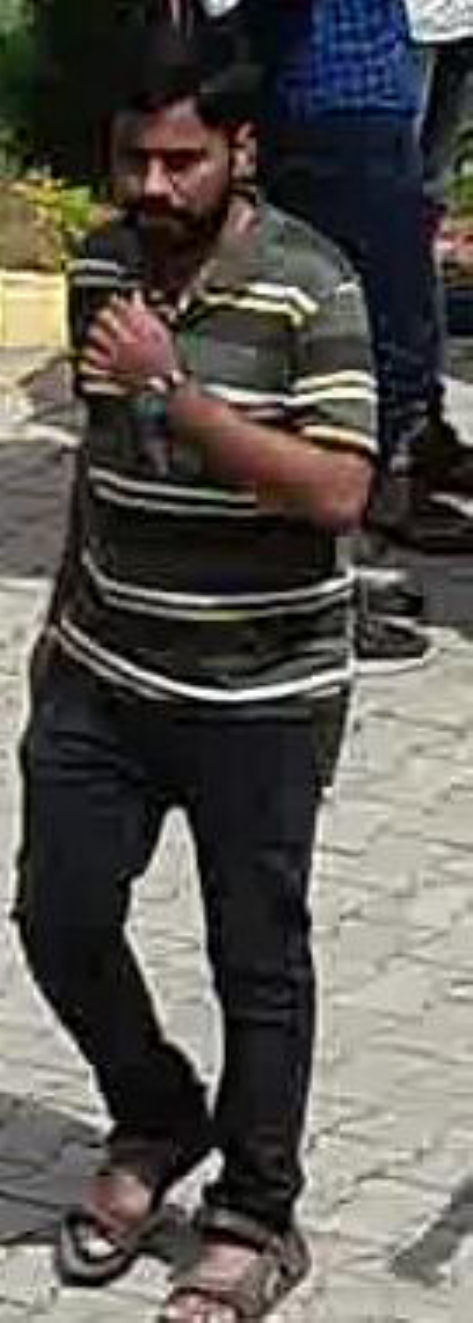}};     
\draw (-1, \deltaY) node[rectangle, rotate=90] {\scriptsize{\textbf{Pose}}};  

\def\deltaY{-11}
\def\deltaX{0.175}
\draw (0.25+\deltaX,\deltaY) node(n1)  {\includegraphics[height=2.0 cm]{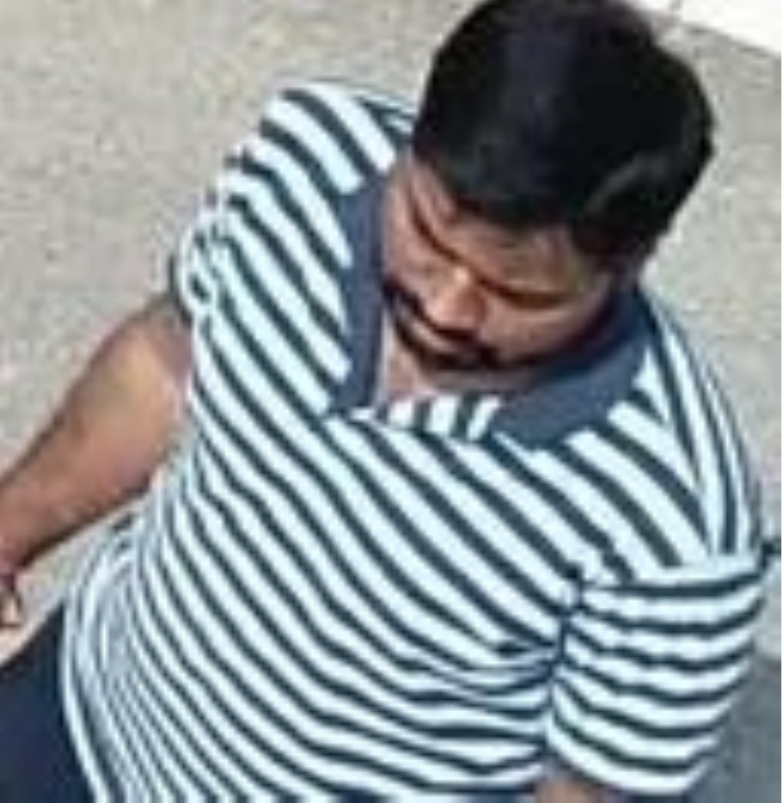}};       
\draw (2.45+\deltaX,\deltaY) node(n1)  {\includegraphics[height=2.0 cm]{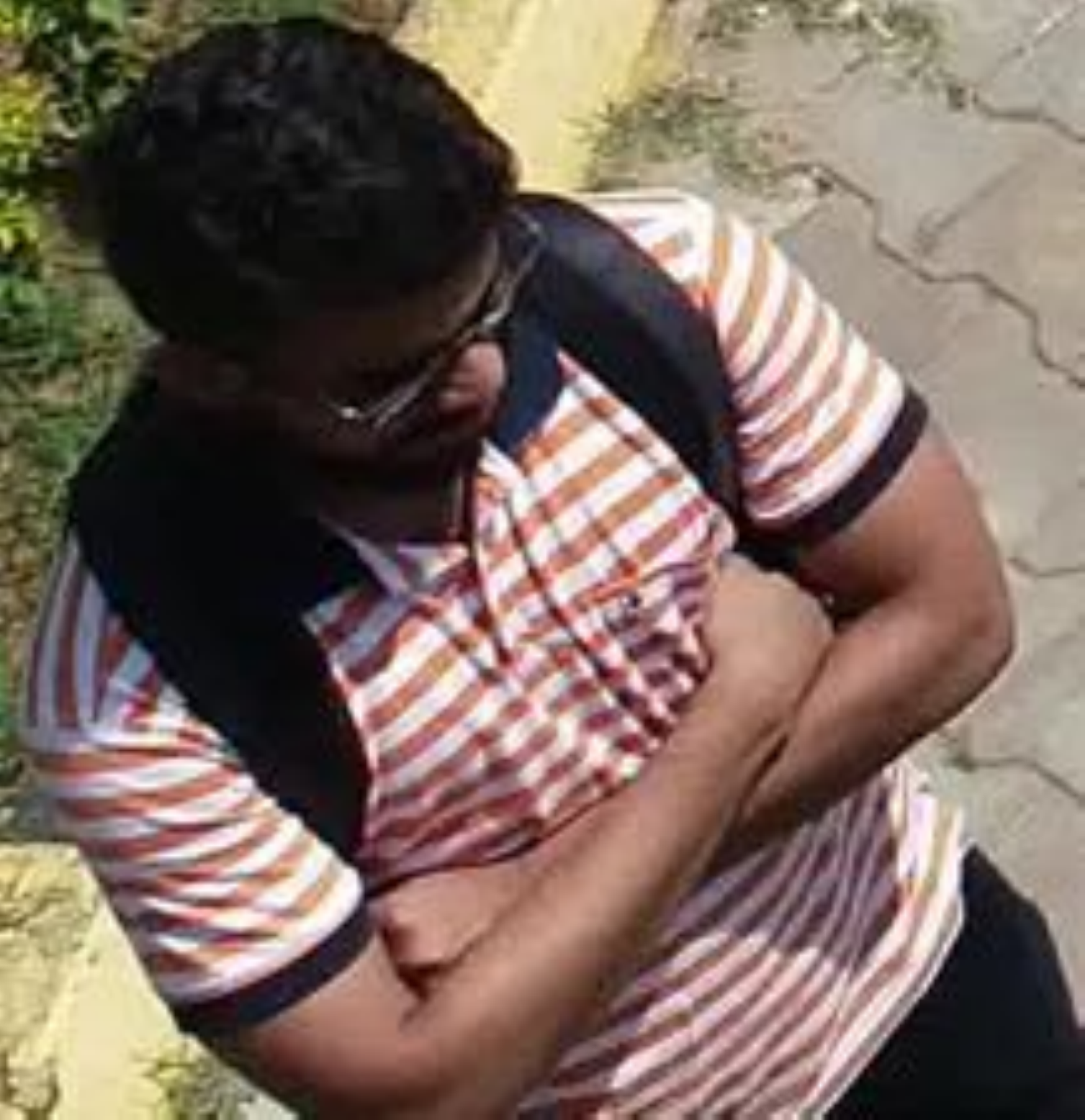}};     
\draw (4.5+\deltaX,\deltaY) node(n1)  {\includegraphics[height=2.0 cm]{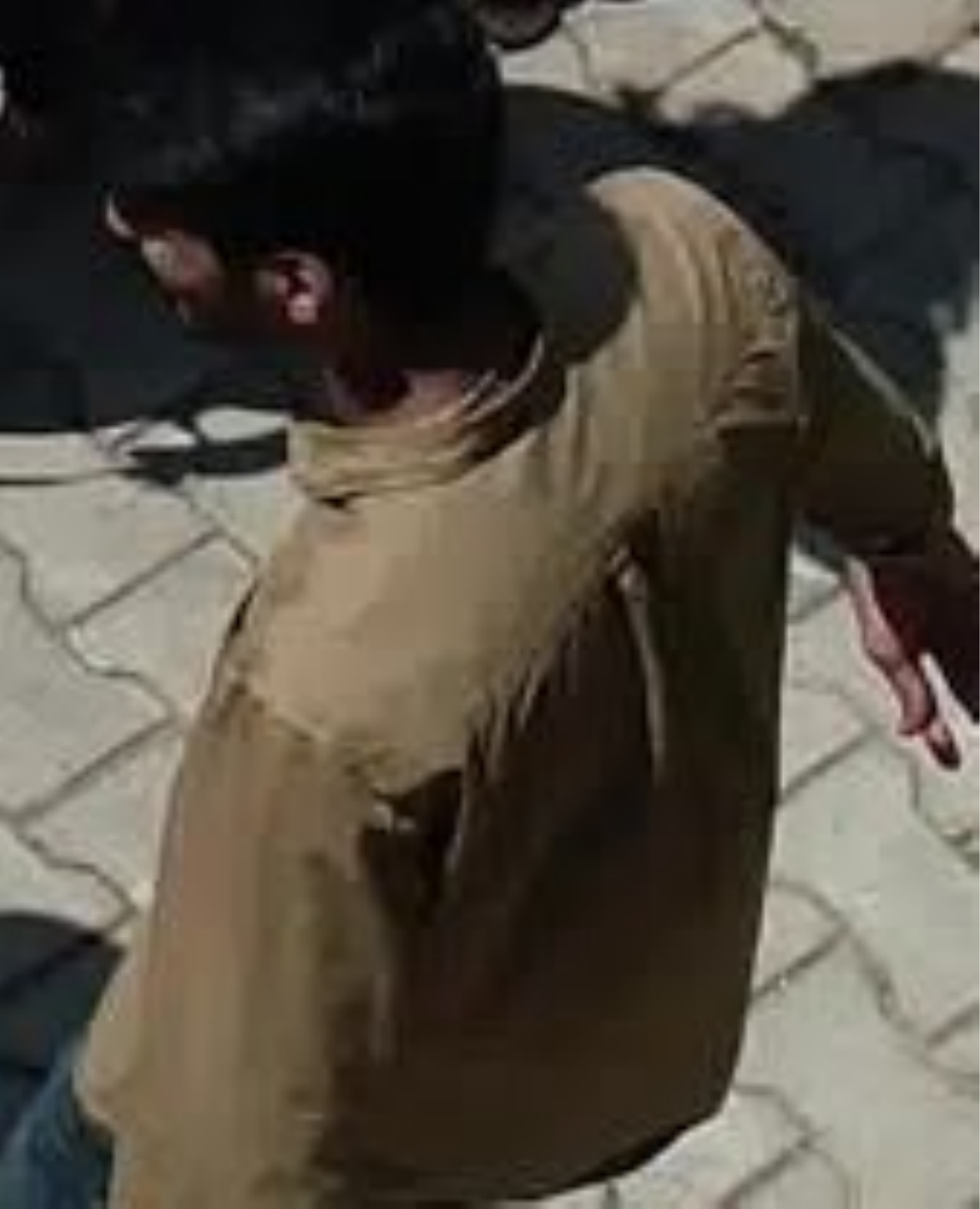}};     
\draw (6.5+\deltaX,\deltaY) node(n1)  {\includegraphics[height=2.0 cm]{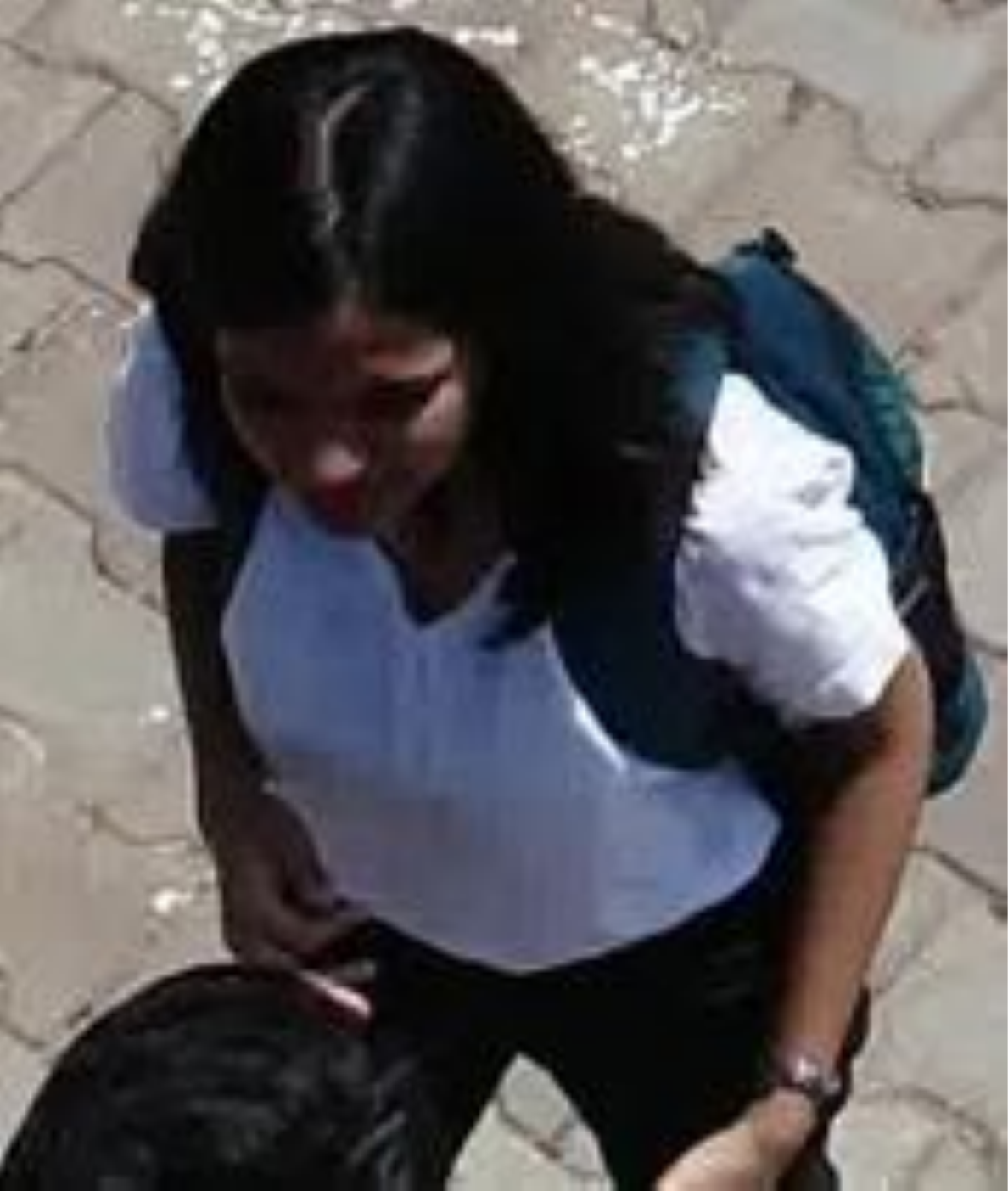}};     
\draw (-1, \deltaY) node[rectangle, rotate=90] {\scriptsize{\textbf{UAV Perspective}}};  

\def\deltaY{-8.8}
\draw (0,\deltaY) node(n1)  {\includegraphics[height=2.0 cm]{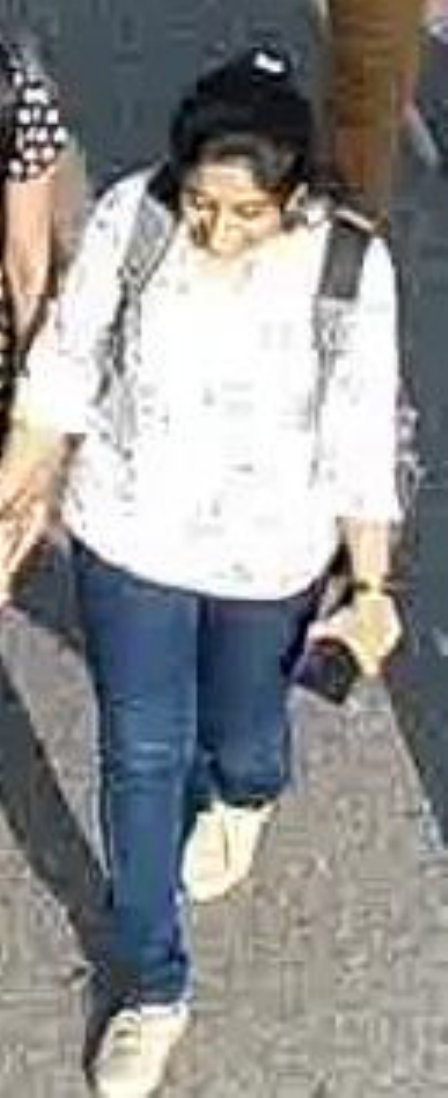}};       
\draw (1.0,\deltaY) node(n1)  {\includegraphics[height=2.0 cm]{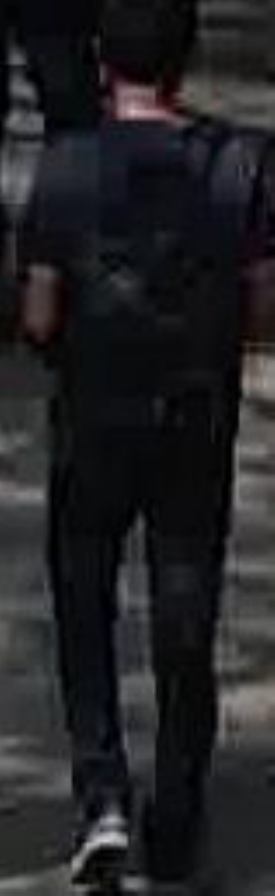}};     
\draw (2.0,\deltaY) node(n1)  {\includegraphics[height=2.0 cm]{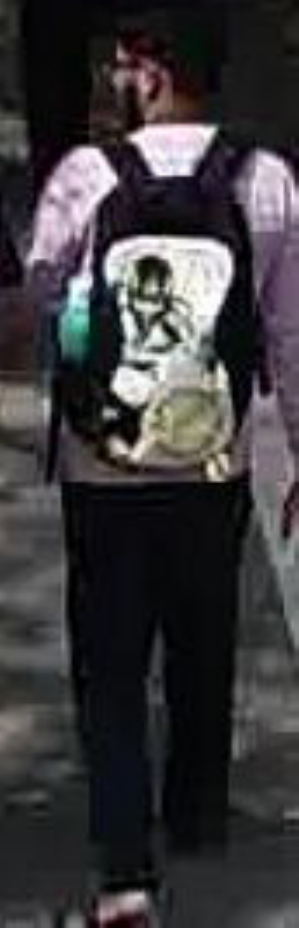}};     
\draw (3.0,\deltaY) node(n1)  {\includegraphics[height=2.0 cm]{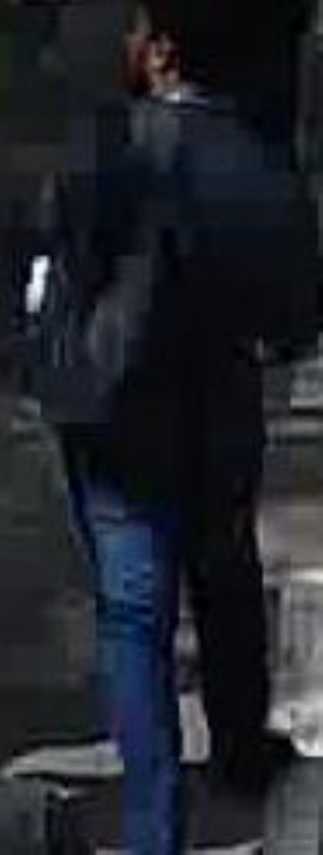}};     
\draw (4.0,\deltaY) node(n1)  {\includegraphics[height=2.0 cm]{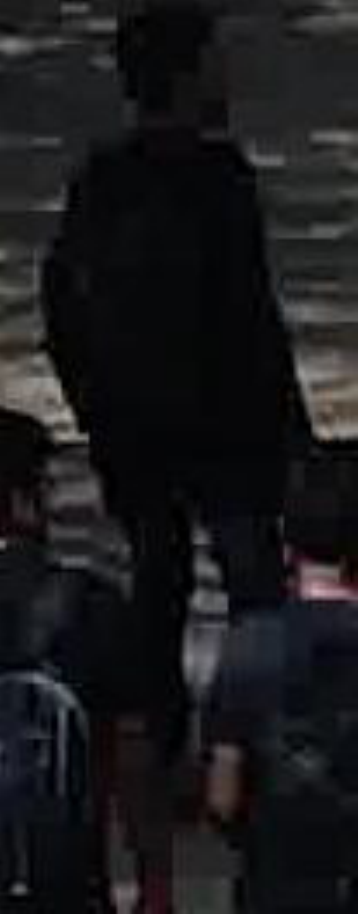}};     
\draw (5.0,\deltaY) node(n1)  {\includegraphics[height=2.0 cm]{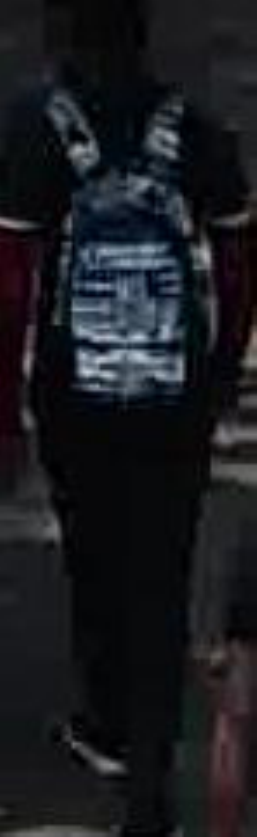}};     
\draw (6.0,\deltaY) node(n1)  {\includegraphics[height=2.0 cm]{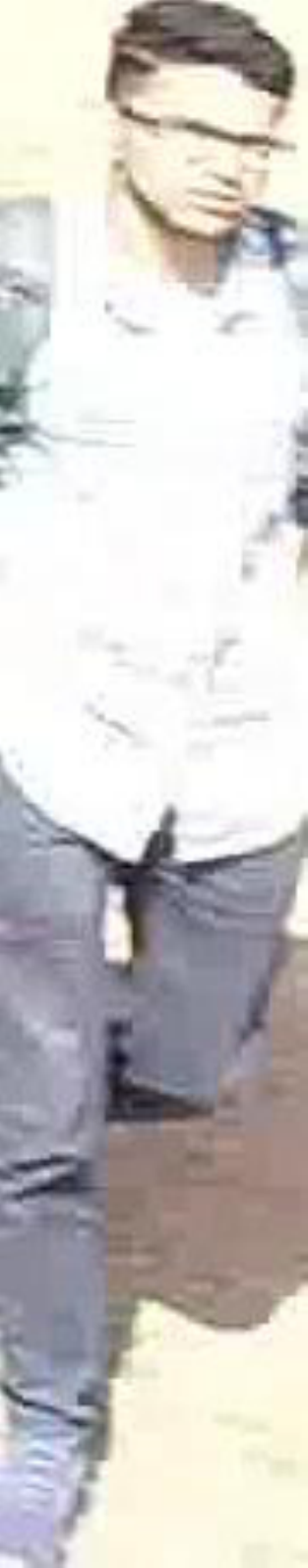}};     
\draw (7.0,\deltaY) node(n1)  {\includegraphics[height=2.0 cm]{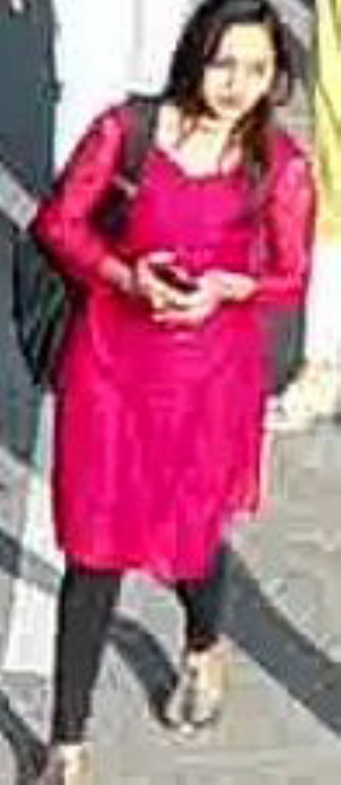}};     
\draw (-1, \deltaY) node[rectangle, rotate=90] {\scriptsize{\textbf{Lighting/Shadows}}};

\end{tikzpicture}
    \caption{Examples of the six factors that - under visual inspection - seem to constitute the major obstacles to perform reliable image analysis in UAV-based data. These are also the predominant data degradation factors in the P-DESTRE dataset.}
        \label{fig:examples}
    \end{center}
\end{figure}

As expected, the acquisition of UAV-based video data in crowded outdoor environments, from at-a-distance and simulating covert protocols, has led to extremely heterogeneous samples, degraded in multiple perspectives. Under visual inspection, we identified six major factors that most frequently reduced the quality of this kind of data, which also augment considerably the challenges of automated image analysis: 

\begin{enumerate}
\item  \textbf{Poor resolution/blur}. As illustrated in the top row of Fig.~\ref{fig:examples}, some subjects were acquired from large distances (over 40 m.), with the corresponding ROIs having very small resolution. Also, some parts of the scenes laid outside the cameras depth-of-field, as a result of a large range in objects depth. This has contributed to the appearance of blurred samples. In both cases, the amount of information available per bounding box is reduced; 

\item  \textbf{Motion blur}. This data degradation factor yielded from the non-stationary nature of the cameras, together with the movements of the subjects. In practice, for some of the bounding boxes, an apparent streaking of the human silhouettes can be observed, which is also an obstacle for automated image analysis;

\item \textbf{Partial occlusions}. As a result of the scene dynamics and due to multiple objects simultaneously in the scenes, partial occlusions of body parts were particularly frequent. According to our perception, this might be the most concerning factor of UAV-based data, as illustrated in the third row of Fig.~\ref{fig:examples}; 

\item  \textbf{Pose}. Under covert data acquisition protocols and without minimally accounting for subjectÕs cooperation, many of the samples regard profile and backside views, in which identification and soft biometric characterisation are particularly difficult to perform;

\item \textbf{Lighting/shadows}. As a consequence of the outdoor conditions, many samples are over/under-illuminated, often with large shadowed regions due to the other objects in the scene (e.g., buildings, cars, trees, traffic signs\ldots); 

\item  \textbf{UAV perspective}. When using gimbal pitch angles close to 90 degrees, the longest axis of some subjectsÕ body is roughly parallel to the camera axis. In such cases, the  images contain almost exclusively a top-view perspective of the heads, and have reduced amounts of discriminating information (bottom row of Fig.~\ref{fig:examples}). 

\end{enumerate}

\subsection{P-DESTRE Statistical Significance}
\label{sec:Statistical}

Let $\alpha$ be a confidence interval. Let $p$ be the error rate of a classifier and $\hat{p}$ be the estimated error rate over a
finite number of test patterns. At an $\alpha$-confidence level, we want that the true error rate does not exceed $\hat{p}$ by an amount larger than  $\varepsilon(n,~\alpha)$. Guyon et al.~\cite{Guyon1989} defined $\varepsilon(n,~\alpha)=\beta p$ as a fraction of $p$. Assuming that recognition errors are Bernoulli trials, authors concluded that the number of required trials $n$ to achieve (1-$\alpha$) confidence in the error rate estimate is given by:

\begin{equation}
\label{eq:guyon} 
n=-ln(\alpha)/(\beta^2 p).
\end{equation}

Using typical values  $\alpha=0.05$ and $\beta=0.2$, authors recommend a simpler form, given by: $n \approx \frac{100}{p}$.

\begin{figure}[ht!]
\begin{center}
\begin{tikzpicture}

\draw (-4.4, 0) node[rectangle, rotate=90] {\scriptsize{\# IDs}};  
\draw (-3, -1.15) node[rectangle] {\scriptsize{Tot. Days}};  
\draw (-3.0,0) node(n1)  {\includegraphics[height=1.9 cm]{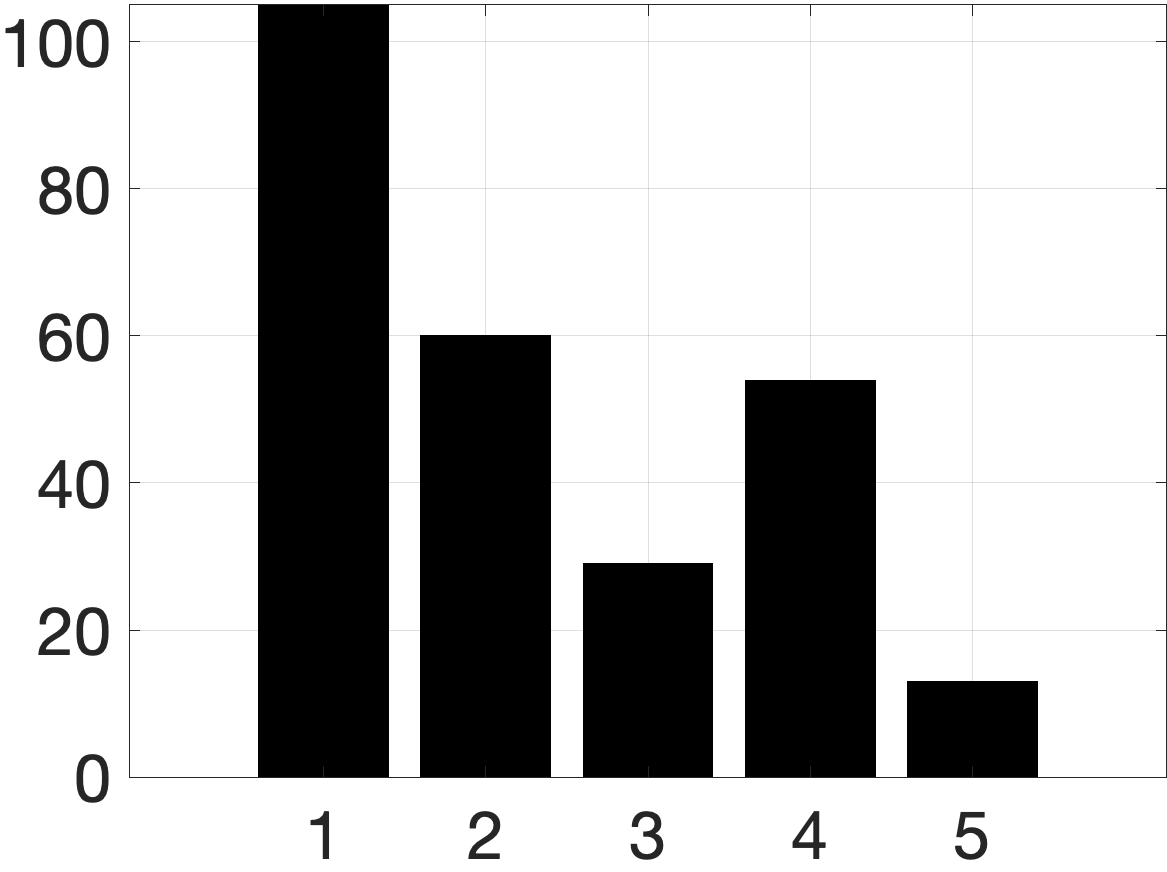}};       

\draw (-1.4, 0) node[rectangle, rotate=90] {\scriptsize{\# IDs}};  
\draw (0, -1.15) node[rectangle] {\scriptsize{Tot. Sessions}};  
\draw (0,0) node(n1)  {\includegraphics[height=1.9 cm]{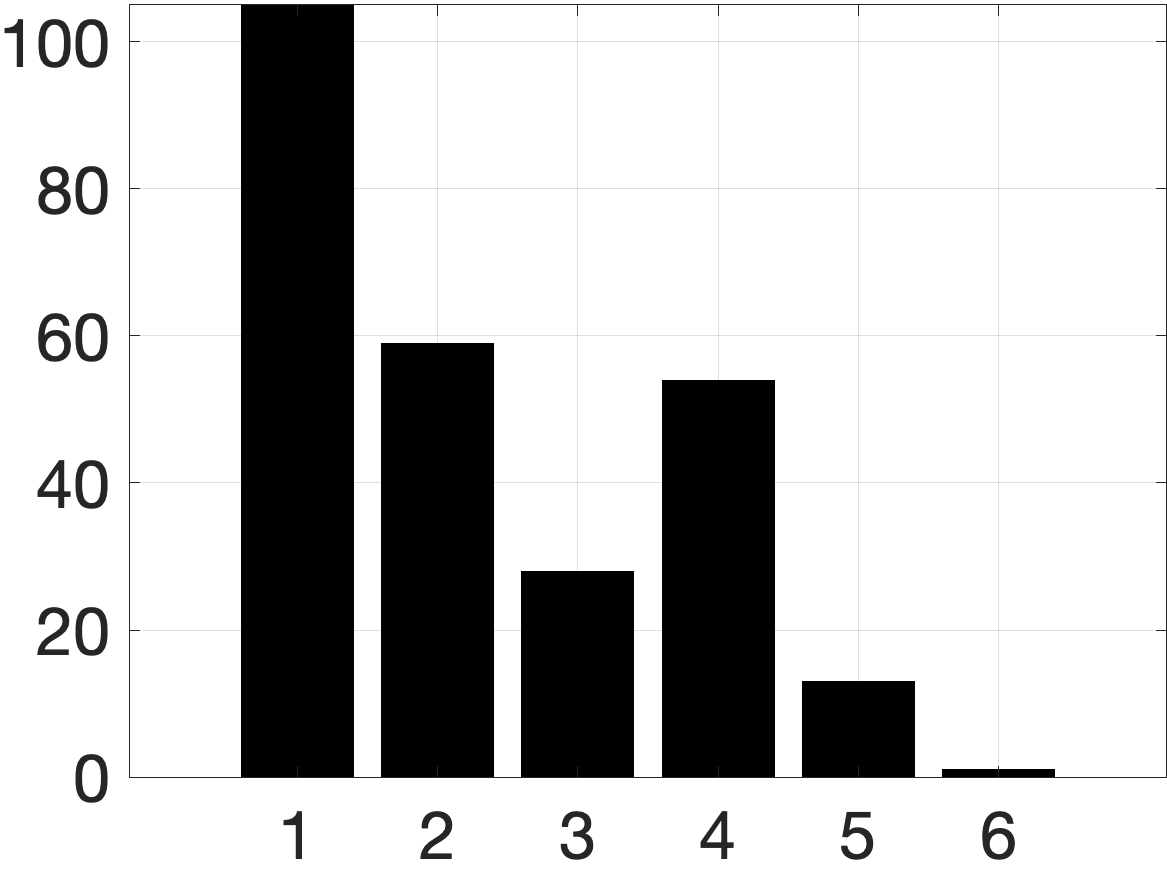}};       

\draw (1.6, 0) node[rectangle, rotate=90] {\scriptsize{\# IDs}};  
\draw (3, -1.15) node[rectangle] {\scriptsize{Tot. Bound. Box}};  
\draw (3,0) node(n1)  {\includegraphics[height=1.9 cm]{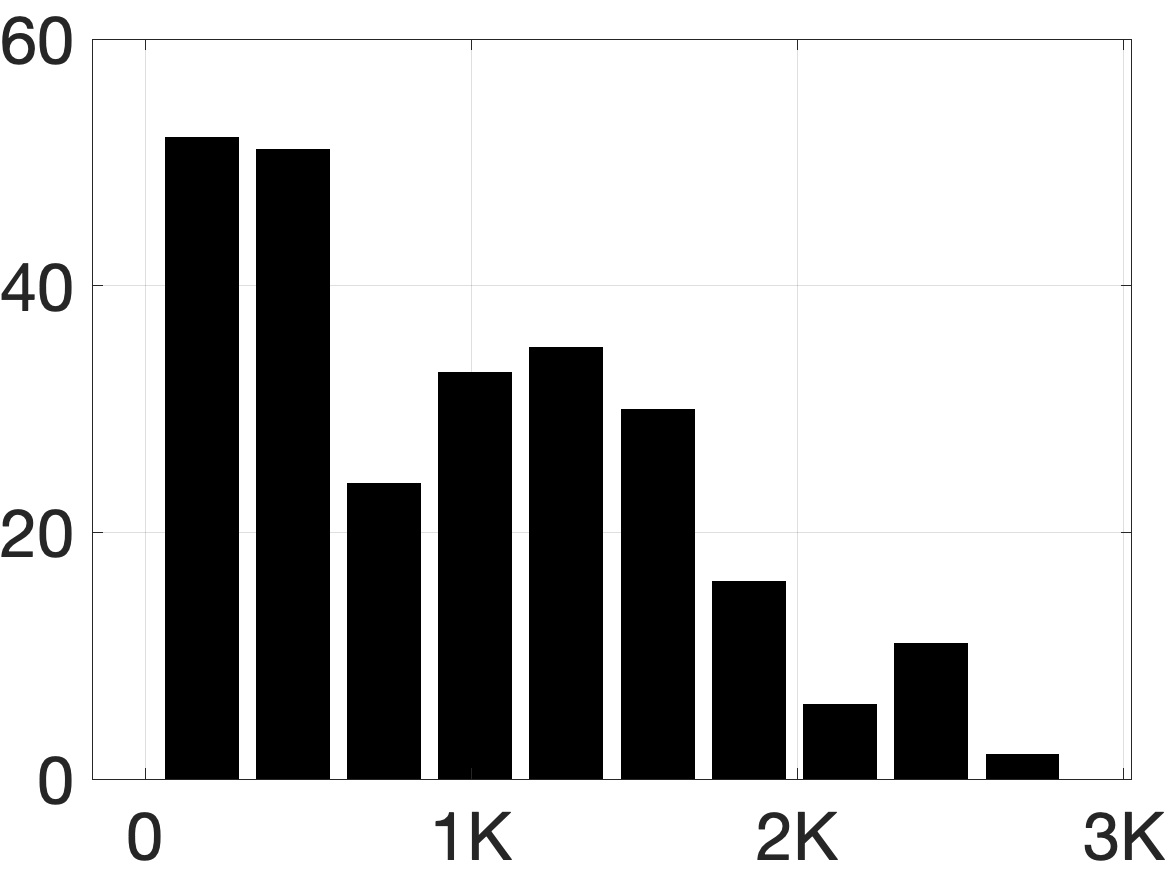}};

\draw (0,-2.3) node(n1)  {\includegraphics[height=1.9 cm]{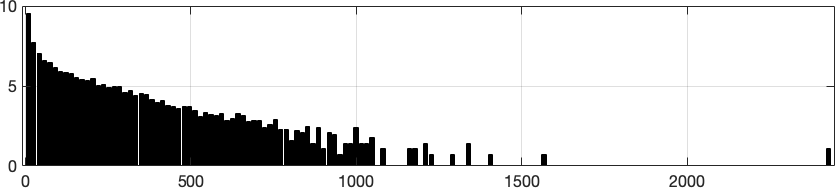}};       
\draw (0, -3.45) node[rectangle] {\scriptsize{Tracklet Length}};  

\draw (-4.4, -2.3) node[rectangle, rotate=90] {\scriptsize{log(\# Sequences)}};  

\end{tikzpicture}
    \caption{P-DESTRE statistics. Top row: number of days with data per volunteer (at left), number of data acquisition sessions per volunteer (at center) and number of bounding boxes per volunteer (at right). The bottom row provides the statistics about the length of the tracklet sequences.}
        \label{fig:statistics}
    \end{center}
\end{figure}
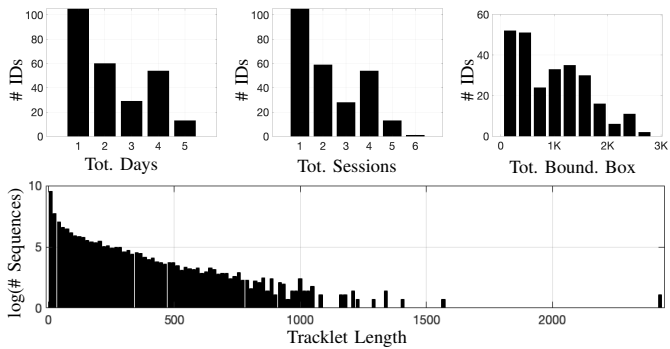

Considering the statistics of the P-DESTRE datasets (Fig.~\ref{fig:statistics}), in terms of the number of data acquisition sessions/days per volunteer and the number of  bounding boxes per volunteer/session, it is possible to obtain the lower bounds for the statistical confidence in experiments related with identity verification at frame level, assuming the 1) re-identification; and 2) search problems.

 In the person re-identification setting, considering that each frame (bounding box) with a known ID ($\geq 1$) generates a valid template, that all frames of the same ID acquired in different sessions of the same day can be used to generate the \emph{genuine} pairs and that frames with different IDs (including '\emph{unknown}') compose the \emph{impostor} set, the P-DESTRE dataset enables to perform  1,246,587,154 (genuine) + 605,599,676,264 (impostor) comparisons, leading to a $\hat{p}$ value with a lower bound of approximately $1.647 \times 10^{-10}$. Regarding the person search problem, where \emph{genuine} pairs must have been acquired in different days, the dataset enables to perform 2,160,586,581 (genuine) + 605,599,676,264 (impostor) comparisons, leading to a $\hat{p}$ value with a lower bound of approximately $1.645 \times 10^{-10}$. Note that these are lower bounds, that do not take into account the portions of data used for learning purposes. Also, these values will increase if we do not assume the independence between images and error correlations are taken into account.

\section{Experiments and Results}
\label{sec:Results}

In this section we report the results obtained by methods considered to represent the state-of-the-art, in four tasks: 1) pedestrian detection; 2) pedestrian tracking; 3) pedestrian re-identification; and 4) pedestrian search. For contextualisation, we report not only the performance obtained by such techniques in the P-DESTRE dataset, but also provide as baseline the results we observed for the same techniques in datasets that are well-known in the computer vision literature. For each problem, we also illustrate the typical failure cases that we have subjectively perceived during our experiments, which should provide the motivation for further advances in each of the problems considered.  

\subsection{Pedestrian  Detection}

 The RetinaNet~\cite{Lin2020} and R-FCN~\cite{Dai2016} methods were considered to represent the state-of-the-art in pedestrian detection, as both outperformed in the PASCAL VOC 2007/2012~\cite{Everingham2011} challenge ('Person Detection' category). In order to perceive the hardness of P-DESTRE for object detection, we compared the P-DESTRE performance of both methods against the values obtained in the PASCAL VOC 2007/2012 set, which is among the most frequently seen in object detection literature. 

In summary, RetinaNet uses a feature pyramid network as backbone, on top of a ResNet architecture. Two disjoint sub-networks respectively classify anchor boxes and adjust the values with respect to the default anchors. R-FCN uses a fully convolutional architecture~\cite{Dai2016}, where the translation invariance is obtained by a set of position-sensitive score maps that uses specialized convolutional layers to encode the deviations with respect to default positions. A position sensitive ROI pooling layer is  appended on top of the fully connected layers. 

For the PASCAL VOC 2007/2012  set, the official development kit\footnote{\url{http://host.robots.ox.ac.uk/pascal/VOC/voc2012/\#devkit}} was used to evaluate both methods on the 'Person' category. For the P-DESTRE set, 10-fold cross validation was used, with the data in each split randomly divided into 60\% for learning, 20\% for validation and 20\% for testing purposes. The full specification of the samples used in each split and of the scores returned by each method is provided in\footnote{\url{http://p-destre.di.ubi.pt/experiments.html}}.  

\begin{table}[h!]
\centering
     \caption{Comparison between the average precision (AP) obtained by two methods considered to represent the state-of-the-art in person detection, in the P-DESTRE and PASCAL VOC 2007/2012 sets.}
     \label{tab:detection}
\begin{tabular}{|c|c|C{2.25cm}|C{2.25cm}|}
\hline
\textbf{\scriptsize{Method}}  & \textbf{\scriptsize{Backbone}} & \textbf{\scriptsize{PASCAL VOC}}  & \textbf{\scriptsize{P-DESTRE}} \\ \hline
\scriptsize{RetinaNet~\cite{Lin2020}}  & \scriptsize{ResNet-50} &\scriptsize{86.44 $\pm$ 1.03}  & \scriptsize{63.10 $\pm$ 1.64}\\ \hline
\scriptsize{R-FCN~\cite{Dai2016}}  & \scriptsize{ResNet-101}& \scriptsize{84.43 $\pm$ 1.85}  & \scriptsize{59.29 $\pm$ 1.31}\\ \hline
\end{tabular}
\end{table}

\begin{figure}[ht!]
\begin{center}
\begin{tikzpicture}

\def\posX{0}
\def\posY{1.6}
\def\deltaY{0}

\fill [rounded corners, gray] (\posX-1, \posY-0.15+\deltaY) rectangle (\posX+1, \posY+0.15+\deltaY);    
\draw  [white] (\posX, \posY+\deltaY) node {\scriptsize{RetinaNet}};    

\draw (0,0+\deltaY) node(n1)  {\includegraphics[width=7.5 cm]{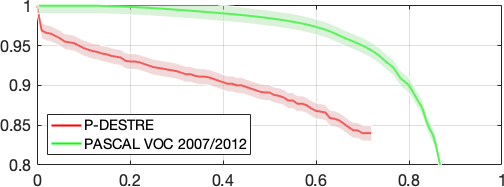}};     
\draw (-3.95, 0+\deltaY) node[rectangle, rotate=90] {\scriptsize{Precision}};  
\draw (0,-1.5+\deltaY) node[rectangle] {\scriptsize{Recall}};  

\def\deltaY{-3.75}
\def\deltaX{0}

\fill [rounded corners, gray] (\posX-1, \posY-0.15+\deltaY) rectangle (\posX+1, \posY+0.15+\deltaY);    
\draw  [white] (\posX, \posY+\deltaY) node {\scriptsize{R-FCN}};    
 
\draw (0,0+\deltaY) node(n1)  {\includegraphics[width=7.5 cm]{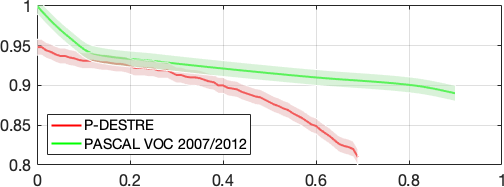}};     
\draw (-3.95, 0+\deltaY) node[rectangle, rotate=90] {\scriptsize{Precision}};  
\draw (0,-1.5+\deltaY) node[rectangle] {\scriptsize{Recall}};

\end{tikzpicture}
    \caption{Comparison between the precision/recall curves observed in the PASCAL VOC 2007/2012 (red lines) and P-DESTRE (green lines) sets for the RetinaNet (top plot) and R-FCN (bottom plot) detection methods.}
        \label{fig:PR}
    \end{center}
\end{figure}
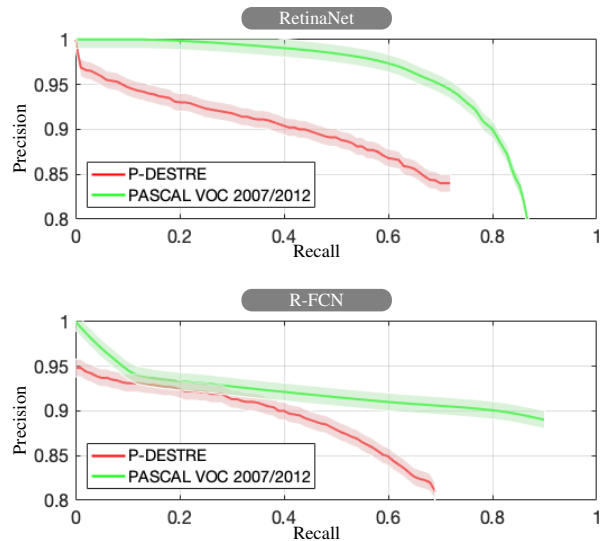

The results are summarized in Table~\ref{tab:detection} for both datasets and methods, in terms of the average precision (at intersection of union values equal to 0.5, AP@IoU=0.5) obtained. Also, Fig.~\ref{fig:PR} provides the precision/recall curves for both data sets and detection techniques, where the P-DESTRE values are represented by red lines and the PASCAL VOC 2007/2012 results by green lines. The shadowed regions denote the standard deviation performance in the 10 splits, at each operating point.  Overall, both methods decreased notoriously their effectiveness from the PASCAL VOC set to the P-DESTRE set, in some cases with error rates increasing over 160\%. In the case of the R-FCN method, in a  small region of the performance space (recall $\approx$ 0.2), the levels of performance for P-DESTRE and PASCAL VOC were approximately equal, yet the precision values then remain stable for much higher recall values in the PASCAL VOC set.

\begin{figure}[ht!]
\begin{center}
\begin{tikzpicture}

\draw (0,0) node(n1)  {\includegraphics[height=2.5 cm]{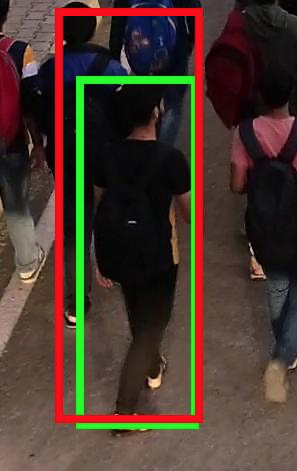}};       
\draw (1.45,0) node(n1)  {\includegraphics[height=2.5 cm]{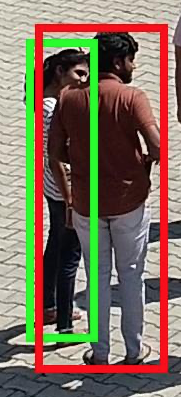}};    
\def\deltaY{-2.75}
\def\deltaX{0}
\draw (0+\deltaX-0.2,0+\deltaY) node(n1)  {\includegraphics[height=2.5 cm]{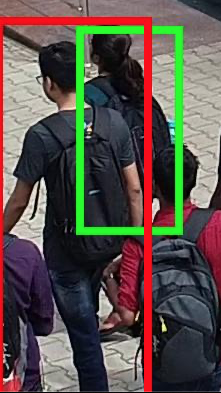}};       
\draw (1.45+\deltaX-0.1,0+\deltaY) node(n1)  {\includegraphics[height=2.5 cm]{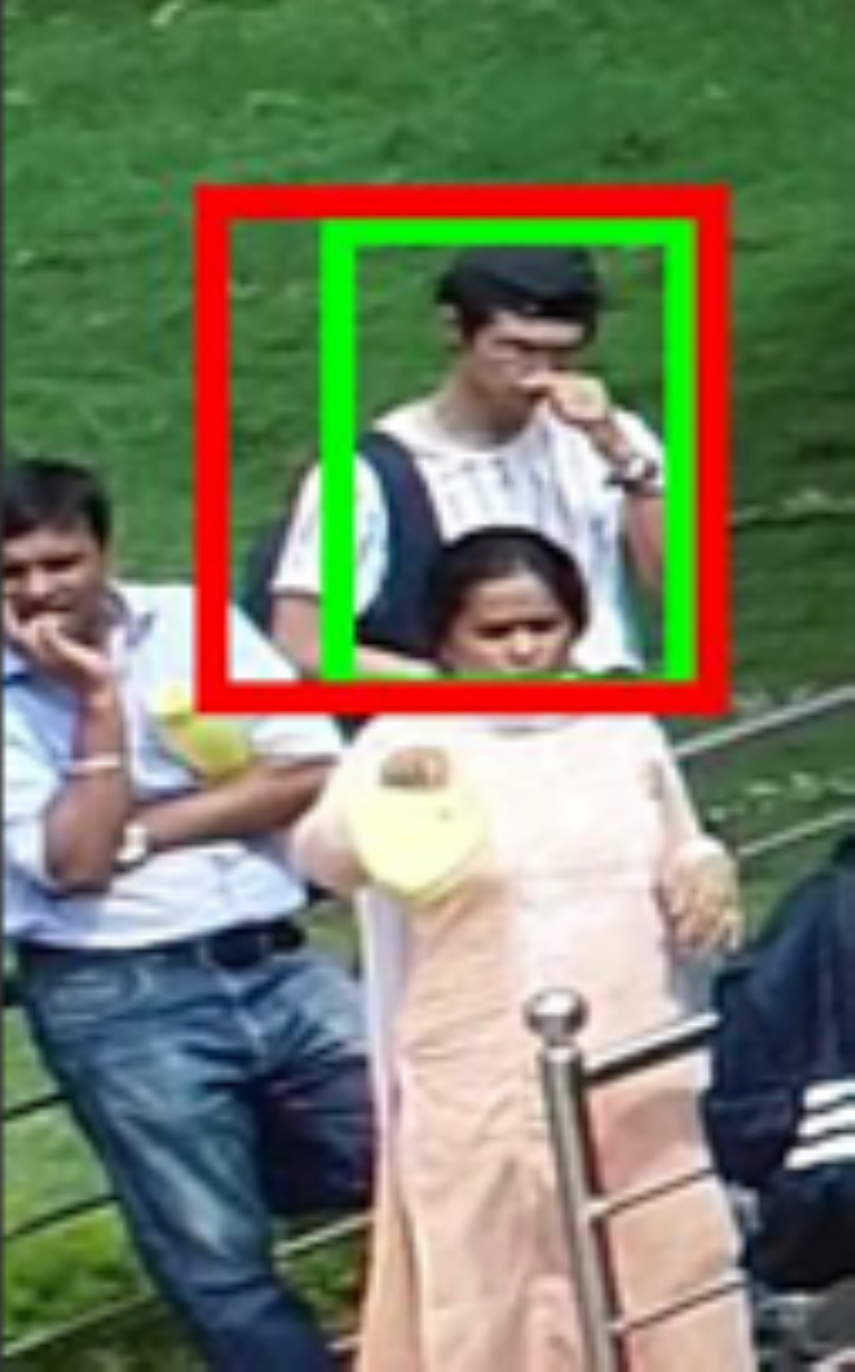}};    
\draw [rounded corners, black, very thick, dashed] (-1, -4.45) rectangle (2.25,1.5);    
\draw (0.65, -4.2) node[rectangle] {\scriptsize{\textbf{Crowded}}};

\draw (3.2,0) node(n1)  {\includegraphics[height=2.5 cm]{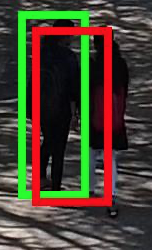}};     
\draw (4.75,0) node(n1)  {\includegraphics[height=2.5 cm]{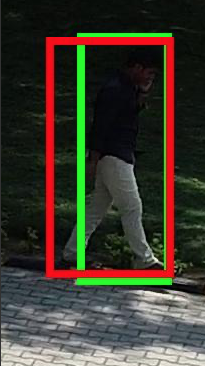}};     
\draw [rounded corners, black, very thick, dashed] (2.35, -1.75) rectangle (5.6,1.5);    
\draw (4.0, -1.5) node[rectangle] {\scriptsize{\textbf{Shadows/Lighting}}};  

\draw (6.7,0) node(n1)  {\includegraphics[height=2.5 cm]{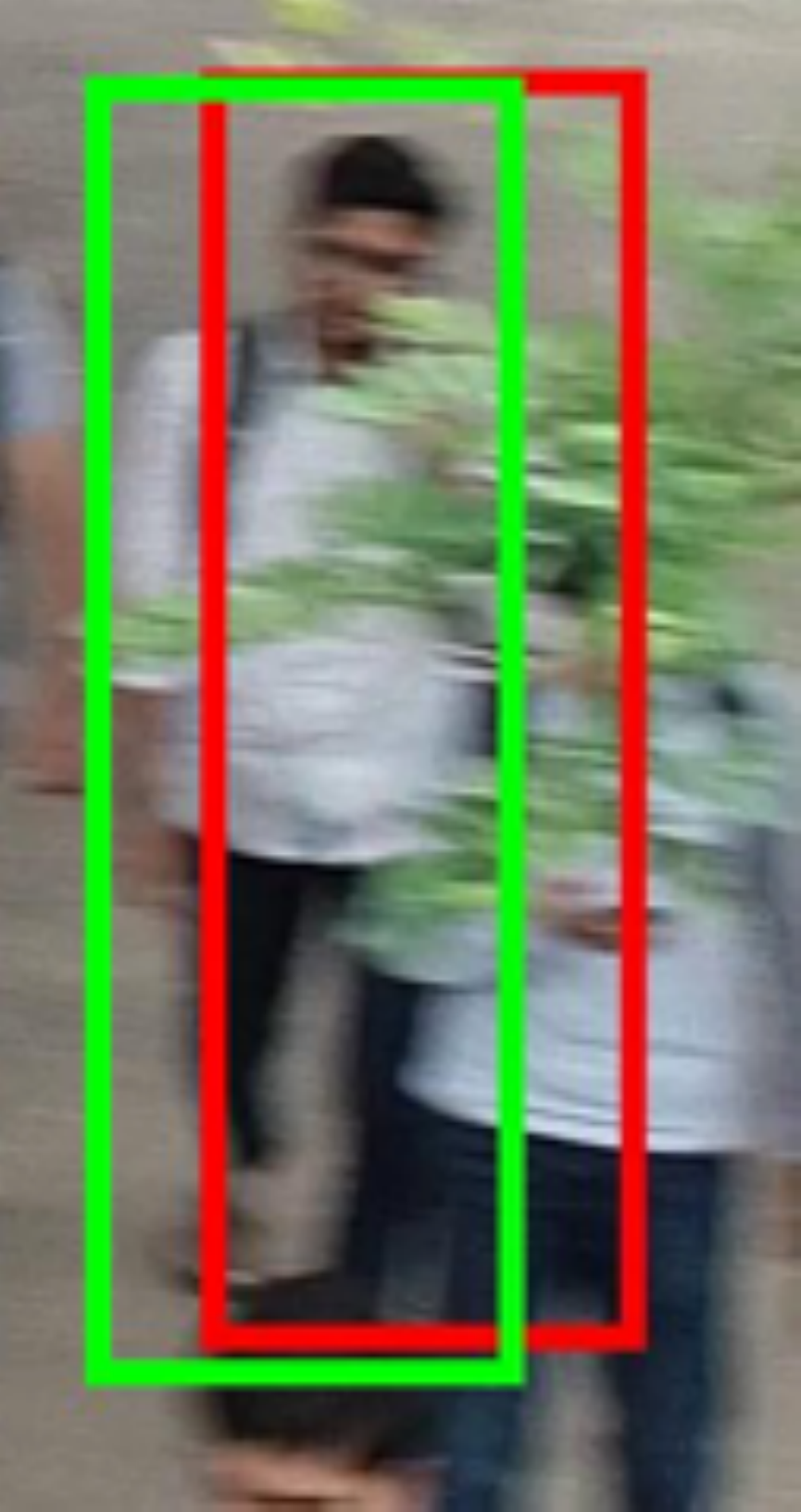}};     
\draw (6.7, -1.5) node[rectangle] {\scriptsize{\textbf{Motion}}}; 
\draw [rounded corners, black, very thick, dashed] (5.85, -1.75) rectangle (7.5,1.5);

\draw (3.0,-3) node(n1)  {\includegraphics[height=2.0 cm]{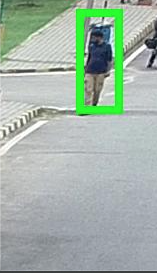}};     
\draw (4.3,-3) node(n1)  {\includegraphics[height=2.0 cm]{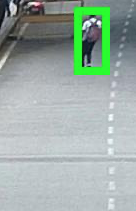}};     
\draw (5.6,-3) node(n1)  {\includegraphics[height=2.0 cm]{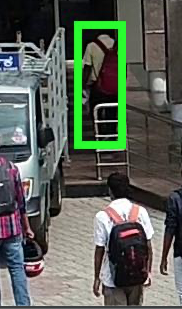}};     
\draw (6.8,-3) node(n1)  {\includegraphics[height=2.0 cm]{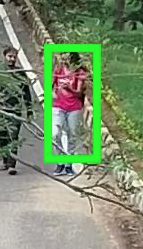}};     
\draw (5.0, -4.2) node[rectangle] {\scriptsize{\textbf{Poor resolution}}}; 
\draw [rounded corners, black, very thick, dashed] (2.35, -4.45) rectangle (7.5,-1.85);



\end{tikzpicture}
    \caption{Typical cases where both methods produced the worst detection scores, i.e., failed to appropriately detect the pedestrians. The green boxes represent the ground-truth, while red colour denotes the detected boxes.}
        \label{fig:examples_detection}
    \end{center}
\end{figure}
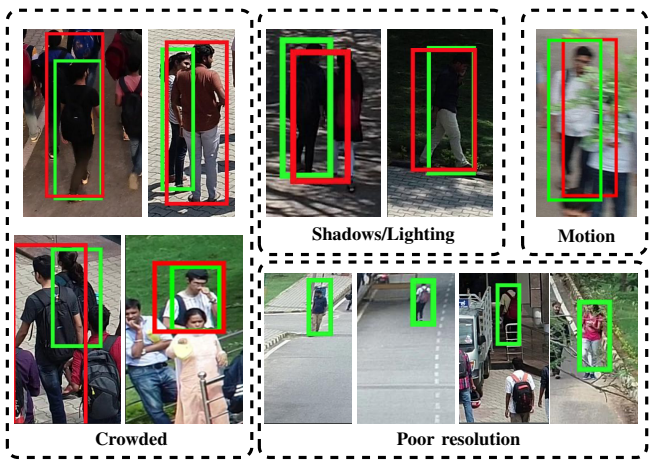

In our qualitative analysis, we observed that both methods faced particular difficulties in crowded scenes, when only a small proportion of the subjects silhouette is available, as illustrated in Fig.~\ref{fig:examples_detection}. Considering that RetinaNet is anchor-based, and that the predefined anchor boxes have a set of handcrafted aspect ratios and scales that are data dependent, performance might have been seriously affected. Even though RetinaNet has clearly outperformed R-FCN, the challenging conditions in the P-DESTRE set had still notoriously degraded its effectiveness, when compared to the PASCAL VOC baseline. By careful analysis of the instances in both sets, we concluded that the P-DESTRE has notoriously more \emph{hard} cases than PASCAL VOC, with severely degraded data (i.e., severe occlusions, poor resolution and local lighting variations/shadows). 

As a summary, these experiments point for the requirement of developing novel strategies to handle the specific features that yield from UAV-based data acquisition. Not only the state-of-the-art solutions provide levels of performance that are still far from the demanded to deploy this kind of solutions in real-environments, but they are also particularly sensitive to some of the most frequent data degradation factors in UAV-based imaging (e.g., motion-blur and shadows). Another particularly concerning factor is the density of subjects in the scene, with crowded environments easily providing severe occlusions in most of the subjects that constraint the effectiveness of the object detection phase.

\subsection{Pedestrian Tracking}

For the tracking task, the TracktorCV~\cite{Bergmann2019} and V-IOU~\cite{Bochinski2018} methods were selected to represent the state-of-the-art, according to two reasons: 1) their performance in the MOT challenge\footnote{\url{https://motchallenge.net}}; and 2) the fact that both provide freely available implementations, which is particularly important to guarantee that we obtain a fair evaluation between datasets. This way, we compared the effectiveness attained by these techniques in the P-DESTRE and in the MOT challenge set, once again to perceive the relative hardness of tracking pedestrians from UAV-based data, in comparison to a stationary-cameras tracking task. 

The TracktorCV method comprises two steps: 1) a regression module, that uses the input of the object detection step to update the position of the bounding box at a subsequent frame; and 2) an object detector that provides the set of bounding boxes for the next frames. The V-IOU algorithm is an extension of the IOU algorithm~\cite{Bochinski2017} that attenuates the problem of false negatives, by associating the detections in consecutive frames according to spatial overlap information. For both methods, the hyper-parameters were tuned according to the way authors suggested, and are given in\footnote{\url{http://p-destre.di.ubi.pt/parameters_tracking.zip}}.

In terms of performance measures, our analysis was based in the Multiple Object Tracking Accuracy (MOTA), Multiple Object Tracking Precision (MOTP) and F1 values, as described in~\cite{Bernardin2008}. The summary results attained by both algorithms and datasets are given in Table~\ref{tab:tracking}.  Once again, a consistent degradation in performance from the MOT-17 to the P-DESTRE set was observed, even though the deterioration was in absolute terms far less than the observed for the detection task (here, an decrease in the F1 values of around 10\% was observed). 

When comparing both methods, the Tracktor-Cv outperformed the V-IOU both in non-aerial and aerial data, decreasing the error rates around 9\%.  For both techniques, we observed a positive correlation between their typical failure cases, which were invariably related to crowded scenes, and two particularly concerning cases: 1) scenes where, due to extreme pedestrian density, subjects' trajectories cross others at every moment; and 2) when severe occlusions of the human silhouettes occur. Both factors augment the likelihood of observing \emph{fragmentations}, i.e., with the trackers erroneously switching identities of two trajectories in the scene, and wrong \emph{merge} cases, with the trackers erroneously merging two ground truth identities into a single one.

When subjectively comparing the data in MOT-17 to the P-DESTRE dataset, it is evident that P-DESTRE contains more complex scenarios, with more cluttered backgrounds (e.g., many scenes have 'grass' grounds and several tree branches) and poor resolution subjects, in result of data acquisition from large distances. Also, we noted that the trackability of pedestrians also depends on the tracklet length (i.e., number of consecutive frames where an object appears), with the values in MOT-17 varying from 1 to 1,050 (average 304) and in the P-DESTRE varying from 4 to 2,476 (average 63.7 $\pm$ 128.8), as illustrated in Fig.~\ref{fig:statistics}.

\begin{table}[h!]
\centering
     \caption{Comparison between the tracking performance attained by two state-of-the-art algorithms in the P-DESTRE and MOT data sets.}
     \label{tab:tracking}
\begin{tabular}{|c|C{1.25cm}|C{1.25cm}|C{1.25cm}|C{1.25cm}|}
\hline
\textbf{\scriptsize{Method}}  & \textbf{\scriptsize{Dataset}}  & \textbf{\scriptsize{MOTA}} & \textbf{\scriptsize{MOTP}}  & \textbf{\scriptsize{F-1}}\\ \hline
 \multirow{2}{*}{ \scriptsize{TracktorCv~\cite{Bergmann2019}}} &   \textbf{\scriptsize{MOT-17}} & \scriptsize{65.20 $\pm$ 9.60} & \scriptsize{62.30 $\pm$ 11.00} & \scriptsize{89.60 $\pm$ 2.80} \\ \cline{2-5}
& \textbf{\scriptsize{P-DESTRE}} & \scriptsize{56.00 $\pm$ 3.70} & \scriptsize{55.90 $\pm$ 2.60} & \scriptsize{87.40 $\pm$ 2.00} \\ \hline
 \multirow{2}{*}{ \scriptsize{V-IOU~\cite{Bochinski2018}}} &   \textbf{\scriptsize{MOT-17}} & \scriptsize{52.50 $\pm$ 8.80} & \scriptsize{57.50 $\pm$ 9.50} & \scriptsize{86.50 $\pm$ 1.90} \\ \cline{2-5}
& \textbf{\scriptsize{P-DESTRE}} & \scriptsize{47.90 $\pm$ 5.10} & \scriptsize{51.10 $\pm$ 5.80} & \scriptsize{83.30 $\pm$ 8.40} \\ \hline

\end{tabular}
\end{table}

\begin{figure}[ht!]
\begin{center}
\begin{tikzpicture}

\draw (0,0) node(n1)  {\includegraphics[width=8.0 cm]{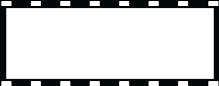}};       

\def\deltaX{-2.55}

\draw (0+\deltaX,0) node(n1)  {\includegraphics[height=2.5 cm]{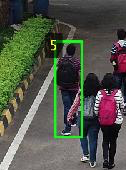}};       
\draw (1.75+\deltaX,0) node(n1)  {\includegraphics[height=2.5 cm]{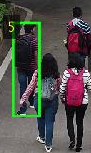}};    
\draw (3.475+\deltaX,0) node(n1)  {\includegraphics[height=2.5 cm]{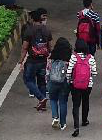}};     
\draw (5.25+\deltaX,0) node(n1)  {\includegraphics[height=2.5 cm]{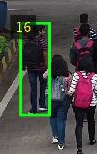}};     

\def\deltaA{-0.3}

\draw (0.0+\deltaX, -1.5+\deltaA) node[rectangle] {{\textbf{\cmark}}}; 
\draw (1.75+\deltaX, -1.5+\deltaA) node[rectangle] {{\textbf{\cmark}}}; 
\draw (3.475+\deltaX, -1.5+\deltaA) node[rectangle] {{\textbf{\xmark \tiny{MD}}}}; 
\draw (5.25+\deltaX, -1.5+\deltaA) node[rectangle] {{\textbf{\xmark \tiny{WL}}}}; 

\def\deltaY{-3.75}
\def\deltaX{-2.65}

\draw (0,0+\deltaY) node(n1)  {\includegraphics[width=8.0 cm]{imgs/images_2.png}};       

\draw (0+\deltaX,0+\deltaY) node(n1)  {\includegraphics[height=2.5 cm]{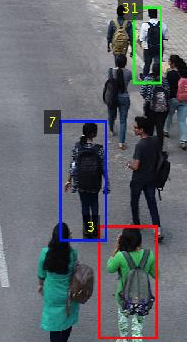}};       
\draw (1.55+\deltaX,0+\deltaY) node(n1)  {\includegraphics[height=2.5 cm]{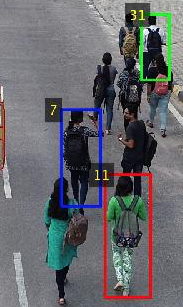}};    
\draw (3.225+\deltaX,0+\deltaY) node(n1)  {\includegraphics[height=2.5 cm]{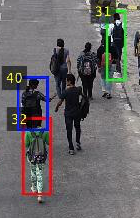}};     
\draw (5.1+\deltaX,0+\deltaY) node(n1)  {\includegraphics[height=2.5 cm]{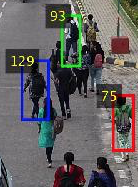}};     

\draw (0.0+\deltaX, -1.5+\deltaY+\deltaA) node[rectangle] {{\textbf{\cmark}~{\cmark}~{\cmark}}}; 
\draw (1.55+\deltaX, -1.5+\deltaY+\deltaA) node[rectangle] {{\textbf{\xmark{\tiny{WL}}~\cmark~\cmark}}}; 
\draw (3.225+\deltaX, -1.5+\deltaY+\deltaA) node[rectangle] {{\textbf{\xmark{\tiny{WL}}~\xmark{\tiny{WL}}~\cmark}}}; 
\draw (5.1+\deltaX, -1.5+\deltaY+\deltaA) node[rectangle] {{\textbf{\xmark{\tiny{WL}}~\xmark{\tiny{WL}}~\xmark{\tiny{WL}}}}}; 

\end{tikzpicture}
    \caption{Examples of sequences where both tracking methods faced difficulties and have - at some point - missed the ground truth targets or a fragmentation occurred.  \emph{MD} stands for ''missed detection'' and \emph{WL} represents ''wrong label'' assignment.}
        \label{fig:examples_tracking}
    \end{center}
\end{figure}

\subsection{Pedestrian Re-Identification}

In a way similar to the previous tasks, the idea is to perceive the relative hardness of performing pedestrian re-identification from UAV-based data, with respect to the levels of performance that are known to be possible by stationary surveillance footage. As such, we selected two well known re-identification algorithms to represent the state-of-the-art and assessed their variation in performance from the stationary to the UAV-based set. The MARS~\cite{Zheng2016} dataset was selected to represent the stationary data, as it is currently the largest video-based source that is freely available. 

According to the results of a challenge on re-identification techniques~\cite{Ye2020}, the GLTR~\cite{Li2019} and COSAM~\cite{Subramaniam2019} were considered to represent the state-of-the-art. The GLTR exploits multi-scale temporal cues in video sequences, by modelling separately short- and long-term features. Short-term components capture the appearance and motion of pedestrians, using parallel dilated convolutions with varying rates. Long-term information is extracted by a temporal self-attention model. The key in COSAM is to capture intra video attention using a co-segmentation module, extracting task-specific regions-of-interest that typically correspond to persons and their accessories. This module is plugged between convolution blocks to induce the notion of co-segmentation, and enables to obtain representations of both the spatial and temporal domains.

For the MARS dataset, the evaluation protocol described in\footnote{\url{http://www.liangzheng.com.cn/Project/project_mars.html}} was used. We considered 1,894 tracklets of 608 IDs, with an average number of frames per tracklet of 67,4. In a 5-fold setting, both datasets were divided into random splits, each one containing the learning, query and gallery sets, in proportions 50:10:40. For the GLTR method, the ResNet50 was used as backbone model, with the learning rate set to 0.01. In the COSAM method, the Se-ResNet50 architecture was used as the backbone model and the COSAM layer was plugged between the forth and fifth convolution layers, with the learning rate set to 0.0001 and the reduction dimension size set to 256.  

The summary results are provided in Table~\ref{tab:reidentification}. In opposition to the detection and tracking problems, no significant decreases in performance were observed between the MARS and P-DESTRE results, which might support for the suitability of the existing re-identification solutions also for UAV-based data. 

\begin{figure}[ht!]
\begin{center}
\begin{tikzpicture}

\def\posX{0}
\def\posY{1.6}
\def\deltaY{0}

\fill [rounded corners, gray] (\posX-1, \posY-0.15+\deltaY) rectangle (\posX+1, \posY+0.15+\deltaY);    
\draw  [white] (\posX, \posY+\deltaY) node {\scriptsize{GLTR}};    

\draw (0,0+\deltaY) node(n1)  {\includegraphics[width=7.5 cm]{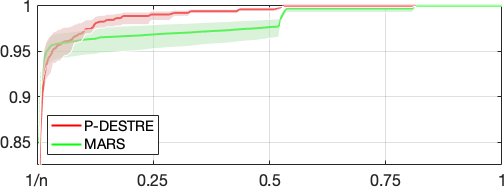}};     
\draw (1.65,0.0+\deltaY+0.05) node(n1)  {\includegraphics[width=4.0 cm]{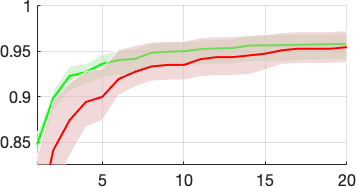}};     
\draw (-3.95, 0+\deltaY) node[rectangle, rotate=90] {\scriptsize{Identification Rate (Hit)}};  
\draw (0,-1.6+\deltaY) node[rectangle] {\scriptsize{Acc. Rank (Penetration)}};  

\def\deltaY{-3.75}
\def\deltaX{0}

\fill [rounded corners, gray] (\posX-1, \posY-0.15+\deltaY) rectangle (\posX+1, \posY+0.15+\deltaY);    
\draw  [white] (\posX, \posY+\deltaY) node {\scriptsize{COSAM}};    
 
\draw (0,0+\deltaY) node(n1)  {\includegraphics[width=7.5 cm]{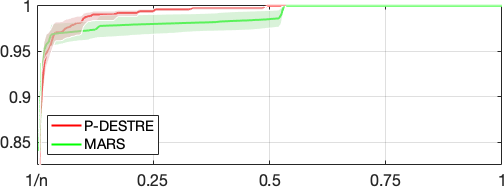}};     
\draw (1.65,0.0+\deltaY+0.05) node(n1)  {\includegraphics[width=4.0 cm]{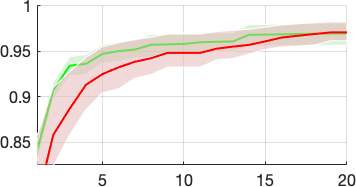}};     
\draw (-3.95, 0+\deltaY) node[rectangle, rotate=90] {\scriptsize{Identification Rate (Hit)}};  
\draw (0,-1.6+\deltaY) node[rectangle] {\scriptsize{Acc. Rank (Penetration)}};  

\draw [very thick, rounded corners, black] (-0.385,-1.025) rectangle (3.675,1.15);     
\draw [very thick, rounded corners, black] (-0.385,-1.025+\deltaY) rectangle (3.675,1.15+\deltaY);     

\end{tikzpicture}
    \caption{Comparison between the closed-set identification (CMC) curves observed in the MARS (green lines) and P-DESTRE (red lines) sets for the GLTR (top plot) and COSAM (bottom plot) re-identification techniques. Zoomed-in regions with the top 1 to 20 results are shown in the inner plots.}
        \label{fig:CMC}
    \end{center}
\end{figure}
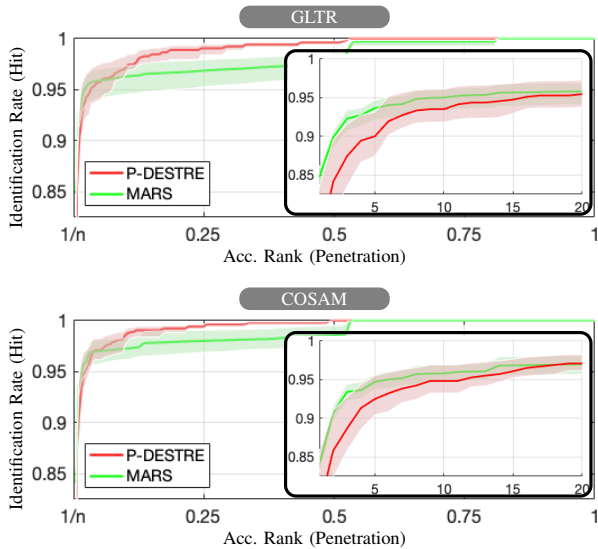

Fig.~\ref{fig:CMC} provides the cumulative rank-n curves for both algorithms and datasets. The red lines represent the P-DESTRE results and the green series denote the MARS values. Results are given in terms of the true identification rate with respect to the proportion of gallery identities retrieved (i.e., equivalent to a hit/penetration plot). It is interesting to note the apparently contradictory results of the GLTR and COSAM algorithms in the MARS and P-DESTRE sets. In both cases, for the top-20 cases, the P-DESTRE results are far worse than the corresponding MARS values. However, for larger ranks (from 5\% of the number of enrolled identities), the P-DESTRE values were solidly better than the ranks observed for MARS. In the case of the former data set, it appears that in case of heavily degraded images, both algorithms tend to produce almost random results, which was not observed for the P-DESTRE. This observation might be justified by the fact that P-DESTRE contains more \emph{poor quality} data than MARS, but does not provide \emph{extremely degraded} samples that almost turn identification into a random process.

\begin{table}[h!]
\centering
     \caption{Comparison between the re-identification performance attained by two state-of-the-art algorithms in the P-DESTRE and MARS data sets.}
     \label{tab:reidentification}
\begin{tabular}{|c|C{1.25cm}|C{1.25cm}|C{1.25cm}|C{1.25cm}|}
\hline
\textbf{\scriptsize{Method}}  & \textbf{\scriptsize{Dataset}}  & \textbf{\scriptsize{mAP}} & \textbf{\scriptsize{Rank-1}}  & \textbf{\scriptsize{Rank-20}}\\ \hline
 \multirow{2}{*}{ \scriptsize{GLTR~\cite{Li2019}}} &   \textbf{\scriptsize{MARS}} & \scriptsize{77.74 $\pm$ 1.07} & \scriptsize{84.72 $\pm$ 2.61} & \scriptsize{95.80 $\pm$ 2.34} \\ \cline{2-5}
& \textbf{\scriptsize{P-DESTRE}} & \scriptsize{77.68 $\pm$ 9.46} & \scriptsize{75.96 $\pm$ 11.77} & \scriptsize{95.48 $\pm$ 3.17} \\ \hline
 \multirow{2}{*}{ \scriptsize{COSAM~\cite{Subramaniam2019}}} &   \textbf{\scriptsize{MARS}} & \scriptsize{78.35 $\pm$ 1.66} & \scriptsize{84.03 $\pm$ 0.91} & \scriptsize{96.97 $\pm$ 0.98} \\ \cline{2-5}
& \textbf{\scriptsize{P-DESTRE}} & \scriptsize{80.64 $\pm$ 9.91} & \scriptsize{79.14 $\pm$ 12.43} & \scriptsize{97.10 $\pm$ 1.85} \\ \hline

\end{tabular}
\end{table}

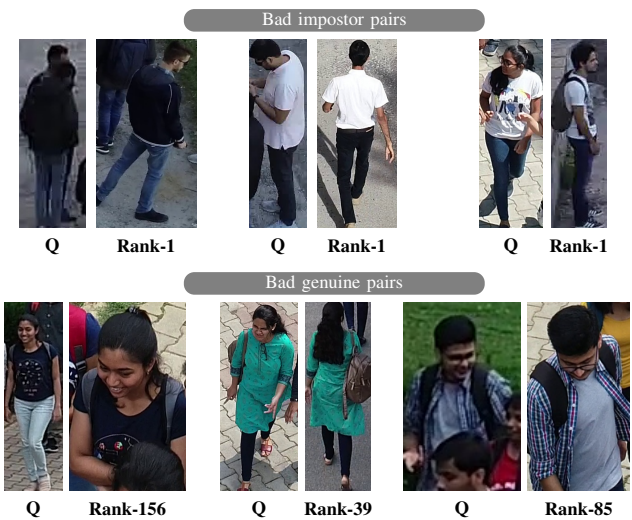
\begin{figure}[ht!]
\begin{center}
\begin{tikzpicture}

\def\posX{4}
\def\posY{1.5}
\def\deltaY{0}

\fill [rounded corners, gray] (\posX-2, \posY-0.15+\deltaY) rectangle (\posX+2, \posY+0.15+\deltaY);    
\draw  [white] (\posX, \posY+\deltaY) node {\scriptsize{Bad impostor pairs}};    

\def\deltaX{0.25}

\draw (0+\deltaX,0) node(n1)  {\includegraphics[height=2.5 cm]{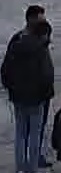}};     
\draw (0.0+\deltaX, -1.5) node[rectangle] {\scriptsize{\textbf{Q}}};

\draw (1.25+\deltaX,0) node(n1)  {\includegraphics[height=2.5 cm]{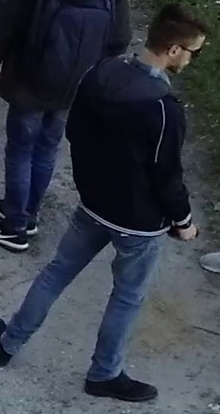}};  
\draw (1.25+\deltaX, -1.5) node[rectangle] {\scriptsize{\textbf{Rank-1}}};     

\draw (3.00+\deltaX,0+\deltaY) node(n1)  {\includegraphics[height=2.5 cm]{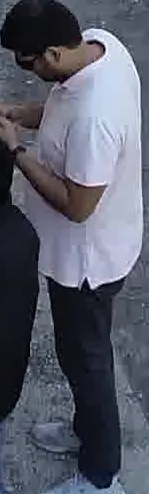}};     
\draw (3.0+\deltaX, -1.5+\deltaY) node[rectangle] {\scriptsize{\textbf{Q}}};  
\draw (4.05+\deltaX,0+\deltaY) node(n1)  {\includegraphics[height=2.5 cm]{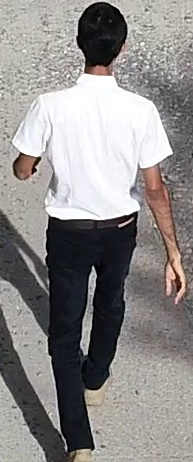}};     
\draw (4.05+\deltaX, -1.5+\deltaY) node[rectangle] {\scriptsize{\textbf{Rank-1}}};  

\draw (6.1+\deltaX,0+\deltaY) node(n1)  {\includegraphics[height=2.5 cm]{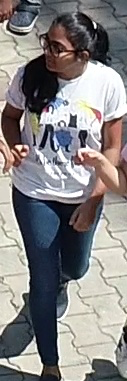}};     
\draw (6.1+\deltaX, -1.5+\deltaY) node[rectangle] {\scriptsize{\textbf{Q}}};  

\draw (7.0+\deltaX,0+\deltaY) node(n1)  {\includegraphics[height=2.5 cm]{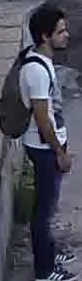}};    
\draw (7.0+\deltaX, -1.5+\deltaY) node[rectangle] {\scriptsize{\textbf{Rank-1}}};   

\def\deltaY{-3.5}
\fill [rounded corners, gray] (\posX-2, \posY-0.15+\deltaY) rectangle (\posX+2, \posY+0.15+\deltaY);    
\draw  [white] (\posX, \posY+\deltaY) node {\scriptsize{Bad genuine pairs}};

\draw (0,0+\deltaY) node(n1)  {\includegraphics[height=2.5 cm]{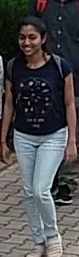}};       
\draw (0.0, -1.5+\deltaY) node[rectangle] {\scriptsize{\textbf{Q}}};  

\draw (1.25,0+\deltaY) node(n1)  {\includegraphics[height=2.5 cm]{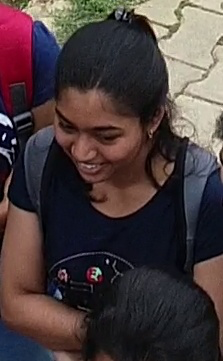}};     
\draw (1.25, -1.5+\deltaY) node[rectangle] {\scriptsize{\textbf{Rank-156}}};  

 
\draw (3.00,0+\deltaY) node(n1)  {\includegraphics[height=2.5 cm]{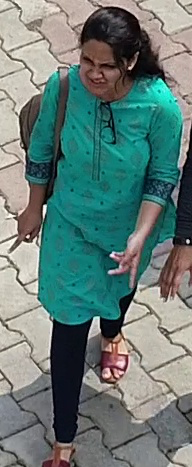}};     
\draw (3.0, -1.5+\deltaY) node[rectangle] {\scriptsize{\textbf{Q}}};  
\draw (4.05,0+\deltaY) node(n1)  {\includegraphics[height=2.5 cm]{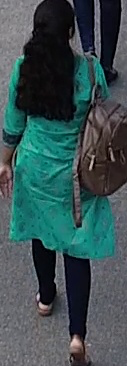}};     
\draw (4.05, -1.5+\deltaY) node[rectangle] {\scriptsize{\textbf{Rank-39}}};  

\draw (5.7,0+\deltaY) node(n1)  {\includegraphics[height=2.5 cm]{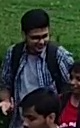}};     
\draw (5.7, -1.5+\deltaY) node[rectangle] {\scriptsize{\textbf{Q}}};  

\draw (7.25,0+\deltaY) node(n1)  {\includegraphics[height=2.5 cm]{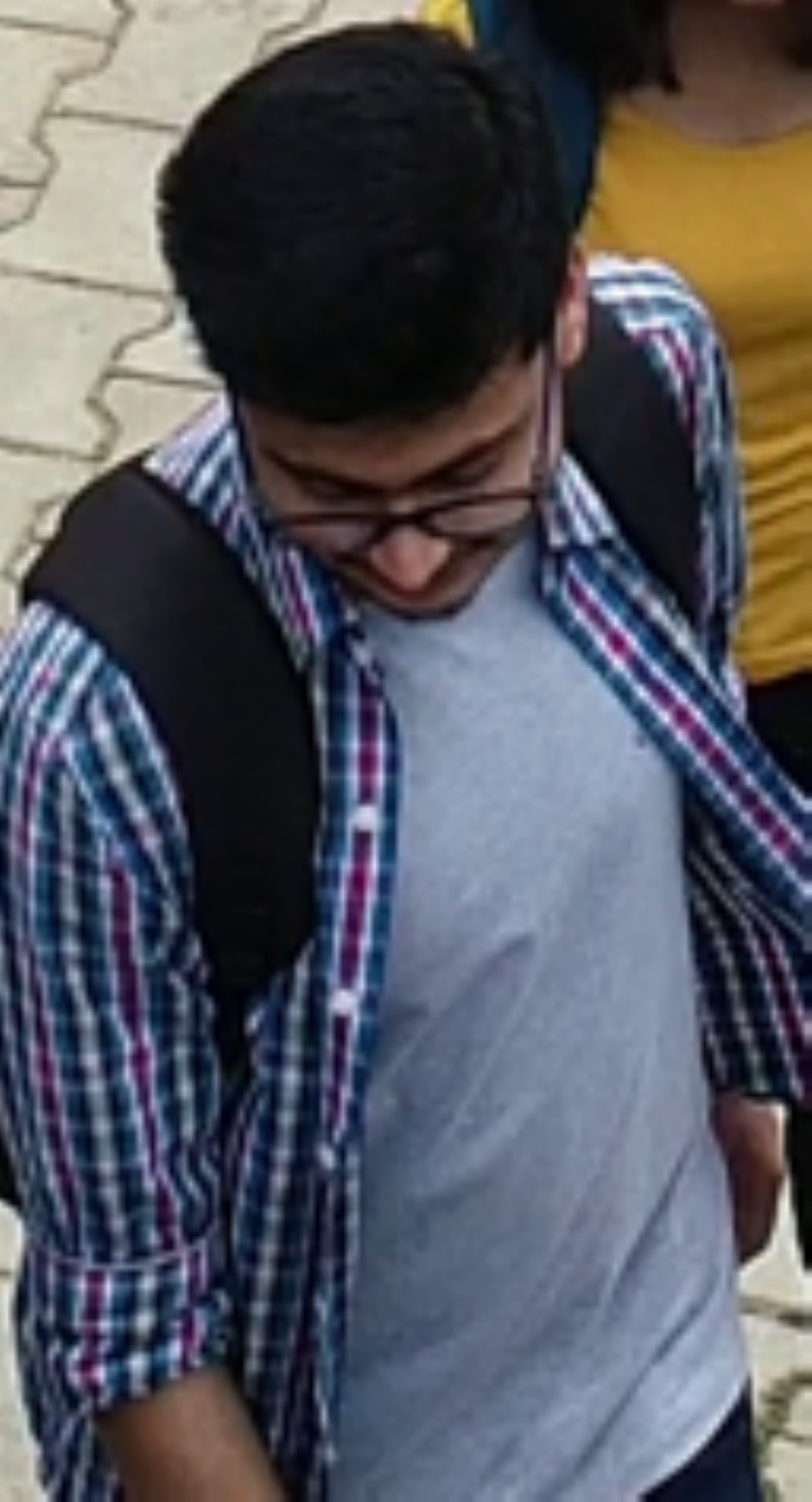}};    
\draw (7.25, -1.5+\deltaY) node[rectangle] {\scriptsize{\textbf{Rank-85}}};   

\end{tikzpicture}
    \caption{Examples of the instances that got the worst re-identification performance. The upper row illustrates typical false matches, almost invariably related with clothing styles and colours. The bottom row provides some examples of cases where (due to differences in pose and scale), the true identities could not be retrieved among the top positions. '''Q'' represents the query image and ''Rank-i'' provides the rank of the corresponding gallery image.}
        \label{fig:cases_reid}
    \end{center}
\end{figure}

Based in these experiments, and according to our subjective evaluation, Fig.~\ref{fig:cases_reid} highlights some of the notorious cases for re-identification purposes. The upper row represents the particularly hazardous cases in terms of \emph{convenience},  where different IDs were erroneously perceived as the same. As can be seen, this was mostly due to similarities in clothing appearance, together with the sharing of most soft biometric labels between the confounded IDs. The bottom row provides the particularly dangerous cases for \emph{security} purposes, where both algorithms had difficulties in identifying a known ID.  It can be seen that such cases often yielded from  notorious differences in pose and scale between the query and gallery data. Along with the background clutter, these factors decrease the effectiveness of the feature representations, and were observed to be among the most concerning for re-identification performance.  

\subsection{Pedestrian Search}

As stated above, the pedestrian search problem was the main motivation for the development of the P-DESTRE dataset. 
Here, in opposition to the re-identification setting, there is not any guarantee about the clothing appearance of subjects, nor about the time elapsed between consecutive observations of one ID. Under such circumstances, the analysis of clothing appearance becomes meaningless and other features should be privileged (i.e., face, gait or soft-biometrics based). 

Considering that there are not methods in the literature specifically designed for the pedestrian search task, we have chosen a combination of two well-known human identification techniques, which combine face and body features. In a way similar to the previous tasks analysed, the idea is to provide an approximation for the effectiveness attained by the existing solutions in UAV-based data. Such levels of performance should provide a baseline for this problem, and can be used as basis for further developments in this topic. 

The facial regions-of-interest were detected by the SSH method~\cite{Najibi2017} (with acceptance threshold set to 0.7), from where a feature representation was obtained using the ArcFace~\cite{Deng2019} model. For the body-based analysis, the COSAM~\cite{Subramaniam2019}  identification model provided the feature representation. Both models were trained \emph{from scratch}. The data were sampled into 5 trials, each one containing learning + gallery + query instances in proportions 50:10:40. In the ArcFace method, MobileNetV2 was used as backbone model, and the learning rate set to 0.01. In COSAM, the Se-ResNet50 was used as backbone model, and the COSAM layer plugged into the forth and fifth convolutional layers, with learning rate equal to 1e$^{-4}$ and reduction dimension size equal to 256.  Each model was trained in a separate way, and during the test phase, the mean value of the ArcFace facial features in the tracklet were appended to the body-based representation yielding from COSAM. The Euclidean norm was used as distance function between such concatenated representations.

\begin{figure}[ht!]
\begin{center}
\begin{tikzpicture}

\def\posX{0}
\def\posY{1.6}
\def\deltaY{0}

\fill [rounded corners, gray] (\posX-2, \posY-0.15+\deltaY) rectangle (\posX+2, \posY+0.15+\deltaY);    
\draw  [white] (\posX, \posY+\deltaY) node {\scriptsize{ArcFace~\cite{Deng2019} + COSAM~\cite{Subramaniam2019}}};    

\draw (0,0+\deltaY) node(n1)  {\includegraphics[width=7.5 cm]{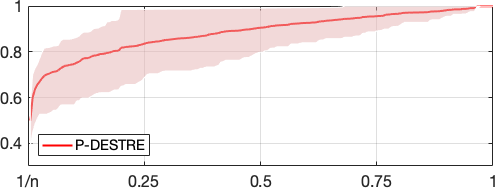}};     
\draw (1.65,0.0+\deltaY+0.05) node(n1)  {\includegraphics[width=4.0 cm]{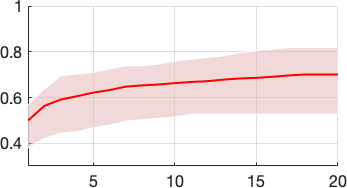}};     
\draw (-3.95, 0+\deltaY) node[rectangle, rotate=90] {\scriptsize{Identification Rate (Hit)}};  
\draw (0,-1.6+\deltaY) node[rectangle] {\scriptsize{Acc. Rank (Penetration)}};  

\draw [very thick, rounded corners, black] (-0.385,-1.025) rectangle (3.675,1.15);     
 
\end{tikzpicture}
    \caption{Closed-set identification (CMC) curves obtained for theP-DESTRE. The zoomed-in region with the top-20 results is shown in the inner plot.}
        \label{fig:CMC_search}
    \end{center}
\end{figure}
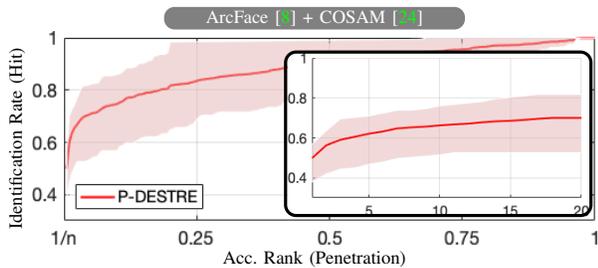

Fig.~\ref{fig:CMC_search} provides the cumulative rank-n curves obtained the P-DESTRE set, in terms of the identification rate with respect to the proportion of gallery identities (i.e., hit/penetration plot). As expected, when compared to the re-identification setting, the performance was substantially lower (rank-1 $\approx$ 79.14\% for re-identification $\rightarrow$ $\approx$ 49.88\% for serach), which accords the human perception for the additional difficulty of \emph{search} with respect to \emph{re-identify}. 

\begin{table}[h!]
\centering
     \caption{Baseline person search performance obtained by an ensemble of  ArcFace~\cite{Deng2019} + COSAM~\cite{Subramaniam2019} in the P-DESTRE data set.}
     \label{tab:search}
\begin{tabular}{|c|C{1.25cm}|C{1.25cm}|C{1.25cm}|}
\hline
\textbf{\scriptsize{Method}}    & \textbf{\scriptsize{mAP}} & \textbf{\scriptsize{Rank-1}}  & \textbf{\scriptsize{Rank-20}}\\ \hline
\scriptsize{ArcFace~\cite{Deng2019} + COSAM~\cite{Subramaniam2019}} & \scriptsize{34.90 $\pm$ 6.43} & \scriptsize{49.88 $\pm$ 8.01} & \scriptsize{70.10 $\pm$ 11.25} \\ \hline
\end{tabular}
\end{table}

\begin{figure}[ht!]
\begin{center}
\begin{tikzpicture}

\def\posX{4}
\def\posY{1.5}
\def\deltaY{0}

\fill [rounded corners, gray] (\posX-2, \posY-0.15+\deltaY) rectangle (\posX+2, \posY+0.15+\deltaY);    
\draw  [white] (\posX, \posY+\deltaY) node {\scriptsize{Good genuine pairs}};    

\def\deltaX{0.25}

\draw (0+\deltaX,0) node(n1)  {\includegraphics[height=2.5 cm]{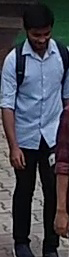}};     
\draw (0.0+\deltaX, -1.5) node[rectangle] {\scriptsize{\textbf{Q}}};

\draw (0.85+\deltaX,0) node(n1)  {\includegraphics[height=2.5 cm]{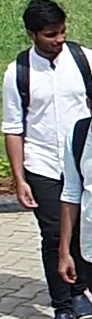}};  
\draw (0.85+\deltaX, -1.5) node[rectangle] {\scriptsize{\textbf{Rank-1}}};

\draw (2.85+\deltaX,0+\deltaY) node(n1)  {\includegraphics[height=2.5 cm]{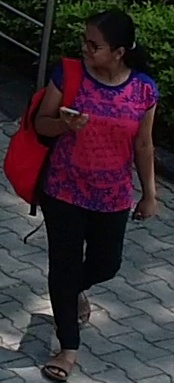}};     
\draw (2.85+\deltaX, -1.5+\deltaY) node[rectangle] {\scriptsize{\textbf{Q}}};  

\draw (4.05+\deltaX,0+\deltaY) node(n1)  {\includegraphics[height=2.5 cm]{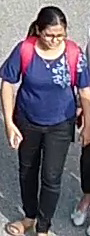}};     
\draw (4.05+\deltaX, -1.5+\deltaY) node[rectangle] {\scriptsize{\textbf{Rank-1}}};  

\draw (5.8+\deltaX,0+\deltaY) node(n1)  {\includegraphics[height=2.5 cm]{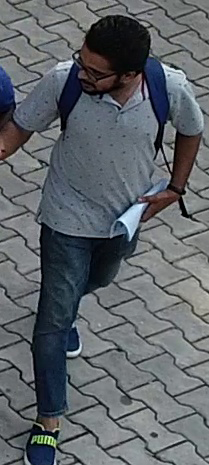}};     
\draw (5.8+\deltaX, -1.5+\deltaY) node[rectangle] {\scriptsize{\textbf{Q}}};  

\draw (7.0+\deltaX,0+\deltaY) node(n1)  {\includegraphics[height=2.5 cm]{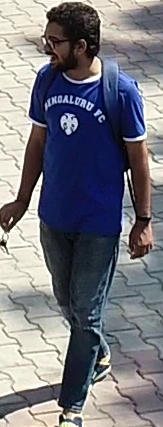}};    
\draw (7.0+\deltaX, -1.5+\deltaY) node[rectangle] {\scriptsize{\textbf{Rank-1}}};   

\def\deltaY{-3.5}
\def\deltaX{0.25}

\fill [rounded corners, gray] (\posX-2, \posY-0.15+\deltaY) rectangle (\posX+2, \posY+0.15+\deltaY);    
\draw  [white] (\posX, \posY+\deltaY) node {\scriptsize{Bad impostor pairs}};    

\draw (0+\deltaX,0+\deltaY) node(n1)  {\includegraphics[height=2.5 cm]{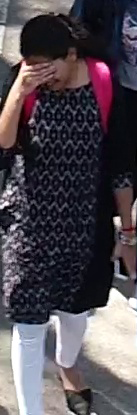}};       
\draw (0.0+\deltaX, -1.5+\deltaY) node[rectangle] {\scriptsize{\textbf{Q}}};  

\draw (1.0+\deltaX,0+\deltaY) node(n1)  {\includegraphics[height=2.5 cm]{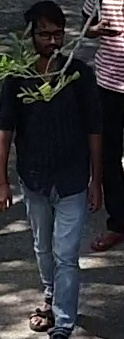}};     
\draw (1.0+\deltaX, -1.5+\deltaY) node[rectangle] {\scriptsize{\textbf{Rank-1}}};  

 
\draw (3.00,0+\deltaY) node(n1)  {\includegraphics[height=2.5 cm]{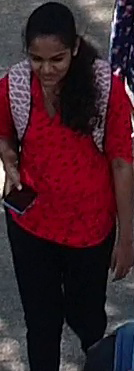}};     
\draw (3.0, -1.5+\deltaY) node[rectangle] {\scriptsize{\textbf{Q}}};  

\draw (4.15,0+\deltaY) node(n1)  {\includegraphics[height=2.5 cm]{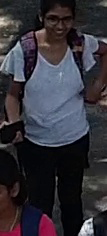}};     
\draw (4.15, -1.5+\deltaY) node[rectangle] {\scriptsize{\textbf{Rank-1}}};  

\draw (6.1,0+\deltaY) node(n1)  {\includegraphics[height=2.5 cm]{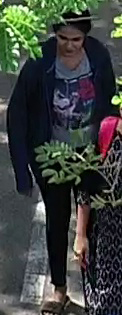}};     
\draw (6.1, -1.5+\deltaY) node[rectangle] {\scriptsize{\textbf{Q}}};  

\draw (7.15,0+\deltaY) node(n1)  {\includegraphics[height=2.5 cm]{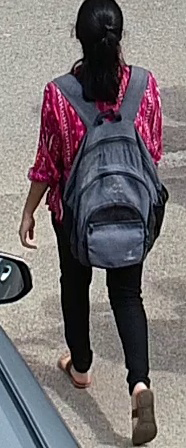}};    
\draw (7.15, -1.5+\deltaY) node[rectangle] {\scriptsize{\textbf{Rank-1}}};   

\def\deltaY{-7.0}
\fill [rounded corners, gray] (\posX-2, \posY-0.15+\deltaY) rectangle (\posX+2, \posY+0.15+\deltaY);    
\draw  [white] (\posX, \posY+\deltaY) node {\scriptsize{Bad genuine pairs}};

\draw (0+0.35,0+\deltaY) node(n1)  {\includegraphics[height=2.5 cm]{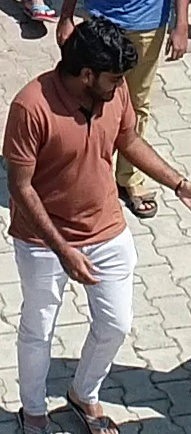}};       
\draw (0.0+0.35, -1.5+\deltaY) node[rectangle] {\scriptsize{\textbf{Q}}};  

\draw (1.5,0+\deltaY) node(n1)  {\includegraphics[height=2.5 cm]{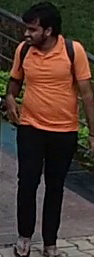}};     
\draw (1.5, -1.5+\deltaY) node[rectangle] {\scriptsize{\textbf{Rank-56}}};  

 
\draw (3.20,0+\deltaY) node(n1)  {\includegraphics[height=2.5 cm]{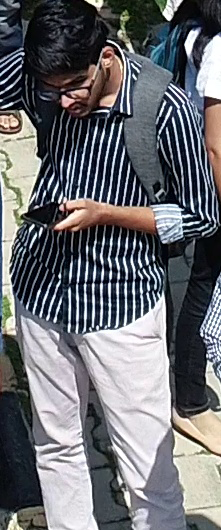}};     
\draw (3.2, -1.5+\deltaY) node[rectangle] {\scriptsize{\textbf{Q}}};  

\draw (4.25,0+\deltaY) node(n1)  {\includegraphics[height=2.5 cm]{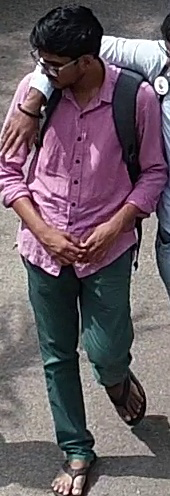}};     
\draw (4.25, -1.5+\deltaY) node[rectangle] {\scriptsize{\textbf{Rank-73}}};  

\draw (5.9,0+\deltaY) node(n1)  {\includegraphics[height=2.5 cm]{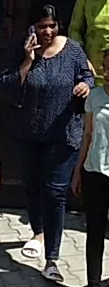}};     
\draw (5.9, -1.5+\deltaY) node[rectangle] {\scriptsize{\textbf{Q}}};  

\draw (7.05,0+\deltaY) node(n1)  {\includegraphics[height=2.5 cm]{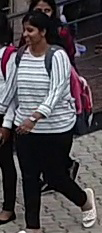}};    
\draw (7.05, -1.5+\deltaY) node[rectangle] {\scriptsize{\textbf{Rank-98}}};

\end{tikzpicture}
    \caption{Examples of the instances where good/poorest person search performance was observed. The upper row illustrates particularly successful cases, while the bottom rows show pairs of images where the used algorithm had notorious difficulties to retrieve the correct identity. '''Q'' represents the query image and ''Rank-i'' provides the rank of the retrieved gallery image.}
        \label{fig:cases_search}
    \end{center}
\end{figure}
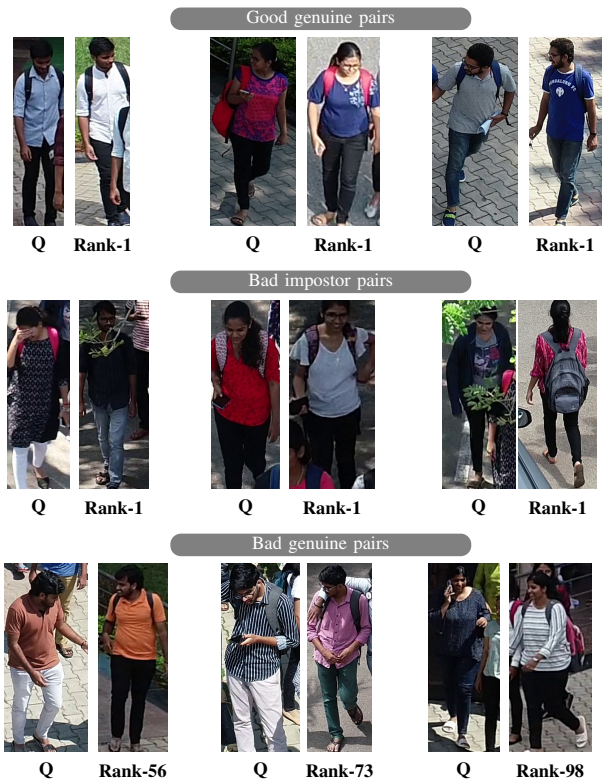

Based in our qualitative analysis of the results, Fig.~\ref{fig:cases_search} provides three different types of examples: the upper row shows some successful identification processes, in which the model was able to retrieve the true identity in the first position. In opposition, the second row provides examples of particularly hazardous cases, in which due to similarities in pose, accessories and soft biometric labels between the query and gallery images, false matches have occurred. Finally, the bottom row provides examples of cases where the corresponding identities of the queries were retrieved in high positions (ranks 56, 73 and 98), i.e., which represent the maximum threat to security, as the system failed to detect a particular subject of interest in a crowd.   

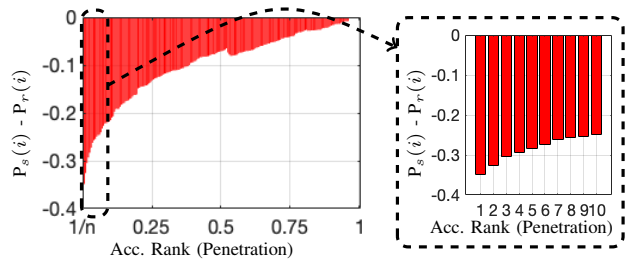
\begin{figure}[ht!]
\begin{center}
\begin{tikzpicture}

\def\posX{0}
\def\posY{1.6}
\def\deltaY{0}

\draw (0,0+\deltaY) node(n1)  {\includegraphics[width=4.25 cm]{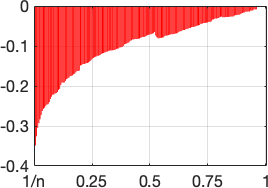}};     
\draw (-2.4, 0+\deltaY) node[rectangle, rotate=90] {\scriptsize{P$_s(i)$ - P$_r(i)$}};  
\draw (0,-1.7+\deltaY) node[rectangle] {\scriptsize{Acc. Rank (Penetration)}};  

\draw (4.25,0+\deltaY) node(n1)  {\includegraphics[width=2.4 cm]{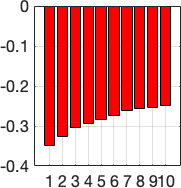}};     
\draw (4.25,-1.4+\deltaY) node[rectangle] {\scriptsize{Acc. Rank (Penetration)}};  
\draw (5.25-2.4, 0+\deltaY) node[rectangle, rotate=90] {\scriptsize{P$_s(i)$ - P$_r(i)$}};  

\draw [rounded corners, very thick, dashed] (-1.6, -1.25) rectangle (-1.25, 1.5);    

\draw [rounded corners, very thick, dashed] (2.6, -1.65) rectangle (5.6, 1.4);    

\draw [very thick, dashed, ->] (-1.25, 0.5) ..controls (1, 1.75) .. (2.6, 1);
 
\end{tikzpicture}
    \caption{Differences between the probability of retrieving the true identity of a query among the top-$i$ positions, $\forall i \in \{1,\ldots,100\}$, for the person search (P$_s$) and re-identification (P$_r$) problems.}
        \label{fig:summary}
    \end{center}
\end{figure}

As concluding remark, the challenges of person search are illustrated in Fig.~\ref{fig:summary}, providing the differences between the probabilities of obtaining a top-$i$ true identification (hit), $\forall i \in \{1,\ldots,n\}$, i.e., retrieve the identity corresponding to a query up to the i$^{th}$ position, for the search and re-identification problems. Here, P$_s(i)$ and P$_r(i)$ denote the probabilities of observing a \emph{hit} in the search P$_s$ and re-identification P$_r$ tasks, i.e., negative \big(P$_s(i)$ - P$_r(i)$\big) denote higher probabilities for re-identification success than for search success. The zoomed-in region given at the right part of the Figure shows the additional difficulty (of almost 40 percentual points) in retrieving the true identity in a single shot (difference between top-1 values). Then, the gap between the accumulated values of P$_s$ and P$_r$ decreases in a monotonous way, and only approaches 0 near the full penetration rate, i.e., when all the known identities are retrieved for a query. In summary, it is much more difficult to identify pedestrians when no clothing information can be used, which paves the way for further developments in this kind of technology. According to our goals in developing this data source, the P-DESTRE set is a tool to support such advances in the state-of-the-art.

 \section{Conclusions}
 \label{sec:Conclusions}
 
In this paper we announced the free availability of the P-DESTRE dataset, which provides video sequences of pedestrians in outdoor urban environments, taken from UAVs and fully annotated at the frame level. Accordingly, our main contributions are two-fold: 1) we provide consistent ID annotations across the observations taken in different days, which is a singularity with respect to previous related sets, and makes the P-DESTRE suitable for the extremely challenging problem of \emph{person search}, i.e., when no clothing-based features can be used for identification purposes. The dataset is also suitable for research on pedestrian detection, tracking, re-identification and soft biometrics; and 2) having carried out a reproducible evaluation of state-of-the-art pedestrian detection, tracking, re-identification and search techniques, we report the performance values attained by such methods in well-known datasets with respect to their effectiveness in UAV-based data. Overall, such experiments point for a consistent degradation in performance for the detection (among all tasks), tracking and search tasks when working with UAV-based data. The exception was the re-identification problem, where the already existing solutions attain results in UAV-based data that are similar to the obtained in data acquired from stationary devices.  As such, further efforts are required to advance the state-of-the-art in pedestrian detection, tracking and search for UAV-based data. The P-DESTRE initiative provided the data and the baselines to support such efforts. 

\section*{Acknowledgements}

This work is funded by FCT/MEC through national funds and  co-funded by FEDER - PT2020 partnership agreement under the projects UID/EEA/50008/2019, POCI-01-0247-FEDER-033395 and C4: Cloud Computing Competence Centre.

{\small
\bibliographystyle{ieee}

}

\end{document}